%% file: main.tex
\documentclass{article}

\usepackage{ijcai25}

\usepackage{times}
\usepackage{soul}
\usepackage{url}
\usepackage[hidelinks]{hyperref}
\usepackage[utf8]{inputenc}
\usepackage[small]{caption}
\usepackage{graphicx}
\usepackage{amsmath}
\usepackage{amsthm}
\usepackage{booktabs}
\usepackage{algorithm}
\usepackage{algorithmic}
\usepackage[switch]{lineno}

\usepackage{graphicx} 
\usepackage{natbib}  
\usepackage{caption} 
\usepackage{algorithm}
\usepackage{algorithmic}

\usepackage{newfloat}
\usepackage{listings}

\usepackage{microtype}
\usepackage{subcaption}
\usepackage{booktabs} 
\usepackage{amsmath}
\usepackage{amssymb}
\usepackage{mathtools}
\usepackage{amsthm}

\newtheorem{theorem}{Theorem}
\newtheorem{assumption}[theorem]{Assumption}
\newtheorem{lemma}[theorem]{Lemma}
\newtheorem{definition}[theorem]{Definition}

\usepackage[textsize=tiny]{todonotes}

\usepackage{import}
\input{math_commands}

\usepackage{adjustbox}
\usepackage{multirow}
\usepackage{colortbl}
\usepackage{enumitem}
\usepackage{lipsum}
\usepackage[capitalize,noabbrev]{cleveref}

\title{A primal-dual perspective for distributed TD-learning}
\author{
Han-Dong Lim \and Donghwan Lee \\ 
\affiliations Department of Electrical Engineering, KAIST\\
\emails \{limaries30, donghwan\}@kaist.ac.kr  }

\begin{document}

\maketitle



\begin{abstract}
The goal of this paper is to investigate distributed temporal difference (TD) learning for a networked multi-agent Markov decision process. The proposed approach is based on distributed optimization algorithms, which can be interpreted as primal-dual ordinary differential equation (ODE) dynamics subject to null-space constraints. Based on the exponential convergence behavior of the primal-dual ODE dynamics subject to null-space constraints, we examine the behavior of the final iterate in various distributed TD-learning scenarios, considering both constant and diminishing step-sizes and incorporating both i.i.d. and Markovian observation models. Unlike existing methods, the proposed algorithm does not require the assumption that the underlying communication network structure is characterized by a doubly stochastic matrix.
\end{abstract}


\section{Introduction}
\import{intro}{intro}

\section{Preliminaries}
\import{./prelim}{mdp}
\import{./prelim}{mamdp}

\section{Analysis of primal-dual gradient dynamics}\label{sec:pd}
\import{./ode}{ode}
\section{Distributed TD-learning }\label{sec:distributed_td}
\import{./dtd}{dtd}
\subsection{Markovian observation case}\label{sec:sa_markov}
\import{./dtd}{dtd_markov}
\section{Experiments}\label{sec:exp}
\import{./experiments}{main}

\section{Conclusion}
\import{./conclusion}{main}
\section*{Acknowledgements}
The work was supported by the Institute of Information Communications Technology Planning Evaluation (IITP) funded by the Korea government under Grant 2022-0-00469. 




\bibliographystyle{named}
\bibliography{bilbio}

\import{./}{app_arxiv.tex}





\end{document}

%% file: math_commands.tex
\usepackage{amsmath,amsfonts,bm}
\usepackage{pifont}
\newcommand{\xmark}{\text{\ding{55}}}

\def\eqref#1{equation~\ref{#1}}

\def\1{\bm{1}}

\def\eps{{\epsilon}}

\def\vtheta{{\bm{\theta}}}
\def\va{{\bm{a}}}
\def\vb{{\bm{b}}}

\def\ve{{\bm{e}}}

\def\vv{{\bm{v}}}
\def\vw{{\bm{w}}}
\def\vx{{\bm{x}}}

\def\vz{{\bm{z}}}

\def\mA{{\bm{A}}}
\def\mB{{\bm{B}}}

\def\mD{{\bm{D}}}
\def\mE{{\bm{E}}}

\def\mG{{\bm{G}}}
\def\mH{{\bm{H}}}
\def\mI{{\bm{I}}}

\def\mL{{\bm{L}}}
\def\mM{{\bm{M}}}

\def\mP{{\bm{P}}}
\def\mQ{{\bm{Q}}}
\def\mR{{\bm{R}}}
\def\mS{{\bm{S}}}
\def\mT{{\bm{T}}}
\def\mU{{\bm{U}}}
\def\mV{{\bm{V}}}
\def\mW{{\bm{W}}}

\def\mPhi{{\bm{\Phi}}}

\DeclareMathAlphabet{\mathsfit}{\encodingdefault}{\sfdefault}{m}{sl}
\SetMathAlphabet{\mathsfit}{bold}{\encodingdefault}{\sfdefault}{bx}{n}

\def\gA{{\mathcal{A}}}

\def\gC{{\mathcal{C}}}

\def\gE{{\mathcal{E}}}
\def\gF{{\mathcal{F}}}
\def\gG{{\mathcal{G}}}

\def\gN{{\mathcal{N}}}
\def\gO{{\mathcal{O}}}
\def\gP{{\mathcal{P}}}

\def\gR{{\mathcal{R}}}
\def\gS{{\mathcal{S}}}

\def\gV{{\mathcal{V}}}

\def\gX{{\mathcal{X}}}

\def\sN{{\mathbb{N}}}

\def\sP{{\mathbb{P}}}

\newcommand{\E}{\mathbb{E}}

\newcommand{\R}{\mathbb{R}}



%% file: intro/intro.tex
Temporal-difference (TD) learning~\citep{sutton1988learning} aims to solve the policy evaluation problem in Markov decision processes (MDPs), serving as the foundational pillar for many reinforcement learning (RL) algorithms~\citep{mnih2015human}. Following the empirical success of RL in various fields~\citep{kober2013reinforcement,li2019deep}, theoretical exploration of TD-learning has become an active area of research. For instance, \citealp{tsitsiklis1996analysis} studied the asymptotic convergence of TD-learning, while non-asymptotic analysis has been examined in \citealp{bhandari2018finite,srikant2019finite,lee2022analysis}.

In contrast to the single-agent case, the theoretical understanding for TD-learning for networked multi-agent Markov decision processes (MAMDPs) has not been fully explored so far. In the networked MAMDPs, each agent follows its own policy and receives different local rewards while sharing their local learning parameters through communication networks. Under this scenario, several distributed TD-learning algorithms~\citep{wang2020decentralized,doan2019finite,doan2021finite,sun2020finite,zeng2022finite} have been developed based on distributed optimization frameworks~\citep{nedic2009distributed,pu2021distributed}.


The main goal of this paper is to provide finite-time analysis of a distributed TD-learning algorithm for networked MAMDPs from the perspectives of the primal-dual algorithms~\citep{wang2011control,mokhtari2016dsa}. The proposed algorithms are inspired by the control system model for distributed optimization problems~\citep{wang2011control,lee2023distributed}, and at the same time, it can also be interpreted as the primal-dual gradient dynamics in~\citealp{qu2018exponential}. In this respect, we first study finite-time analysis of continuous-time primal-dual gradient dynamics in~\citealp{qu2018exponential} with special nullity structures on the system matrix. Based on the analysis of primal-dual gradient dynamics, we further provide a finite-time analysis of the proposed distributed TD-learning under both i.i.d. observation and Markov observation models. The main contributions are summarized as follows:
\begin{enumerate}
    \item An improved or comparable to the state of art convergence rate for continuous-time primal-dual gradient dynamics~\citep{qu2018exponential} with null-space constraints under specific conditions: the results can be applied to general classes of distributed optimization problems that can be reformulated as saddle-point problems~\citep{wang2011control,mokhtari2016dsa};
    \item Development of new distributed TD-learning algorithm inspired by~\citealp{wang2011control,lee2023distributed}, which does not require a double stochastic matrix. This offers a significant advantage in specific scenarios, such as wireless ad hoc networks or broadcast-based communication, where node degrees (number of neighbours) are often unknown due to factors like message loss during transmission~\citep{hendrickx2015fundamental}. This uncertainty makes it challenging to construct a doubly stochastic matrix, as most existing methods rely on precise knowledge of node degrees. In contrast, our algorithm does not require such additional information and thus remains effective in these environments;
    \item New mean-squared error bounds of the distributed TD-learning under our consideration for both i.i.d. and Markovian observation models and under various conditions of the step-sizes: the distributed TD-learning is based on the control system model in~\citealp{wang2011control,lee2023distributed} which does not require doubly stochastic matrix corresponding to its associated network graph. Note that the doubly stochastic assumption is required in other distributed TD-learning algorithms based on the classical distributed optimization algorithms~\citep{nedic2009distributed,pu2021distributed}; 
    \item Empirical demonstrations of both the convergence and the rate of convergence of the algorithm are provided.
\end{enumerate}

\textbf{Related Works.} 
Distributed optimization has been an active research field. In this context,~\citealp{nedic2009distributed} investigated a distributed optimization algorithm over a communication network whose structure graph is represented by a doubly stochastic matrix. In this approach, each agent exchanges information with its neighbors, with the exchange being weighted by the corresponding element in the doubly stochastic matrix. Meanwhile,~\citealp{wang2011control,notarnicola2023gradient} provided control system approach to study distributed optimization problem.

The asymptotic convergence of distributed TD-learning has been studied in~\citealp{mathkar2016distributed,stankovic2023distributed}.~\citealp{doan2019finite} provided finite-time analysis of distributed TD-learning based on the distributed optimization algorithm~\citep{nedic2009distributed} with i.i.d. observation model. Their analysis was extended to the Markovian observation model~\citep{doan2021finite}.~\citealp{sun2020finite} studied distributed TD-learning based on~\citealp{nedic2009distributed} with the Markovian observation model using multi-step Lyapunov function~\citep{wang2019multistep}.~\citealp{wang2020decentralized} studied distributed TD-learning motivated by the gradient tracking method~\citep{pu2021distributed}.~\citealp{zeng2022finite} studied finite-time behavior of distributed stochastic approximation algorithms~\citep{robbins1951stochastic} with general mapping including TD-learning and Q-learning, using Lyapunov-Razumikhin function~\citep{zhou2018improved}.
 
In the context of policy evaluation,~\citealp{macua2014distributed,lee2018primal,wai2018multi,cassano2020multiagent} studied distributed versions of gradient-TD~\citep{sutton2009fast}. The Gradient-TD method can be reformulated as saddle-point problem~\citep{macua2014distributed,lee2022new}, and the aforementioned works can be understood as distributed optimization over a saddle-point problem~\citep{boyd2004convex}.

%% file: prelim/mdp.tex
\subsection{Markov decision process}
Markov decision process (MDP) consists of five tuples \((\gS,\gA,\gamma,\gP,r)\), where \(\gS:=\{1,2,\dots,|\gS|\}\) is the collection of states, \(\gA\) is the collection of actions, \(\gamma\in(0,1)\) is the discount factor, \(\gP:\gS\times\gA\times\gS\to[0,1]\) is the transition kernel, and \(r :\gS\times\gA\times\gS \to \R\) is the reward function. If action \(a\in \gA\) is chosen at state \(s\in\gS\), the transition to state \(s^{\prime}\in\gS\) occurs with probability \(\gP(s,a,s^{\prime})\), and incurs reward \(r(s,a,s^{\prime})\). Given a stochastic policy \(\pi:\gS\times\gA \to [0,1]\), the quantity \(\pi(a\mid s)\) denotes the probability of taking action \(a\in\gA\) at state \(s\in\gS\). We will denote \(\gP^{\pi}(s,s^{\prime}):=\sum_{a\in\gA} \gP(s,a,s^{\prime}) \pi(a\mid s)\), and \(\gR^{\pi}(s):=\sum_{a\in\gA}\sum_{s^{\prime}\in\gS}\gP(s,a,s^{\prime})\pi(a\mid s)r(s,a,s^{\prime}) \), which is the transition probability from state \(s\in\gS\) to \(s^{\prime}\in\gS \) under policy \(\pi\), and expected reward at state \(s\in\gS\), respectively. \(d:\gS\to [0,1] \) denotes the stationary distribution of the state \(s\in \gS \) under policy \(\pi\). The policy evaluation problem aims to estimate the expected sum of discounted rewards following policy \(\pi\), the so-called the value function, \(  v^{\pi}(s)=\E\left[\sum_{k=0}^{\infty} \gamma^k r(s_k,a_k,s_{k+1})\middle | s_0 = s, \pi \right]\) for \(s\in\gS\). 

Given a feature function \(\bm{\phi}:\gS\to\R^q\), our aim is to estimate the value function through learnable parameter \(\vtheta\), i.e., \(v^{\pi}(s)\approx  \bm{\phi}(s)^{\top}\vtheta\), for \(s\in\gS\), which can be achieved through solving the optimization problem, \(
\min_{\vtheta \in \R^q} \frac{1}{2}\left\| \mR^{\pi}+\gamma\mP^{\pi}\mPhi\vtheta-\mPhi\vtheta \right\|^2_{\mD^{\pi}}\), where \(\mD^{\pi}\) is a diagonal matrix whose elements are \(d(1),d(2),\dots,d(|\gS|) \), \(\mP^{\pi}\in\R^{|\gS|\times|\gS|}\) whose elements are \([\mP^{\pi}]_{ij}:=\gP^{\pi}(i,j)\) for \(i,j\in\gS\), \(\mR^{\pi}\in\R^{|\gS|}\), \([\mR^{\pi}]_i:=\E_{}\left[ r(s,a,s^{\prime}) \middle|s=i\right]\) for \(i\in\gS\), and \(\mPhi:= \begin{bmatrix}
    \bm{\phi}(1) & \bm{\phi}(2) & \cdots & \bm{\phi}(|\gS|)
\end{bmatrix}^{\top} \in\R^{|\gS|\times q}\). The solution of the optimization problem satisfies the so-called projected Bellman equation~\citep{sutton2009fast}: 
\begin{align*}
    \mPhi^{\top}\mD^{\pi}\mPhi\vtheta = \mPhi^{\top}\mD^{\pi}\mR^{\pi}+\gamma \mPhi^{\top}\mD^{\pi}\mP^{\pi}\mPhi\vtheta. 
\end{align*}
Throughout the paper, we adopt the common assumption on the feature matrix, which is widely used in the literature~\citep{bhandari2018finite,wang2020decentralized}.
\begin{assumption}\label{assmp:feature}
    \( \left\|\bm{\phi}(s) \right\|_2\leq 1\) for all \(s\in\gS \) and \(\mPhi\) is full-column rank matrix.
\end{assumption}

%% file: prelim/mamdp.tex
\subsection{Multi-agent MDP}
Multi-agent Markov decision process (MAMDP) considers a set of agents cooperatively computing the value function for a shared environment. Considering \(N\) agents, each agent can be denoted by \(i\in \gV:=\{1,2,\dots, N\}\), and the agents communicate over networks that can be described by a connected and undirected simple graph \(\gG:=(\gV,\gE) \), where \( \gE \subset \gV \times \gV\) is the set of edges. \(\gN_i \subset \gV\) denotes the neighbour of agent \(i \in \gV\), i.e., \(j \in \gN_i\) if and only if \((i,j) \in \gE\) for \(i,j\in \gV\). Each agent \(i\in\gV\) has its local policy \(\pi^i:\gS\times\gA_i \to [0,1]\), where \(\gA_i\) is the action space of agent \(i\), and receives reward following its local reward function \(r^i:\gS\times\gA \times \gS \to \R\) where $\gA:=\Pi^N_{i=1}\gA_i$.  MAMDP consists of five tuples \((\gS,\gA,\gamma ,\gP,\{r^i\}_{i=1}^N)\), where \(\gP:\gS\times\gA\times \gS \to [0,1]\) is the Markov transition kernel. The agents share the same state \(s\in \gS\), and when action \(\va:=(a_1,a_2,\dots,a_N)\in \gA\) is taken, the state transits to \(s^{\prime}\in \gS \) with probability \(\gP(s,\va,s^{\prime})\), and for \(i\in \gV\), agent \(i\) receives \( r^i(s,\va,s^{\prime})\). The aim of the policy evaluation under MAMDP is to estimate the expected sum of discounted rewards averaged over \(N\) agents, i.e., \(v^{\pi}(s)=\E\left[\sum^{\infty}_{k=0}\gamma^k \frac{1}{N} \sum_{i=1}^N r^i(s_k,\va,s_{k+1})\right]\), for \(s\in\gS\). While learning, each agent $i\in {\cal V}$ can share its learning parameter over the communication network with its neighboring agents \(j \in \gN_i\). Following the spirit of single-agent MDP, the aim of each agent is now to compute the solution of the following equation:
\begin{align}
        \mPhi^{\top}\mD^{\pi}\mPhi\vtheta = \mPhi^{\top}\mD^{\pi}\left(\frac{1}{N}\sum_{i=1}^N \mR^{\pi}_i +\gamma \mP^{\pi}\mPhi\vtheta\right),
        \label{eq:distributed_bellman}
\end{align}
where \(\mR^{\pi}_i\in \R^{|\gS|}\) for \(i\in\gV\), whose elements are \([\mR^{\pi}_i]_j=\E\left[r^i(s,\va,s^{\prime})\mid s=j\right]\) for \(j \in \gS\). The equation~(\ref{eq:distributed_bellman}) admits a unique solution \(\vtheta_c\in\R^{q}\), given by
\begin{align}
    \vtheta_c = (\mPhi^{\top}\mD^{\pi}(\mPhi-\gamma \mP^{\pi}\mPhi))^{-1}\mPhi^{\top}\mD^{\pi}\left(\frac{1}{N}\sum_{i=1}^N \mR^{\pi}_i \right). \label{def:vtheta_c}
\end{align}
Note that the solution corresponds to the value function associated with the global reward $\sum^{\infty}_{k=0}\gamma^k \frac{1}{N} \sum_{i=1}^N r^i(s_k,\va_k,s_{k+1})$. Moreover, we will denote, for $1\leq i \leq N$,
\begin{align}
    \mA := & \gamma \mPhi^{\top}\mD^{\pi}\mPhi-\mPhi^{\top}\mD^{\pi}\mP^{\pi}\mPhi,\quad \vb_i:= \mPhi^{\top}\mD^{\pi}\mR^{\pi}_i,\label{def:A,b}
\end{align}
and \( w := \lambda_{\min}(\mPhi^{\top}\mD^{\pi}\mPhi) \). The bound on the reward will be denoted by a positive constant \(R_{\max}\in\R\), i.e., \(|r^i(s,\va,s^{\prime})| \leq R_{\max},\;1\leq i \leq N, \forall{s,\va,s^{\prime}}\in\gS\times \gA \times \gS\).

%% file: ode/ode.tex
The so-called primal-dual gradient dynamics~\citep{arrow1958studies} will be the key tool for the analysis of the proposed distributed TD-learning. The analysis provided in this section will serve as the foundation for the subsequent analysis in Section~\ref{sec:distributed_td}. This section establishes exponential convergent behavior of the primal-dual gradient dynamics in terms of the Lyapunov method. To this end, let us consider the following constrained optimization problem:
\begin{align}
     \min_{\vtheta \in \R^n}  \quad & f( \vtheta) \quad
   \text{such that} \quad  \mM \vtheta = \bm{0}_n  , \label{eq:opt}
\end{align}
where \( \vtheta \in \R^n\), \(\mM \in \R^{n\times n}\) and \(f:\R^n\to\R\) is a differentiable, smooth, and strongly convex function~\citep{boyd2004convex}. One of the popular approaches for solving~(\ref{eq:opt}) is to formulate it into the saddle-point problem~\citep{boyd2004convex}, \(
    L(\vtheta,\vw) = \min_{\vtheta\in\R^n} \max_{\vw\in\R^n} (f(\vtheta)+\vw^{\top}\mM\vtheta),
\) whose solution, \(\vtheta^*,\vw^*\in\R^n\), exists and is unique when \(\mM\) has full-column rank~\citep{qu2018exponential}. If \(\mM\) is rank-deficient, i.e., it is not full-column rank, there exists multiple \(\vw^*\) solving the saddle-point problem. 
It is known that its solution \(\vtheta^*,\vw^*\) can be obtained by investigating the solution \(\vtheta_t,\vw_t\in\R^n\) of the so-called primal-dual gradient dynamics~\citep{qu2018exponential}, with initial points \(\vtheta_0,\vw_0\in\R^n\), 
\begin{align*}
     \dot{\vtheta}_t =&- \nabla f(\vtheta_t) - \mM^{\top}\vw_t,\quad
\dot{\vw}_t = \mM \vtheta_t.
\end{align*}
~\citealp{qu2018exponential} studied exponential stability of the primal-dual gradient dynamics when \(\mM\) is full column-rank, using the classical Lyapunov approach~\citep{sontag2013mathematical}. However, the proof relies on the invertibility of $\mM$, and cannot be extended to the case when \(\mM\) is rank-deficient. As for such case,~\citealp{ozaslan2023global,cisneros2020distributed,gokhale2023contractivity} proved exponential convergence to a particular solution \(\vtheta^*,\vw^*\) using the tools based on singular value decomposition~\citep{horn2012matrix}. In this paper, we will consider the following particular scenarios:
\begin{enumerate}
\item \( \nabla f(\vtheta_t) = \mU \vtheta_t\), where \(\mU\in \R^{n\times n}\), which is not necessarily symmetric but positive definite matrix, i.e., $\mU + \mU^{\top}  \succ 0$;
\item \(\mM\) is symmetric and rank-deficient. Distributed algorithms are typical examples satisfying such condition and will be elaborated in subsequent sections. 
\end{enumerate}
We note that previous works considered general matrix \(\mM\), not necessarily a symmetric matrix. Moreover, note that the primal-dual gradient dynamics under such scenarios will appear in Section~\ref{sec:distributed_td} as an ODE model of the proposed distributed TD-learning. The corresponding system can be rewritten as
\begin{align}
   \frac{d}{dt} \begin{bmatrix}
        \vtheta_t\\
        \vw_t
    \end{bmatrix}
    =\begin{bmatrix}
        -\mU & -\mM^{\top}\\
        \mM & \bm{0}_{n\times n}
    \end{bmatrix}
    \begin{bmatrix}
        \vtheta_t\\
        \vw_t
    \end{bmatrix},\quad \vtheta_0,\vw_0 \in \R^n.\label{eq:primal_dual_ode}
\end{align}  
To study its exponential stability, let us introduce the Lyapunov function candidate \(V(\vtheta,\vw) =   \begin{bmatrix}
        \vtheta\\
        \mM\mM^{\dagger}\vw
    \end{bmatrix}^{\top} \mS  \begin{bmatrix}
        \vtheta\\
        \mM\mM^{\dagger}\vw
    \end{bmatrix}   \), where \(\mS\in \R^{2n\times 2n}\) is some symmetric positive definite matrix, and \(\vtheta,\vw\in\R^n\). The candidate Lyapunov function considers projection of the iterate $\vw_t$ to the range space of $\mM$. As in previous works, the difficulty coming from singularity of \(\mM\) can be avoided by considering the range space and null space conditions of \(\mM\). In particular,~\citealp{ozaslan2023global} employed a Lyapunov function that involves the gradient of the Lagrangian function, and considered the projected iterate \(\mM\mM^{\dagger}\vw_t\), where \(\mM\mM^{\dagger}\) is the projection matrix onto range space of \(\mM\).~\citealp{cisneros2020distributed} exploited a quadratic Lyapunov function in~\citealp{qu2018exponential} for the iterate \(\vtheta_t\) and \(\mV\vw_t\), where \(\mM:=\mT\bm{\Sigma}\mV^{\top}\), which is the singular value decomposition of \(\mM\).~\citealp{gokhale2023contractivity} considered a positive semi-definite matrix $\mS$ and used semi-contraction theory~\citep{de2023dual} to prove exponential convergence of the primal-dual gradient dynamics.
    
    In this paper, we will adopt the quadratic Lyapunov function in~\citealp{qu2018exponential} with the projected iterate \(\mM\mM^{\dagger}\vw_t\), and leverage the symmetric property of \(\mM\) to show improved or comparable to the state of art convergence rate under the particular conditions newly imposed in this paper. In particular, when \(\mM\) is symmetric, the fact that the projection onto the column space of \(\mM\) and row space of \(\mM\) being identical simplifies the overall bounds. We first present the following Lyapunov inequality.
\begin{lemma}\label{lem:ode_lyapunov}
    Let \(\mS:=\begin{bmatrix}
        \beta\mI_{n} & \mM\\
        \mM & \beta \mI_n
    \end{bmatrix}\) where  \(\beta := \max\left\{ \frac{2\lambda_{\max}(\mM)^2 +2+ \left\| \mU \right\|_2^2}{\lambda_{\min}(\mU+\mU^{\top})} ,4\lambda_{\max}(\mM) \right\}\). Then, \(
        \frac{\beta}{2} \mI_{2n} \prec \mS \prec 2\beta \mI_{2n}\), and we have, for any \(\vtheta,\vw \in \R^{n}\),
    \begin{align*}
   & \begin{bmatrix}
        \vtheta\\
        \mM\mM^{\dagger}\vw
    \end{bmatrix}^{\top} \mS   \begin{bmatrix}
        -\mU & -\mM\\
        \mM & \bm{0}_{n\times n}
    \end{bmatrix}       \begin{bmatrix}
        \vtheta\\
        \mM\mM^{\dagger}\vw
    \end{bmatrix} \\
    \leq &  - \min \{1,\lambda^+_{\min}(\mM)^2 \}\left\|\begin{bmatrix}
        \vtheta\\
        \mM\mM^{\dagger}\vw
    \end{bmatrix} \right\|^2_2. \end{align*}
\end{lemma}
The proof is given in \footnote{The Appendix can be found in \url{https://arxiv.org/pdf/2310.00638}}{Appendix Section~\ref{app:proof:lem:ode_lyapunov}}. Using the above Lemma~\ref{lem:ode_lyapunov}, we can now prove the exponential stability of the ODE dynamics in~(\ref{eq:primal_dual_ode}). 
\begin{theorem}\label{thm:ode:bound}
    Let \(V(\vtheta,\vw)=\begin{bmatrix}
    \vtheta \\
    \mM\mM^{\dagger}\vw
\end{bmatrix}^{\top}\mS\begin{bmatrix}
    \vtheta \\
    \mM\mM^{\dagger}\vw
\end{bmatrix}\). For \(\vtheta_0,\vw_0\in\R^n\) and \(t\in\R^+\), we have
\begin{align*}
     & V(\vtheta_t,\vw_t) \\=&   \gO\left(\exp \left(-\frac{ \min \{1,\lambda^+_{\min}(\mM)^2\}  }{\max\left\{ \frac{2\lambda_{\max}(\mM)^2 +2+ \left\| \mU \right\|_2^2}{\lambda_{\min}(\mU+\mU^{\top})} ,4\lambda_{\max}(\mM) \right\}} t \right) \right).
\end{align*}
\end{theorem}   

The proof is given in Appendix Section~\ref{app:thm:ode:bound}. We show that the above bound enjoys sharper or comparable to the state of the art convergence rate under particular conditions. With slight modifications, the Lyapunov function becomes identical to that of~\citealp{gokhale2023contractivity}. However, we directly rely on classical Lyapunov theory~\citep{khalil2015nonlinear} rather than the result from semi-contraction theory~\citep{de2023dual} used in~\citealp{gokhale2023contractivity}. The classical Lyapunov approach simplifies the proof steps compared to that of semi-contraction theory. The detailed comparative analysis is in Appendix Section~\ref{app:sec:compairosn}. The fact that $\mM$ is symmetric and considering the projected iterate \(\mM\mM^{\dagger}\vw_t\), provides improved and comparable bound. Furthermore, as will be clear in Section~\ref{sec:distributed_td}, this enables us to extend the analysis to stochastic algorithms (TD-learning) without introducing involved analysis including (semi)-contraction theory or intricate Lyapunov function.



%% file: dtd/dtd.tex
In this section, we propose a new distributed TD-learning algorithm to solve~(\ref{eq:distributed_bellman}) based on the result in~\citealp{wang2011control}.
In this scenario, each agent keeps its own parameter estimate \(\vtheta^i\in\R^q,\;1\leq i\leq N\), and the goal of each agent is to estimate the value function \(v^{\pi}(s)\approx  \bm{\phi}(s)^{\top}\vtheta_c \) satisfying~(\ref{eq:distributed_bellman}) (the value function associated with the global reward $\sum^{\infty}_{k=0}\gamma^k \frac{1}{N} \sum_{i=1}^N r^i$) under the assumption that each agent has access only to its local reward $r^i$. The parameter of each agent can be shared over the communication network whose structure is represented by the graph \(\gG\), i.e., agents can share their parameters only with their neighbors over the network to solve the global problem. The connections among the agents can be represented by graph Laplacian matrix~\citep{anderson1985eigenvalues}, \(\mL\in \R^{|\gS|\times |\gS|}\), which characterizes the graph \(\gG\), i.e., \([\mL]_{ij}=-1\) if \((i,j)\in\gE\) and \([\mL]_{ij}=0\) if \((i,j)\notin\gE\), and \([\mL]_{ii}= |\gN_i|\) for \(i\in \gV\). Note that \(\mL\) is symmetric positive semi-definite matrix and \(\mL\bm{1}_{|\gS|} = 0\). To proceed, let us first introduce a set of matrix notations:
\begin{align*}
       & \bar{\mL}:=\mL \otimes \mI_q ,\quad \bar{\mD}^{\pi}:=\mI_N\otimes \mD^{\pi} ,\quad \bar{\mP}^{\pi}:=\mI_N\otimes \mP^{\pi}, \\ 
  &  \bar{\mR}^{\pi}=\begin{bmatrix}
        (\mR^{\pi}_1)^{\top} &(\mR^{\pi}_2)^{\top} &\cdots &(\mR^{\pi}_N)^{\top}
    \end{bmatrix}^{\top},\quad \bar{\mPhi}:=\mI_N\otimes \mPhi,\\
    & \bar{\mA}=\mI_{N}\otimes \mA ,\quad \bar{\vb}=\begin{bmatrix}
        \vb_1\\
        \vb_2\\
        \vdots\\
        \vb_N
    \end{bmatrix},\quad \bar{\vtheta}=\begin{bmatrix}
        \vtheta^1\\
        \vtheta^2\\
        \vdots\\
        \vtheta^N
    \end{bmatrix},\quad \bar{\vw}=\begin{bmatrix}
        \vw^1\\
        \vw^2\\
        \vdots\\
        \vw^N
    \end{bmatrix},
\end{align*}
where \(\otimes\) denotes Kronecker product, and \(\bar{\vw}\) is another collection of learnable parameters \(\{\vw^i\in\R^{q}\}_{i=1}^N\), where \(\vw^i\) assigned to each agent \(i\) and $\vb_i$ is defined in~(\ref{def:A,b}).

Meanwhile,~\citealp{wang2011control} studied distributed optimization algorithms~\citep{tsitsiklis1984problems} from the control system perspectives in continuous-time domain, which can be represented as an Lagrangian problem~\citep{hestenes1969multiplier}. Compared to other distributed optimization algorithms~\citep{nedic2009distributed,pu2021distributed}, the method in~\citealp{wang2011control} does not require any specific initialization, diminishing step-sizes, and doubly stochastic matrix that corresponds to the underlying communication graph. Due to these advantages, this framework has been further studied in~\citealp{hatanaka2018passivity,bin2022stability}. 
Inspired by~\citealp{wang2011control},~\citealp{lee2023distributed} developed a continuous-time distributed TD-learning algorithm. The analysis relies on Barbalat's lemma~\citep{khalil2015nonlinear}, which makes extension to the non-asymptotic finite-time analysis difficult for its discrete-time counterpart. Moreover, they focus on the deterministic continuous-time algorithms. The corresponding discrete-time distributed TD-learning is summarized in Algorithm~\ref{algo:1}, where each agent updates its local parameter using the local TD-error in~(\ref{eq:local_td_error}). The updates in~(\ref{eq:theta_update}) and~(\ref{eq:w_update}) in Algorithm~\ref{algo:1} can be obtained by discretizing the continuous-time ODE introduced in~\citealp{wang2011control} with stochastic samples.

\begin{algorithm}
\caption{Distributed TD-learning}\label{algo:1}
\begin{algorithmic}
\STATE Initialize \(\alpha_0\in (0,1),\{\vtheta^i_0,\vw^i_0\in\R^q\}_{i=1}^N, \eta \in (0,\infty)\).
\FOR{$k=1,2,\dots, T$}
\FOR{$i=1,2,\dots,N$}
\STATE  Agent \(i\) observes  \(o^i_k:= (s_k,s_k^{\prime},r^i_k)\).
\STATE Update as follows:
 \begin{align}
        \delta(o^i_k;\vtheta^i_k) =& r^i_k + \gamma \bm{\phi}^{\top}(s_k^{\prime})\vtheta^i_k - \bm{\phi}^{\top}(s_k) \vtheta_k^i\label{eq:local_td_error} \\
        \vtheta^i_{k+1} 
        =& \vtheta^i_{k}  + \alpha_k (\delta(o^i_k;\vtheta^i_k)\bm{\phi}(s_k) \nonumber\\
        &-\eta(|\gN_i| \vtheta^i_k-\textstyle\sum_{j\in \gN_i}\vtheta^j_k ) \nonumber\\
        &-\eta(|\gN_i|\vw^i_k-\textstyle\sum_{j\in \gN_i} \vw^j_k )) \label{eq:theta_update} \\
        \vw^i_{k+1} =& \vw_k^i + \alpha_k \eta ( |\gN_i| \vtheta^i_k-\textstyle\sum_{j\in \gN_i}\vtheta^j_k ) \label{eq:w_update}
\end{align}
\ENDFOR
\ENDFOR
\end{algorithmic}
\end{algorithm}

Using the stacked vector representation, the updates in~(\ref{eq:theta_update}) and~(\ref{eq:w_update}) in Algorithm~\ref{algo:1} can be rewritten in compact form:
\begin{align}
\begin{bmatrix}
    \bar{\vtheta}_{k+1}\\
    \bar{\vw}_{k+1}
\end{bmatrix}
=& \begin{bmatrix}
    \bar{\vtheta}_k\\
    \bar{\vw}_k
\end{bmatrix} 
+\alpha_k 
\begin{bmatrix}
    \bar{\mA}-\eta\bar{\mL} & -\eta\bar{\mL}\\
    \eta\bar{\mL} & \bm{0}
\end{bmatrix}
\begin{bmatrix}
    \bar{\vtheta}_k\\
    \bar{\vw}_k
\end{bmatrix} 
 \nonumber \\
&+\alpha_k\begin{bmatrix}
    \bar{\vb}\\
    \bm{0}
\end{bmatrix}+  \alpha_k \bar{\bm{\eps}}(o_k;\bar{\vtheta}_k) , \label{eq:stacked_form_1}
\end{align}
where, \(o_k:=\{o^i_k\}_{i=1}^N\), and for \(1\leq i \leq N\), 
\begin{align}
   \bm{\eps}^i(o_k^i;\vtheta^i_k):=&\delta(o^i_k;\vtheta^i_k)\bm{\phi}(s_k) - \mA\vtheta_k^i-\vb^i , \nonumber\\
    \bm{\bar{\eps}}(o_k;\bar{\vtheta}_k):=&
    \begin{bmatrix}
    \bm{\eps}^{1\top}_k &
    \bm{\eps}^{2\top}_k &
    \cdots&
    \bm{\eps}^{N\top}_k &
    \bm{0}^{\top}
    \end{bmatrix}^{\top} ,\label{eq:bar_eps}
\end{align}
where we denoted $\bm{\eps}^i_k:=\bm{\eps}^i(o^i_k;\vtheta^i_k)$. Note that the superscript of \(\bm{\eps}^i_k\) corresponds to the $i$-th agent. Compared to the continuous-time algorithm in~\citealp{lee2023distributed}, we introduce an additional positive variable $\eta >0$ multiplied with the graph Laplacian matrix, which results in the factor $\eta$ multiplied with the mixing part in Algorithm~\ref{algo:1} in order to control the variance of the update. 
We note that when the the number of neighbors of an agent $i\in \cal V$ is large, then so is the variance of the corresponding updates of the agent. In this case, the variance can be controlled by adjusting $\eta$ to be small.

The behavior of stochastic algorithm is known to be closely related to its continuous-time O.D.E. counterpart~\citep{borkar2000ode,srikant2019finite}.
In this respect, the corresponding O.D.E. model of~(\ref{eq:stacked_form_1}) is given by
\begin{align}
\frac{d}{dt}
\begin{bmatrix}
        \bar{\vtheta}_t \\
        \bar{\vw}_t
\end{bmatrix}
= \begin{bmatrix}
    \bar{\mA}-\eta\bar{\mL} & -\eta\bar{\mL}\\
    \eta\bar{\mL} & \bm{0}
\end{bmatrix}
\begin{bmatrix}
    \bar{\vtheta}_t\\
    \bar{\vw}_t
\end{bmatrix}+\begin{bmatrix}
    \bar{\vb}\\
    \bm{0}
\end{bmatrix}, \label{eq:algo1:continuous}
\end{align}
for $\bar{\vtheta}_0,\bar{\vw}_0\in\R^{Nq}$,  and \(t\in \R^+\). The above linear system is closely related to the primal-dual gradient dynamics in~(\ref{eq:primal_dual_ode}) in Section~\ref{sec:pd}. Compared to~(\ref{eq:primal_dual_ode}), the difference lies in the fact that the above system corresponds to the the dynamics of the distributed TD-learning represented by matrix $\bar{\mA}$ instead of the gradient of a particular objective function. It is straightforward to check that the equilibrium point of the above system is \(\bm{1}_N\otimes \vtheta_c\) and \( \frac{1}{\eta}\bar{\vw}_{\infty}\) such that \( \bar{\mL}\bar{\vw}_{\infty} = \bar{\mA}(\bm{1}_N\otimes \vtheta_c) + \bar{\vb} \).

In what follows, we will analyze finite-time behavior of~(\ref{eq:stacked_form_1}) based on the Lyapunov equation in Lemma~\ref{lem:algo1:lyapunov_equation_for_projected_iterate}. For the analysis, we will follow the spirit of~\citealp{srikant2019finite}, which studied the standard single-agent TD-learning based on the Lyapunov method~\citep{sontag2013mathematical}. To proceed further, let us consider the coordinate change of \(\tilde{\vtheta}_k:=
    \bar{\vtheta}_k-\bm{1}_N\otimes \vtheta_c\) and \( \tilde{\vw}_k:=\bar{\vw}_k-\frac{1}{\eta}\bar{\vw}_{\infty}\), with which we can rewrite~(\ref{eq:stacked_form_1}) by
\begin{align}
    \begin{bmatrix}
    \tilde{\vtheta}_{k+1}\\
     \tilde{\vw}_{k+1}
\end{bmatrix}
=& \begin{bmatrix}
    \tilde{\vtheta}_{k}\\
     \tilde{\vw}_{k}
\end{bmatrix}
+\alpha_k 
\begin{bmatrix}
    \bar{\mA}-\eta\bar{\mL} & -\eta\bar{\mL}\\
    \eta\bar{\mL} & \bm{0}
\end{bmatrix}
\begin{bmatrix}
    \tilde{\vtheta}_{k}\\
     \tilde{\vw}_{k}
\end{bmatrix} \nonumber \\
&+  \alpha_k \bar{\bm{\eps}}(o_k;\bar{\vtheta}_k) . \label{eq:coordinate_chagne}
\end{align}
We will now derive a Lyapunov inequality for the above system based on the results in Lemma~\ref{lem:algo1:lyapunov_equation_for_projected_iterate}, 
To this end, we will rely on the analysis in~\citealp{qu2018exponential}, which proved exponential convergence of the continuous-time primal-dual gradient dynamics based on the Lyapunov method. However, the newly introduced singularity of \(\bar{\mL}\) imposes difficulty in directly applying the results from~\citealp{qu2018exponential} which does not allow the singularity. To overcome this difficulty, we will multiply \(\bar{\mL}\bar{\mL}^{\dagger}\) to the dual update \(\tilde{\vw}_{k+1}\) in~(\ref{eq:coordinate_chagne}), which is the projection to the range space of \(\bar{\mL}\). The symmetric assumption of \(\bar{\mL}\) helps to construct an explicit solution of the Lyapunov inequality in Lemma~\ref{lem:algo1:lyapunov_equation_for_projected_iterate}. Multiplying \(\bar{\mL}\bar{\mL}^{\dagger}\) to \(\tilde{\vw}_{k+1}\) in~(\ref{eq:coordinate_chagne}) yields
\begin{align}
   \begin{bmatrix}
    \tilde{\vtheta}_{k+1}\\
   \bar{\mL}\bar{\mL}^{\dagger} \tilde{\vw}_{k+1}
\end{bmatrix}
=& \left(\mI_{2N}+\alpha_k 
\begin{bmatrix}
    \bar{\mA}-\eta\bar{\mL} & -\eta\bar{\mL}\\
    \eta\bar{\mL} & \bm{0}
\end{bmatrix}\right)
\begin{bmatrix}
    \tilde{\vtheta}_k\\
   \bar{\mL}\bar{\mL}^{\dagger}  \tilde{\vw}_k
\end{bmatrix} \nonumber \\
&+  \alpha_k \bar{\bm{\eps}}_k(o_k;\bar{\vtheta}_k) , \label{eq:algo1_transformed_update}
\end{align}
which can be proved using Lemma~\ref{lem:moore_penrose} in the Appendix~\ref{subsec:technical}.
For this system, we now derive the following Lyapunov inequality.

\begin{lemma}\label{lem:algo1:lyapunov_equation_for_projected_iterate}
There exists a positive symmetric definite matrix \(\mG \in \R^{2Nq\times 2Nq}\) such that \(  \frac{8+\eta+4\eta^2 \lambda_{\max}(\bar{\mL})^2}{2\eta (1-\gamma)w} \mI_{2Nq}\prec  \mG \prec  2\frac{8+\eta+4\eta^2 \lambda_{\max}(\bar{\mL})^2}{\eta (1-\gamma)w} \mI_{2Nq} \), and  for \(\tilde{\vtheta},\tilde{\vw}\in\R^{Nq}\),
    \begin{align*}
       & 2\begin{bmatrix}
            \tilde{\vtheta}\\
         \bar{\mL}\bar{\mL}^{\dagger}  \tilde{\vw}
        \end{bmatrix}^{\top}
        \mG\begin{bmatrix}
    \bar{\mA}-\eta\bar{\mL} & -\eta\bar{\mL}\\
    \eta\bar{\mL} & \bm{0},\\
\end{bmatrix}        
\begin{bmatrix}
            \tilde{\vtheta}\\
            \bar{\mL}\bar{\mL}^{\dagger}  \tilde{\vw}
        \end{bmatrix}\\
        \leq &  -\min\{1,\eta\lambda_{\min}^+(\bar{\mL})^2 \}\left\|\begin{bmatrix}
        \tilde{\vtheta}\\
        \bar{\mL}\bar{\mL}^{\dagger}\tilde{\vw}
    \end{bmatrix}  \right\|_2^2.
    \end{align*}
\end{lemma}

The proof is given in Appendix Section~\ref{app:proof:lem:lyapunov_equation_for_projected_iterate}. The proof can be done by noting that $\bar{\mA}-\eta\bar{\mL}$ is negative semi-definite and \(\bar{\mL}\) is rank-deficient, and applying Lemma~\ref{lem:ode_lyapunov}.

\subsection{i.i.d. observation case}
We are now in position to provide the first main result, a finite-time analysis of Algorithm~\ref{algo:1} under the i.i.d. observation model, which is a common assumption in the literature, and provides simple and clean theoretical insights.
\begin{theorem}\label{thm:algo1}
\begin{enumerate}
    \item[1.] Suppose we use constant step-size \(\alpha_0=\alpha_1=\cdots=\alpha_k\) for \(k\in\sN_0\), and  \(\alpha_0\leq \bar{\alpha}\) for some positive constant $\bar{\alpha}\in(0,1)$.
    Then, we have
    \begin{align*}
        &\frac{1}{N}\E\left[\left\|
        \begin{bmatrix}
            \tilde{\vtheta}_{k+1} \\
           \bar{\mL}\bar{\mL}^{\dagger}  \tilde{\vw}_{k+1}
        \end{bmatrix}
      \right\|_2^2 \right]\\
      =&\gO \left( \exp\left( -(1-\gamma)w \frac{\min\{1,\eta\lambda^+_{\min}(\bar{\mL})^2\}}{\frac{8}{\eta}+4\eta \lambda_{\max}(\bar{\mL})^2} k\alpha_0\right) \right) \\&+\gO\left(  \alpha_0\frac{R^2_{\max}}{w^3(1-\gamma)^3}  \frac{2+\eta^2\lambda_{\max}(\bar{\mL})^2}{\eta \min\{1,\eta\lambda_{\min}(\bar{\mL})^2\}}   \right).
    \end{align*}
    \item[2.]     Suppose we have \(\alpha_k=\frac{h_1}{k+h_2}\). There exist \(\bar{h}_1\) and \(\bar{h}_2\) such that letting \(h_1=\Theta(\bar{h}_1)\) and \(h_2=\Theta(\bar{h}_2)\) yields
    \begin{align*}
        &\frac{1}{N}\E\left[\left\|
        \begin{bmatrix}
            \tilde{\vtheta}_{k+1} \\
             \bar{\mL}\bar{\mL}^{\dagger}\tilde{\vw}_{k+1}
        \end{bmatrix}
      \right\|_2^2 \right]\\
        =& \gO\left( \frac{1}{k} \frac{(2+\eta^2\lambda_{\max}(\bar{\mL})^2)^2}{\eta^2 \min\{1,\eta\lambda^+_{\min}(\bar{\mL})^2 \}^2} \frac{R^2_{\max}}{w^4(1-\gamma)^4} \right).
    \end{align*}
\end{enumerate}
\end{theorem}

The proof and the exact constants can be found in Appendix Section~\ref{app:sec:proof:algo1_iid}. Using constant step-size, we can guarantee exponential convergence rate with small bias term \(\gO\left( \alpha_0\frac{R^2_{\max}\lambda_{\max}(\bar{\mL})}{w^3(1-\gamma)^3} \right)\) when \(\eta \approx \frac{\sqrt{2}}{\lambda_{\max}(\bar{\mL})}\) and \( \lambda^+_{\min}(\bar{\mL})^2\geq  \sqrt{2}\lambda_{\max}(\bar{\mL})\). Appropriate choice of $\eta$ allows wider range of step-size, and this will be clear in the experimental results in Section~\ref{sec:exp}. Furthermore, the algorithm's performance is closely tied to the properties of the graph structure. $\lambda^+_{\min}(\bar{\mL})$, the smallest non-zero eigenvalue of graph Laplacian, characterizes the connectivity of the graph~\cite{chung1997spectral}, and a graph with lower connectivity will yield slower convergence rate and larger bias. \(\lambda_{\max}(\bar{\mL})\) is the largest eigenvalue of the graph Laplacian, and it can be upper bounded by twice the maximum degree of the graph~\citep{anderson1985eigenvalues}. That is, a graph with higher maximum degree could incur slower convergence rate and larger bias. However, compared to $\lambda_{\min}^+(\mM)$, we experimentally verify in Section~\ref{sec:exp} that $\lambda_{\max}(\bar{\mL})$ does not appear to be an important factor under particular cases, and there could exist a tighter bound without $\lambda_{\max}(\bar{\mL})$. As for diminishing step-size, we achieve \( \gO\left( \frac{1}{k}\right) \) convergence rate from the second item in Theorem~\ref{thm:algo1}, and similar observations hold as in the constant step-size, i.e., the convergence rate depends on the smallest non-zero and maximum eigenvalue of graph Laplacian. Lastly, as in~\citealp{wang2020decentralized}, our bound does not explicitly depend on the number of agents, $N$, compared to the bound in~\citealp{doan2019finite} and~\citealp{sun2020finite}, where the bias term and convergence rate scale at the order of $N$.

 Furthermore, the known constant error bound for (single-agent) TD-learning, which is Theorem 2 of~\citealp{bhandari2018finite} is $O\left(  \frac{1}{(1-\gamma)^4w^2} \right)$. Meanwhile our bound in Theorem 4.2 is $O\left( \frac{1}{(1-\gamma)^3w^3} \right)$ for the constant step-size case. The difference only comes from the choice on the bound in $\theta_c$, the solution of the Bellman equation. We use the bound $\left\|\theta_c\right\|_2\leq O\left(\frac{1}{(1-\gamma)w}\right)$ in Lemma~\ref{theta_c_bound} in Appendix~\ref{subsec:technical}, whereas the bound $O\left(\frac{1}{(1-\gamma)^{\frac{3}{2}}w^{\frac{1}{2}}}\right)$ is used in~\citealp{bhandari2018finite}. 

%% file: dtd/dtd_markov.tex
In this section, we consider the Markovian observation model, where the sequence of observations \(\{s_k\}_{k=1}^{\infty}\) follows a Markov chain. Compared to the i.i.d. observation model, the correlation between the observation and the updated iterates imposes difficulty in the analysis. To overcome this issue, an assumption on the Markov chain that ensures a geometric mixing property is helpful. In particular, the so-called ergodic Markov chain can be characterized by the metric called total variation distance~\citep{levin2017markov}, \(
     d_{\mathrm{TV}}(P,Q)=\frac{1}{2}\sum_{x\in\gS}|P(x)-Q(x)|,
 \)
 where \(P\) and \(Q\) is probability measure on \(\gS\). A Markov chain is said to be ergodic if it is irreducible and aperiodic~\citep{levin2017markov}. An ergodic Markov chain is known to converge to its unique stationary exponentially fast, i.e., for \( k \in \sN_0\), \(
    \sup_{1\leq i \leq |\gS|} d_{\mathrm{TV}}(\ve_i^{\top}(\mP^{\pi})^k,\mu_{\infty}) \leq m \rho^k , 
\)
where \(\ve_i \in \R^{|\gS|}\) for \(1\leq i \leq N \) is the \(|\gS|\)-dimensional vector whose \(i\)-th element is one and others are zero,  \(\mu_\infty \in \R^{|\gS|}\) is the stationary distribution of the Markov chain induced by transition matrix \(\mP^{\pi}\), \(m\in\R\) is a positive constant, and \(\rho \in (0,1)\). The assumption on the geometric mixing property of the Markov chain is common in the literature~\citep{srikant2019finite,wang2020decentralized}. The mixing time of Markov chain is an important quantity of a Markov chain, defined as
\begin{align}
    \tau(\delta):= \min\{ k \in \sN \mid m\rho^k \leq \delta \}.\label{eq:mixing_time}
\end{align}
For simplicity, we will use \(\tau:=\tau(\alpha_T)\), where \(T\in\sN_0\) denotes the total number of iterations, and \(\alpha_k\), is the step-size at $k$-th iteration. If we use the step-size \(\alpha_k=\frac{1}{1+k}\), the mixing time \(\tau\) only contributes to the logarithmic factor, \(\log T \) in the finite-time bound~\citep{bhandari2018finite}. As in the proof of i.i.d. case, using the Lypaunov argument in Lemma~\ref{lem:algo1:lyapunov_equation_for_projected_iterate}, we can prove the finite-time bound on the mean-squared error, following the spirit of~\citealp{srikant2019finite}. To simplify the proof, we will investigate the case \(\eta =1\).
\begin{theorem}\label{thm:markov_td}
    \begin{enumerate}
        \item[1.] Suppose we use constant step-size \(\alpha_0=\alpha_1=\cdots=\alpha_T\) such that $\alpha_0 \leq \bar{\alpha}$ for some positive constant $\bar{\alpha}\in(0,1)$. 
     Then, we have, for \(\tau \leq k \leq T\),
       \begin{align*}
    &\frac{1}{N}\E\left[\left\|
        \begin{bmatrix}
            \tilde{\vtheta}_{k+1} \\
             \bar{\mL}\bar{\mL}^{\dagger}\tilde{\vw}_{k+1}
        \end{bmatrix}
      \right\|_2^2 \right]\\
      =& \gO\left(  \exp\left( -\frac{(1-\gamma)w\min\{1,\lambda_{\min}^+(\mL)^2 \}}{\lambda_{\max}(\mL)^2}\alpha_0( k-\tau)  \right)  \right)\\
       &+ \gO\left(\alpha_0 \tau  \frac{R^2_{\max}}{w^3(1-\gamma)^3}     \frac{\lambda_{\max}(\mL)^2 }{\min\{1,\lambda_{\min}^+(\mL)^2 \}}
    \right) . 
    \end{align*}
    
        \item[2.] Considering diminishing step-size, with \(\alpha_k = \frac{h_1}{k+h_2}\) for \(k \in \sN_0\), there exits \(\bar{h}_1\) and \(\bar{h}_2\) such that for \(h_1 = \Theta( \bar{h}_1 )\) and \(h_2 =\Theta( \bar{h}_2) \),        
        we have for \(\tau \leq k \leq T\),
\begin{align*}
&\frac{1}{N}\E\left[\left\|
        \begin{bmatrix}
            \tilde{\vtheta}_{k+1} \\
             \bar{\mL}\bar{\mL}^{\dagger}\tilde{\vw}_{k+1}
        \end{bmatrix}
      \right\|_2^2 \right] \\
      =&  \gO\left(  \frac{\tau}{k} \frac{qR_{\max}^2}{w^4(1-\gamma)^4} \frac{\lambda_{\max}(\mL)^5}{\min\{1,\lambda_{\min}^+(\mL)^2 \}^2}  \right).
\end{align*}
    \end{enumerate}
\end{theorem}

The proof and the exact values can be found in Appendix~\ref{app:proof:thm_markov_td}. For the constant step-size, we can see that the bounds have additional mixing time factors compared to the i.i.d. case. Considering diminishing step-size, the convergence rate of \( \gO\left(\frac{\tau}{k}\right)\) can be verified, incorporating a multiplication by the mixing time \(\tau\). 

As summarized in Table~\ref{tab:comparison_finite_bound}, the proposed distributed TD-learning does not require doubly stochastic matrix or any specific initializations. The algorithms requiring the doubly stochastic matrix, whose definition is given in Appendix~\ref{sec:dsm}, face challenges when extending to directed graph and time-varying graph scenarios. However, our algorithm does not require major modifications. Meanwhile, push-sum~\citep{nedic2014distributed} or push-pull~\citep{pu2020push} algorithms have been developed to cope with the assumption of doubly stochastic matrix in directed graph scenario. Nonetheless, both methods require knowledge of out-degree, which are often difficult to know in presence including broadcast communications~\citep{hendrickx2015fundamental}. Moreover, the performance of the algorithm is sensitive to the choice of doubly stochastic matrix as can be seen in Appendix~\ref{app:exp:comparison}.
\begin{table*}[h!]
\begin{adjustbox}{width=\textwidth,center}
\centering
 \begin{tabular}{|c| c c c c c|} 
 \hline
  & Method & Observation model  & Step-size & Requirement & Doubly stochastic matrix \\ [0.5ex] 
 \hline\hline
 ~\citealp{doan2019finite} & ~\citealp{nedic2009distributed} & i.i.d. & Constant/$\frac{1}{\sqrt{k+1}}$ & Projection & \checkmark \\
  ~\citealp{doan2021finite} & ~\citealp{nedic2009distributed} & Markovian & Constant/$\frac{h_1}{k+1}$ & \xmark & \checkmark \\
 ~\citealp{sun2020finite} &~\citealp{nedic2009distributed} & i.i.d./Markovian & Constant & \xmark &\checkmark\\
  ~\citealp{zeng2022finite} &~\citealp{nedic2009distributed} & i.i.d./Markovian & Constant & \xmark &\checkmark\\
 ~\citealp{wang2020decentralized} & ~\citealp{pu2021distributed} & i.i.d./Markovian & Constant & Specific initialization &\checkmark \\
 Ours & ~\citealp{wang2011control} & i.i.d./Markovian & Constant/$\frac{h_1}{k+h_2}$ & \xmark & \xmark\\ 
 \hline
 \end{tabular}
 \end{adjustbox}
 \caption{Comparison with existing works.}\label{tab:comparison_finite_bound}
\end{table*}



%% file: experiments/main.tex

\begin{figure*}[!htb]
     \centering
     \begin{subfigure}[t]{0.31\textwidth}
         \centering
         \includegraphics[width=\textwidth]{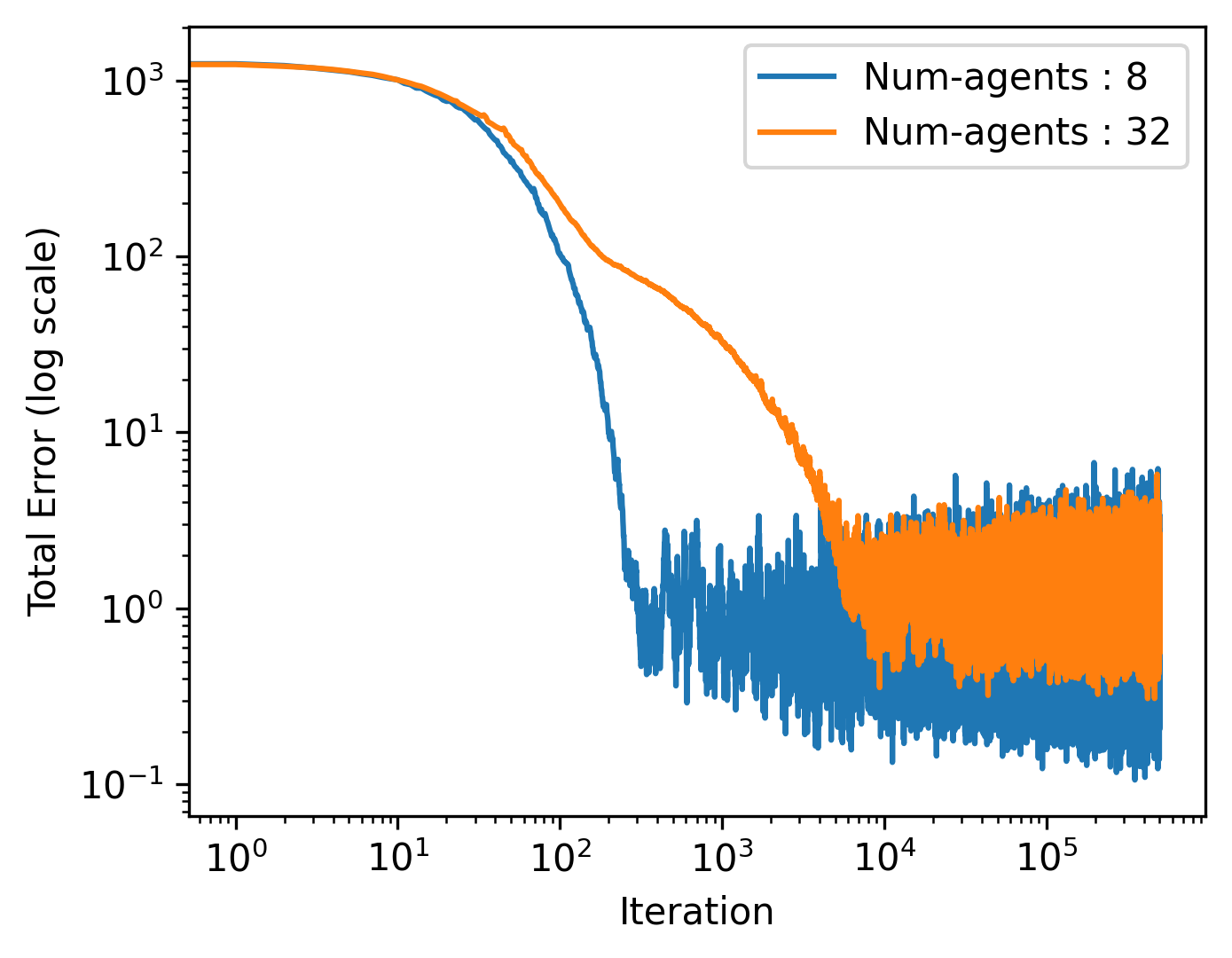}
         \caption{The result shows mean-squared error with $\eta=1$ on cycle graph. }
         \label{fig:step-size}
     \end{subfigure}
     \hfill
          \begin{subfigure}[t]{0.32\textwidth}
         \centering
         \includegraphics[width=\textwidth]{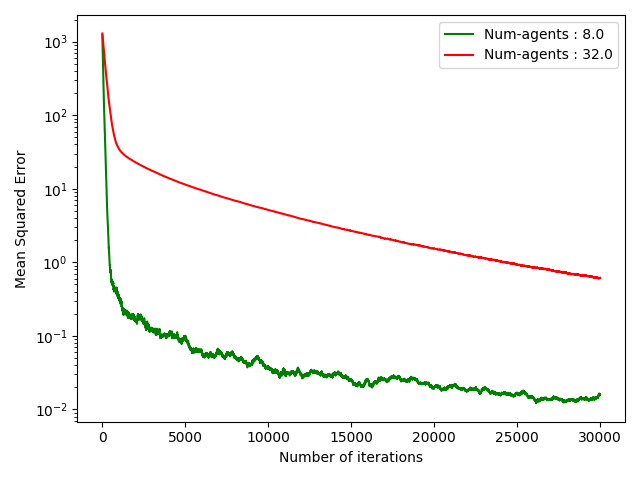}
         \caption{The result shows mean-squared error for the step-size, $\alpha_k=\frac{N^2}{N^3+k}$ on cycle graph. }
         \label{fig:dim-ss-ring}
     \end{subfigure}
     \hfill
     \begin{subfigure}[t]{0.32\textwidth}
         \centering
         \includegraphics[width=\textwidth]{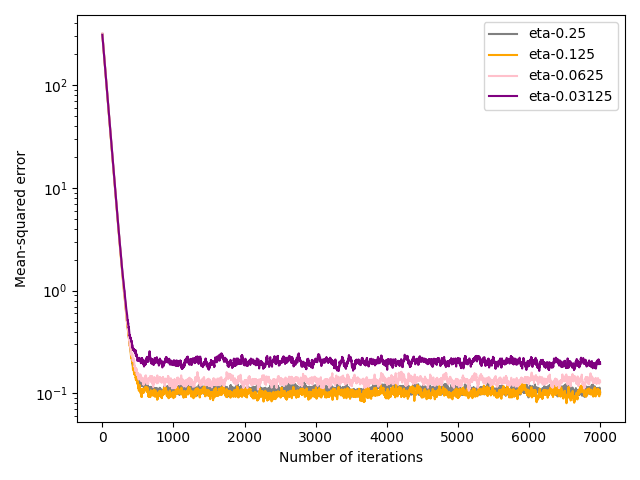}
         \caption{The result shows mean-squared error of the iterates of Algorithm~\ref{algo:1} on random graph with $N=32$, and different values of $\eta$.  }
         \label{fig:eta}
     \end{subfigure}
     \begin{subfigure}[t]{0.32\textwidth}
         \centering
         \includegraphics[width=\textwidth]{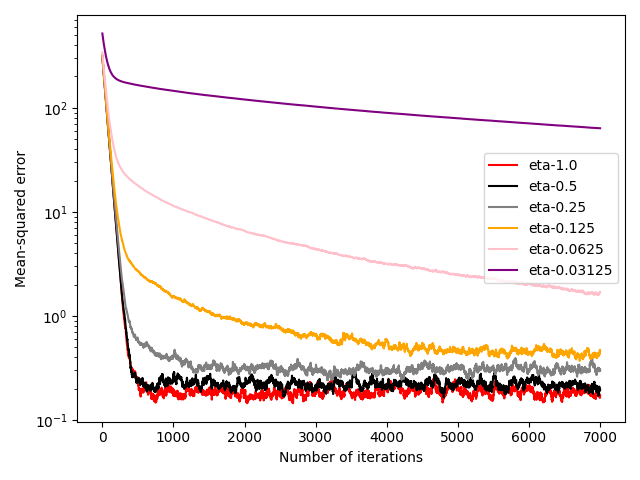}
         \caption{Mean-squared error for Algorithm~\ref{algo:1}, $N=8$, $\alpha=0.1$ on random graph with different values of $\eta$. If $\eta=2$, the algorithm diverges.}
         \label{fig:ramdom-graph-eta-16}
     \end{subfigure}
     \hfill
       \begin{subfigure}[t]{0.32\textwidth}
         \centering
    \includegraphics[width=\textwidth]{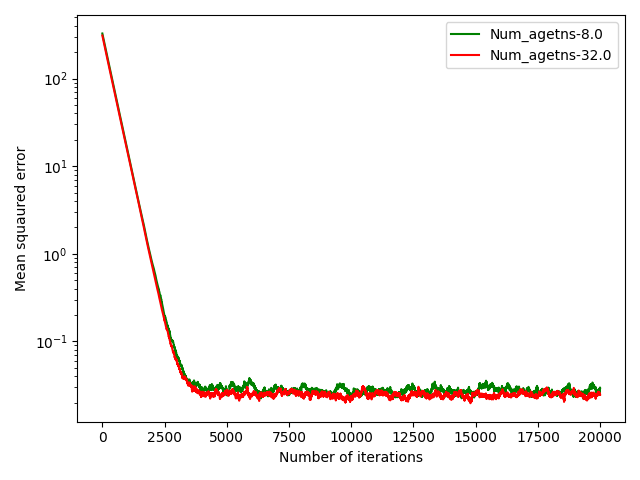}
         \caption{Mean squared error of Algorithm~\ref{algo:1} on star graph after 20,000 iterations. We set $\eta=1$ with step-size $1/2^6$.}
         \label{fig:star-mse-1}
     \end{subfigure}
          \hfill
               \begin{subfigure}[t]{0.32\textwidth}
         \centering
    \includegraphics[width=\textwidth]{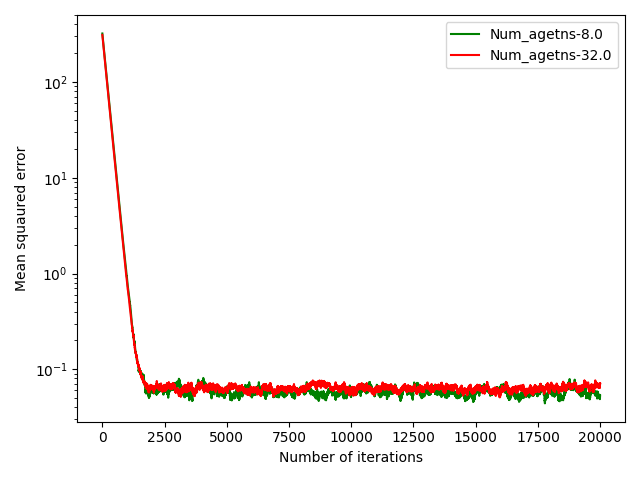}
         \caption{Mean squared error of Algorithm~\ref{algo:1} on star graph after 20,000 iterations. We set $\eta=1$ with step-size $1/2^5$.}
         \label{fig:star-mse-2}
     \end{subfigure}
     \caption{Experiment results of Algorithm~\ref{algo:1}. The experiments were averaged over 50 runs.}
\end{figure*}


\footnote{The code is provided in this \href{https://github.com/limaries30/distributed-TD-learning}{link}.}{This section provides the experimental results of Algorithm~\ref{algo:1}}. First, we give an explanation of the MAMDP setup, where the number of states is three and the dimension of the feature is two. An agent can transit to every state with uniform probability. The feature matrix is set as \(\mPhi^{\top}=\begin{bmatrix}
    1 & 0 & 1\\
    0  &1 & 0
\end{bmatrix}\). The rewards are generated uniformly random between the interval \((0,10)\). The discount factor is set as \(0.8\). 

For each experiment with \(N\in \{2^3,2^5\}\) number of agents, we construct a cycle, a graph \(\gG\) consisting of \(\gV:=\{1,2,\dots,N\}\) and $\gE:=\{(i,i+1)\}_{i=1}^{N-1}\cup\{(N,1)\}$. The smallest non-zero eigenvalue of graph Laplacian corresponding to a cycle with even number of vertices decreases as the number of vertices increases, while maximum eigenvalue remains same. The smallest non-zero eigenvalue is \(2-2\cos\left(\frac{2\pi}{N}\right)\), and the largest eigenvalue is four~\citep{mohar1997some}. As \(N\) gets larger, the smallest non-zero eigenvalue gets smaller, which becomes \(0.59\) and \(0.04\) for \(N=2^3,2^5\), respectively. Therefore, as number of agents increases, the convergence rate will be slower as expected in Theorem~\ref{thm:algo1}, and this can be verified in Figure~(\ref{fig:step-size}) and Figure~(\ref{fig:constant-step-size-ring}) in the Appendix. The plots show the result for constant step-size \(\alpha_0\in\{2^{-3},2^{-4},2^{-5},2^{-6}\}\). Moreover, the convergence under a diminishing step-size can be seen in Figure~(\ref{fig:dim-ss-ring}). To investigate the effect of $\lambda_{\max}(\bar{\mL})$, we construct a star graph, where one vertex has degree $N-1$ and the others have degree one. The maximum eigenvalue of star graph is $N$ and the smallest non-zero eigenvalue is one~\citep{nica2016brief}. Even though $N$ gets larger, we could see in Figure~(\ref{fig:star-mse-1}) and~(\ref{fig:star-mse-2}) that the convergence rate or bias term does not vary. Therefore, we can expect that there could be a tighter bound without $\lambda_{\max}(\bar{\mL})$ under particular cases.

To verify the effect of $\eta$, we use a random graph model~\citep{erdHos1960evolution}, where among possible \(N(N-1)/2\) edges, \((N-3)(N-4)/2\) edges are randomly selected. Figure~(\ref{fig:eta}) shows the evolution of the mean squared error for \(N=32\), and step-size \(0.1\) with different $\eta$ values. When \(\eta = 0.5\) or \(\eta = 1\), the algorithm diverges. Moreover, the bias gets smaller around \(\frac{\sqrt{2}}{\lambda_{\max} (\mL)} \approx 0.046\). This implies that appropriate choice of $\eta$ can control the variance when the number of neighbors is large but if \(\eta\) is too small or large, Algorithm~\ref{algo:1} may cause divergence or large bias. This matches the result of the bound in Theorem~\ref{thm:algo1}. Similar arguments hold when $N=8$, and the result is given in Figure~(\ref{fig:ramdom-graph-eta-16}). 

Lastly, the comparison with other algorithms are given in Appendix~\ref{app:exp:comparison}. In summary, while no single algorithm consistently outperforms the others, the performance of methods that rely on the doubly stochastic matrix is highly sensitive to the choice of this matrix.



%% file: conclusion/main.tex
In this study, we have studied primal-dual gradient dynamics subject to some null-space constraints and its application to a distributed TD-learning. We have derived finite-time error bounds for both the gradient dynamics and the distributed TD-learning. The results have been experimentally demonstrated. Potential future studies include extending the study to finite-time bounds of distributed TD-learning with nonlinear function approximation.

%% file: app_arxiv.tex
\appendix
\onecolumn
\section{Appendix}

\subsection{Notations}
\import{intro}{notations}

\section{Doubly stochastic matrix}\label{sec:dsm}
\import{app}{dsm}

\section{Technical lemmas}\label{subsec:technical}
\import{app}{techincal}

\import{app}{ode}

\import{app}{algo1}

\section{Markovian observation model}\label{app:subsec:markovian}
\import{app}{markov_proof}

\import{app}{exp}

%% file: intro/notations.tex
 \(\R\): set of real numbers; \(\R^+\): set of positive real numbers ; \(\sN\): set of natural numbers;  \(\sN_0\): union of set of natural numbers and element zero; $\mathrm{diag}(\mA_1,\mA_2,\dots,\mA_n)\in\R^{m\times m}$ : block diagonal matrix constructed from \(\mA_1 \in \R^{d_1\times d_1},\mA_2 \in \R^{d_2\times d_2},\dots,\mA_n \in \R^{d_{n}\times d_n}\) where $m=\sum_{i=1}^n d_i$; \(\bm{1}_p\in\R^p\) : \(p\)-dimensional vector whose elements are all one; \(\bm{0}_N\in\R^N\) : \(N\)-dimensional vector whose elements are all zero; \(\bm{0}_{m\times n}\in\R^{m\times n}\) : \(m\times n\)-dimensional matrix whose elements are all zero; \(\mI_n\in\R^{n\times n}\): \(n\times n\)-dimensional identity matrix; \(\mA^{\dagger}\in \R^{n\times n}\): Moore-Penrose inverse of \(\mA\in \R^{n\times n}\); \(\mA \succeq \mB \) for \(\mA,\mB\in\R^{n\times n}\): \(\mA-\mB\) is positive semi-definite matrix; \(\left\|\vx\right\|^2_{\mQ}\) for positive-semi definite matrix \(\mQ\in\R^{n\times n}\) and \(\vx\in \R^{n}\): \(\vx^{\top}\mQ\vx\) ;\([\vv]_i,\; 1\leq i\leq n\) for \(\vv\in \R^n\): \(i\)-th element of \(\vv\); \([\mA]_{ij},\; 1\leq i,j\leq n\) for \(\mA\in \R^{n\times n}\): \(i\)-th row and \(j\)-th column element of \(\mA\); \(\lambda_{\max}(\mA)\) for \(\mA \in \R^{n\times n}\): maximum eigenvalue of \(\mA\); \(\lambda_{\min}(\mA)\) for \(\mA \in \R^{n\times n}\): minimum eigenvalue of \(\mA\); \(\lambda_{\min}^+(\mA)\) for \(\mA\in\R^{n\times n}\): minimum non-zero eigenvalue of \(\mA\); \(\sigma(\gC)\): sigma algebra generated by a family of sets \(\gC\).

%% file: app/dsm.tex
\begin{definition}[Doubly stochastic matrix~\citep{doan2019finite}]
A doubly stochastic matrix \(\mW\in\R^{N\times N}\) is a stochastic matrix of which the row sum and column sum equal one, i.e., $\sum_{i=1}^N[\mW]_{ji}=1$ and $\sum_{i=1}^N[\mW]_{ij}=1$ for $1\leq j \leq N$. A doubly stochastic corresponding to a graph $\gG:=(\gV,\gE)$ requires additional assumption that $[\mW]_{ii}>0$ for $i\in\gV$, and $[\mW]_{uv}=0$ for $(u,v)\notin\gE$. 
\end{definition}



One of the key advantage of our algorithm over other distributed TD algorithms is that we do not require doubly stochastic matrix corresponding to the graph network. We have outlined several reasons highlighting the importance of removing the requirement on doubly stochastic matrix:

To start with, in many real world scenarios, constructing a doubly stochastic matrix is known to be difficult, or even impossible. A typical example is the directed graph scenario. There are graph structures, which do not allow a construction of doubly stochastic matrix~\cite{gharesifard2010does}. However, our algorithm can be extended to the directed graph setting without major modifications. For example, as shown in~\citealp{gokhale2023contractivity,kia2015distributed}, only with appropriate constant multiplication, we can guarantee convergence under strongly-connected and weight-balanced digraph scenario. Meanwhile, in the distributed optimization literature, push-sum~\citep{nedic2014distributed} or push-pull~\citep{pu2020push} algorithms have been developed to cope with the assumption of mixing matrix in directed graph scenario. Nonetheless, both methods require knowledge of out-degree, which are sometimes not possible including the broadcast communication setting~\citep{hendrickx2015fundamental}. 

Moreover, when dealing with a time-varying graph, whenever the graph changes, the doubly stochastic matrix needs to be constructed again. However, our analysis can be easily extended to the time-varying graph setting without any modifications.

Lastly, as from our experiment, the performance of distributed TD algorithms using doubly stochastic matrix is quite sensitive to the choice of doubly stochastic matrix, and the results can be found in Appendix~\ref{app:exp:comparison}.






%% file: app/techincal.tex
\begin{lemma}[~\citealp{pavlikova2023moore}, p. 2]\label{lem:moore_penrose}
    For real symmetric matrix \(\mA \in \R^n\), and its Moore-Penrose pseudo inverse \(\mA^{\dagger}\), the following holds:
    \begin{align*}
    \mA\mA^{\dagger}=\mA^{\dagger}\mA,\quad \mA\mA^{\dagger}\mA = \mA.
    \end{align*}
\end{lemma}

\begin{lemma}[Schur complement and symmetric positive definite matrices, Theorem 1.12 in~\citealp{horn2005basic}]\label{lem:schur_complement}
Let \(\mH \in \R^{(n+m)\times (n+m)}\) be a symmetric matrix partitioned as
\begin{align*}
    \mH:=\begin{bmatrix}
        \mH_{11} & \mH_{12}\\
        \mH_{12}^{\top} & \mH_{22}
    \end{bmatrix},
\end{align*}
where \(\mH_{11}\in \R^{n\times n},\mH_{12}\in\R^{n\times m},\mH_{22}\in\R^{m\times n}\). Then, the following holds:
\begin{align*}
    \mH \succ 0 \iff \mH_{11}\succ 0,\;\mathrm{and}\;\mH_{22}-\mH_{12}^{\top}\mH_{11}^{-1}\mH_{12} \succ 0.
\end{align*}
\end{lemma}

\begin{lemma}[Proposition 4.5 in~\citealp{levin2017markov}]\label{lem:tv_expression_f}
Let \(\mu\) and \(\nu\) be two probability distributions on \(\gX\). For \(f:\gX\to\R\), the total variation distance can be represented as
\begin{align*}
    d_{\mathrm{TV}}(\mu,\nu):=\frac{1}{2}\sup_{f:\sup_{x\in\gX} |f(x)|\leq 1}\left| \sum_{x\in \gX}f(x)\mu(x)-f(x)\nu(x)\right|.
\end{align*}
\end{lemma}

\begin{lemma}\label{lem:geometric_mixing}
    Consider the Markov chain in Section~\ref{sec:sa_markov}. Let \(Y:=(s_{k+\tau},s_{k+\tau+1})\) for \(k, \tau\in\sN_0\), and \((s_{k+\tau},s_{k+\tau+1})\in\gS \times \gS\). For bounded function \(f:\gS\times\gS \to \R\), i.e., \(\sup_{x\in \gS\times\gS}|f(x)|<\infty\) , we have
    \begin{align*}
       | \E[f(Y)\mid s_k]-\E[f(Y)]| \leq 2 \sup_{x\in \gS\times \gS} |f(x)| m \rho^{\tau}.
    \end{align*}
    Moreover, for \(\vv:\gS\times\gS\to\R^{Nq}\), whose elements are bounded, we have
    \begin{align*}
        \left\| \E[\vv(Y)\mid s_k]-\E[\vv(Y)]\right\|_2 \leq 2\sqrt{Nq} \sup_{x\in\gS\times\gS}\left\| \vv(x)\right\|_{\infty} m \rho^{\tau}.
    \end{align*}
    For \(\mM: \gS\times\gS\to \R^{Nq\times Nq}\), whose elements are bounded, we have
    \begin{align*}
        \left\|\E[\mM(Y)\mid s_k]-\E[\mM(Y) \right\|_2 \leq 2Nq \sup_{x\in\gS\times\gS} \max_{1\leq i,j \leq Nq}|[\mM(x)]_{ij} |m\rho^{\tau}.
    \end{align*} 
\end{lemma}
\begin{proof}
    Let the probability measure \(P(Y\in\cdot)=\sP[Y\in\cdot \mid s_k]\) and \(Q(Y\in \cdot)=\sP[Y\in \cdot]\). For simplicity of the proof, let \(f_{\infty}:=2\sup_{x\in \gS\times \gS} |f(x)|\). Then, we have
    \begin{align*}
         &| \E[f(Y)\mid s_k]-\E[f(Y)]| \\
        =& \left| \int f(Y)dP-\int f(Y)dQ \right|\\
        = & 2f_{\infty}\left|\int \frac{f}{2f_{\infty}}dP-\int \frac{f}{2f_{\infty}}dQ\right|\\
        \leq & 2f_{\infty} d_{\mathrm{TV}} (\sP[Y\in \cdot \mid s_k], \sP[Y\in\cdot])\\
        =&   f_{\infty} \sum_{s,s^{\prime}\in\gS\times\gS}| \sP[s_{k+\tau}=s,s_{k+\tau+1}=s^{\prime}\mid s_k]-\sP[s_{k+\tau}=s,s_{k+\tau+1}=s^{\prime}]|\\
        =& f_{\infty} \sum_{s,s^{\prime}\in\gS\times\gS} |\sP[s_{k+\tau+1}=s^{\prime}\mid s_k,s_{k+\tau}=s]\sP[s_{k+\tau}=s\mid s_k]-\sP[s_{k+\tau+1}=s^{\prime}\mid s_{k+\tau}=s]\sP[s_{k+\tau}=s]|\\
        =&f_{\infty} \sum_{s^{\prime}\in\gS}\sum_{s\in\gS} |\sP[s_{k+\tau+1}=s^{\prime}\mid s_{k+\tau}=s]\sP[s_{k+\tau}=s\mid s_k]-\sP[s_{k+\tau+1}=s^{\prime}\mid s_{k+\tau}=s]\sP[s_{k+\tau}=s]|\\
        \leq & f_{\infty}
        \sum_{s^{\prime}\in\gS}\sum_{s\in\gS} |\sP[s_{k+\tau+1}=s^{\prime}\mid s_{k+\tau}=s]||\sP[s_{k+\tau}=s\mid s_k]-\sP[s_{k+\tau}=s]|\\
        =&f_{\infty}\sum_{s\in\gS} |\sP[s_{k+\tau}=s\mid s_k]-\sP[s_{k+\tau}=s]|\sum_{s^{\prime}\in\gS}|\sP[s_{k+\tau+1}=s^{\prime}\mid s_{k+\tau}=s]|\\
        =& 2f_{\infty}d_{\mathrm{TV}}(\sP[s_{k+\tau}=s\mid s_k],\sP[s_{k+\tau}=s]).
    \end{align*}
    The first inequality follows from the definition of total variation distance in Lemma~\ref{lem:tv_expression_f}. The last equality follows from the fact that \(\sum_{s^{\prime}\in\gS}|\sP[s_{k+\tau+1}=s^{\prime}\mid s_{k+\tau}=s]|=1\). We obtain the desired result from the ergodicity of the Markov chain.

    For the second item, we have
    \begin{align*}
        \left\|\E[\vv(Y)\mid s_k]-\E[\vv(Y)]\right\|_2=\sqrt{\sum_{i=1}^{Nq}(\E[\vv_i(Y)\mid s_k]-\E[\vv_i(Y)])^2 },
    \end{align*}
    where \(\vv_i\) denotes the \(i\)-th element of \(\vv\). The rest of the proof follows as in the proof of first item.

    For the third item, we have

    \begin{align*}
        \left\|\E\left[ \mM(Y) \mid s_k\right] -\E\left[\mM(Y)\right]\right\|_2 \leq & \left\|\E\left[ \mM(Y) \mid s_k\right] -\E\left[\mM(Y)\right]\right\|_F\\
        =& \sqrt{\sum_{i=1}^{Nq}\sum_{j=1}^{Nq}(\E[\mM(Y)]_{ij}\mid s_k]-\E[\mM(Y)]_{ij})^2},
    \end{align*}
    where \( \left\| \mB \right\|_F = \sqrt{\sum_{i=1}^n\sum_{j=1}^n [\mB]_{ij}^2} \) for \( \mB \in \R^{n \times n }\). The rest of the proof follows as in the proof of first item.
\end{proof}

The following lemma provides similar bound as in Lemma 7 in~\citealp{bhandari2018finite}:
\begin{lemma}\label{theta_c_bound}
Consider \(\vtheta_c\) in~(\ref{def:vtheta_c}). We have
\begin{align*}
    \left\|\vtheta_c \right\|_2\leq \frac{R_{\max}}{(1-\gamma)w} ,
\end{align*}
where \( w = \lambda_{\min} ( \mPhi^{\top}\mD^{\pi}\mPhi\)).
\end{lemma}
\begin{proof}
From~(\ref{def:vtheta_c}), \(\vtheta_c\) satisfies
    \begin{align*}
        \mPhi^{\top}\mD^{\pi}\mPhi \vtheta_c -\gamma\mPhi^{\top}\mD^{\pi}\mP^{\pi}\mPhi\vtheta_c = \frac{1}{N}\sum_{i=1}^N\vb_i,
    \end{align*} 
 where $\vb_i$ is defined in~(\ref{def:A,b}).
    Multiplying  \(\vtheta_c\) on both sides of the equations, we have
    \begin{align*}
        \vtheta_c^{\top}(\mPhi^{\top}\mD^{\pi}\mPhi-\gamma\mPhi^{\top}\mD^{\pi}\mP^{\pi}\mPhi)\vtheta_c =& \vtheta_c^{\top}\left(\frac{1}{N}\sum_{i=1}^N\vb_i \right)\\
                        \leq & \left\|\vtheta_c\right\|_2 R_{\max},
    \end{align*}
    where the inequality follows from Cauchy-Schwartz inequality. From Lemma~\ref{lem:A_bound} in the Appendix Section~\ref{subsec:technical}, we have
    \begin{align*}
      (-\mA-\mA^{\top}) \succeq 2(1-\gamma)\mPhi^{\top}\mD^{\pi}\mPhi,
    \end{align*}
    which leads to 
    \begin{align*}
        (1-\gamma) w \left\|\vtheta_c\right\|_2^2\leq  \left\|\vtheta_c\right\|_2 R_{\max}.
    \end{align*}
    Therefore, we have
    \begin{align*}
        \left\|  \vtheta_c \right\|_2 \leq \frac{R_{\max}}{(1-\gamma)w} .
    \end{align*}
\end{proof}

The negative definiteness of \(\mA\) and upper bound on norm of \(\mA\) are established in the following lemma, which resembles that of Lemma 3 and 4 in~\citealp{bhandari2018finite}:
\begin{lemma}\label{lem:A_bound}
We have
    \begin{align*}
        &\mA^{\top}+\mA \preceq 2(\gamma-1) \mPhi^{\top}\mD^{\pi}\mPhi,\quad \left\| \mA \right\|_2 \leq 2.
    \end{align*}
\end{lemma}
\begin{proof}

We will first prove the negative definiteness of \(\mA\). For any \(\vv\in \R^{|\gS|}\), we have
\begin{align*}
    \left\|\mP^{\pi} \vv \right\|_{\mD^{\pi}} =& \sqrt{\sum_{i=1}^{|\gS|} d(i) \left(\sum_{j=1}^{|\gS|} \gP^{\pi}(i,j)[\vv]_j \right)^2 }\\
    \leq & \sqrt{\sum^{|\gS|}_{i=1} d(i)\sum_{j=1}^{|\gS|} \gP^{\pi}(i,j) [\vv]_j^2}\\
    =& \sqrt{\sum_{j=1}^{|\gS|}[\vv]_j^2 \sum_{i=1}^{|\gS|} d(i)\gP^{\pi}(i,j)  }\\
    =& \sqrt{ \sum^{|\gS|}_{j=1} [\vv]_j^2 d(j)  }\\
    =& \left\|\vv\right\|_{\mD^{\pi}},
\end{align*}
where the first inequality follow from Jensen's inequality and the second last equality follows from the fact that \(d(s),s\in\gS\) is the stationary distribution of Markov chain induced by \(\gP^{\pi}\). Therefore, we get 
\begin{align*}
    \vv^{\top}\mA \vv =& \gamma \vv^{\top}\mPhi^{\top}\mD^{\pi}\mP^{\pi}\mPhi\vv-\vv^{\top}\mPhi^{\top}\mD^{\pi}\mPhi\vv\\
                    \leq & \gamma \left\| \mPhi\vv\right\|_{\mD^{\pi}}\left\| \mP^{\pi} \mPhi\vv \right\|_{\mD^{\pi}}-\vv^{\top}\mPhi^{\top}\mD^{\pi}\mPhi\vv\\
                    \leq & \gamma \left\| \mPhi\vv \right\|^2_{\mD^{\pi}} -\left\|\mPhi\vv\right\|_{\mD^{\pi}}^2\\
                    =& (\gamma-1) \vv^{\top}\mPhi^{\top}\mD^{\pi}\mPhi\vv. 
\end{align*}

Now, we will prove the upper bound on \(\left\|\mA\right\|_2\). First, note that the following holds:
\begin{align*}
    \left\|\mPhi^{\top}\mD^{\pi}\mPhi\right\|_2 =& \left\| \sum_{i=1}^{|\gS|} d(i)\bm{\phi}(i)\bm{\phi}(i)^{\top}\right\|_2 \\
    \leq & \sum^{|\gS|}_{i=1} d(i) \left\| \bm{\phi}(i)\right\|_2^2\\
    \leq & \sum^{|\gS|}_{i=1} d(i)\\
    =&1,
\end{align*}
where the first inequality follows from triangle inequality, and the second inequality follows from the assumption that \(\left\| \bm{\phi}(s) \right\|_2 \leq 1\) for \(s\in\gS\). Now, we have



    \begin{align*}
       \left\| \mA  \right\|_2 =& \left\|\sum_{s\in\gS} d(s) \bm{\phi}(s)\left(-\bm{\phi}(s)^{\top}+ \gamma \sum_{s^{\prime}\in\gS}  \gP^{\pi}(s,s^{\prime})  \bm{\phi}(s^{\prime})^{\top}\right) \right\|_2\\
       \leq & \left\| \sum_{s\in\gS} d(s) \bm{\phi}(s)\bm{\phi}(s)^{\top}\right\|_2 +\gamma \left\| \sum_{s\in\gS} d(s)\sum_{s^{\prime}\in\gS} \gP^{\pi}(s,s^{\prime}) \bm{\phi}(s)\bm{\phi}(s^{\prime})^{\top}\right\|_2 \\
       \leq & \sum_{s\in \gS} d(s) + \gamma \sum_{s\in \gS} d(s) \sum_{s^{\prime}\gS}  \gP^{\pi}(s,s^{\prime}) \\
       \leq & 2.
    \end{align*}
    The first inequality follows from triangle inequality. Then second inequality follows from the assumption that \( \left\|\bm{\phi}(s) \right\|_2\leq 1\) for \(s\in\gS\).
\end{proof}

\begin{lemma}\label{lem:vb_bound}
For \(1\leq i \leq N\), consider \(\vb_i\) in~(\ref{def:A,b}). We have
 \begin{align*}
     \left\| \vb_i \right\|_2 \leq R_{\max}.
\end{align*}
\end{lemma}
\begin{proof}
    For \(1\leq i \leq N\), we have
    \begin{align*}
         \left\|\vb_i \right\|_2 =& \left\| \sum_{s\in\gS }\bm{\phi}(s) d^{\pi}(s) [\mR^{\pi}_i]_s \right\|_2 \\
                    \leq & \sum_{s\in\gS} d^{\pi}(s) R_{\max}\\
                    =& R_{\max},
    \end{align*}
where the first inequality follows from \(\left\|\bm{\phi}(s)\right\|_2 \leq 1 \) for \(s \in \gS\) and boundedness on the reward.
\end{proof}


\begin{lemma}\label{lem:algo1:saddle_point_matrix_norm}
We have
    \begin{align*}
        \left\|\begin{bmatrix}
    \bar{\mA}-\bar{\mL} & -\bar{\mL}\\
    \bar{\mL} & \bm{0}_{Nq\times Nq}
\end{bmatrix}    \right\|_2 \leq 2+2\lambda_{\max}(\bar{\mL}).
    \end{align*}
\end{lemma}
\begin{proof}

Applying triangle inequality, we have
\begin{align*}
    \left\|\begin{bmatrix}
    \bar{\mA}-\bar{\mL} & -\bar{\mL}\\
    \bar{\mL} & \bm{0}_{Nq\times Nq}
\end{bmatrix}    \right\|_2 =& \left\|\begin{bmatrix}
    \bar{\mA}-\bar{\mL} & \bm{0}_{Nq\times Nq}\\
    \bm{0}_{Nq\times Nq} & \bm{0}_{Nq\times Nq}
\end{bmatrix}+\begin{bmatrix}
    \bm{0}_{Nq\times Nq} & -\bar{\mL}\\
    \bar{\mL} & \bm{0}_{Nq\times Nq}
\end{bmatrix}    \right\|_2 \\
\leq & \left\|\bar{\mA}-\bar{\mL}\right\|_2 + \left\| \bar{\mL}\right\|_2\\
\leq & 2+2\lambda_{\max}(\bar{\mL}).
\end{align*}
The last inequality follows again from triangle inequality and Lemma~\ref{lem:A_bound}.

\end{proof}

\begin{lemma}\label{lem:bar_eps_theta_bound}
For \(k\in\sN_0\), consider a sequence of observations \(\{ o_i \}_{i=1}^k\). Then, we have
    \begin{align}
    \left\| \bar{\bm{\eps}}(o_k;\bar{\vtheta}_k ) \right\|_2  \leq  6\left\|\tilde{\vtheta}_k \right\|_2+\frac{9\sqrt{N}R_{\max}}{(1-\gamma)w} . \nonumber 
    \end{align}
In particular, if \(\{o_i\}_{i=1}^k\) is sampled from i.i.d. distribution, we have
    \begin{align}
        \E\left[\left\|\bm{\bar{\eps}}(o_k;\bar{\vtheta}_k)\right\|_2^2\right] \leq 16\left\|\tilde{\vtheta}_k\right\|^2+\frac{32NR_{\max}^2}{w^2(1-\gamma)^2}  .\label{ienq:eps_iid_bound}
    \end{align}
\end{lemma}
\begin{proof}

First, consider that for \(1 \leq i \leq N \), we have
\begin{align}
    &\left\|\bm{\eps}^i(o^i_k;\vtheta^i_k) \right\|_2^2 \nonumber\\
    =& \left\| (r^i_k + \gamma \bm{\phi}^{\top}(s_k^{\prime})\vtheta^i_k - \bm{\phi}^{\top}(s_k) \vtheta_k^i) \bm{\phi}(s_k)- \mA\vtheta_k^i-\vb_i\right\|_2^2 \nonumber\\
    \leq & 2 \left\|(r^i_k + \gamma \bm{\phi}^{\top}(s_k^{\prime})\vtheta^i_k - \bm{\phi}^{\top}(s_k) \vtheta_k^i) \bm{\phi}(s_k) \right\|_2^2 + 2 \left\|\mA \vtheta^i_k+\vb_i  \right\|_2^2 \nonumber\\
    \leq & 4 \left\| r^i_k \bm{\phi}(s_k) \right\|_2^2+4 \left\|(\gamma \bm{\phi}^{\top}(s_k^{\prime})\vtheta^i_k - \bm{\phi}^{\top}(s_k) \vtheta_k^i) \bm{\phi}(s_k) \right\|_2^2+4 \sigma_{\max}(\mA)^2\left\| \vtheta^i_k\right\|^2_2 + 4R_{\max}^2 \nonumber\\
    \leq & \left(4 \sigma_{\max}(\mA)^2+16\right)\left\| \vtheta^i_k\right\|^2_2  +  8R_{\max}^2, \label{ineq:eps_i_bound}
\end{align}
where \(\bm{\eps}^i(o^i_k;\vtheta^i_k)\) is defined in~(\ref{eq:bar_eps}). The second inequality follows from Lemma~\ref{lem:vb_bound}. The last inequality follows from the assumption that \(\left\|\bm{\phi}(s)\right\|_2 \leq 1\) for \(s \in \gS\) in Assumption~\ref{assmp:feature}, and \(\left\|\va+\vb\right\|^2_2\leq 2\left\|\va\right\|^2_2+2\left\|\vb\right\|^2_2 \) for \(\va,\vb\in\R^{Nq}\). 

Now, we have
\begin{align}
\left\| \bar{\bm{\eps}}(o_k;\bar{\vtheta}_k ) \right\|_2 =& 
    \left\|\begin{bmatrix}
    \bm{\eps}^1(o^1_k;\vtheta^1_k)\\
    \bm{\eps}^2(o^2_k;\vtheta^2_k)\\
    \vdots\\
    \bm{\eps}^N(o^N_k;\vtheta^N_k)\\
    \bm{0}_{Nq}
    \end{bmatrix} \right\|_2 \nonumber \\
    = & \sqrt{\sum_{i=1}^N\left\| \bm{\eps}^i(o^i_k;\vtheta^i_k) \right\|^2_2}\nonumber \\
    \leq & \sqrt{\sum^N_{i=1} \left(4 \sigma_{\max}(\mA)^2+16\right)\left\| \vtheta^i_k\right\|^2_2  +  8R_{\max}^2 } \nonumber\\
    \leq & \sqrt{\left(4 \sigma_{\max}(\mA)^2+16\right)}\sqrt{\sum^N_{i=1} \left\| \vtheta^i_k\right\|^2_2}+\sqrt{8N R_{\max}^2} \nonumber \\
    \leq &  6 \left\|  \bar{\vtheta}_k \right\|_2+ 3\sqrt{N} R_{\max} \label{ienq:eps_pre}\\
    \leq & 6\left\|\tilde{\vtheta}_k \right\|_2+6\left\|\bm{1}_N\otimes \vtheta_c\right\|_2+3\sqrt{N} R_{\max} \nonumber \\
    \leq & 6\left\|\tilde{\vtheta}_k \right\|_2+6 \sqrt{N}\frac{R_{\max}}{(1-\gamma)w}+3\sqrt{N}R_{\max} \nonumber \\
    \leq & 6\left\|\tilde{\vtheta}_k \right\|_2+\frac{9\sqrt{N}R_{\max}}{(1-\gamma)w} \nonumber.
\end{align}
The second equality follows from the definition of Euclidean norm. The first inequality follows from~(\ref{ineq:eps_i_bound}).  The third inequality follows from bound on \(\sigma_{\max}(\mA)\) in Lemma~\ref{lem:A_bound}. The fourth inequality follows from triangle inequality. The second last inequality follows from Lemma~\ref{theta_c_bound}. This proves the first statement.

We will now prove the inequality~(\ref{ienq:eps_iid_bound}). For simplicity of the proof, let 
\begin{align*}
    \bar{\bm{\delta}}(o_k;\bar{\vtheta}_k) := \begin{bmatrix}
        \delta(o^1_k;\vtheta^1_k) \bm{\phi}(s_k)\\
        \delta(o^2_k;\vtheta^2_k) \bm{\phi}(s_k)\\
        \vdots\\
        \delta(o^N_k;\vtheta^N_k) \bm{\phi}(s_k)
    \end{bmatrix} \in \R^{Nq},
\end{align*}

where \(\delta(o^i_k;\vtheta^i_k) , \; 1 \leq i \leq N \) is defined in ~(\ref{eq:local_td_error}). Since \(\E\left[\bar{\bm{\delta}}(o_k;\bar{\vtheta}_k) \middle|\gF_{k-1}\right]=\bar{\mA}\bar{\vtheta}_k+\bar{\vb}\), we have

\begin{align}
    &\E\left[\left\|\bm{\bar{\eps}}(o_k;\bar{\vtheta}_k)\right\|_2^2 \middle |\gF_{k-1}\right] \nonumber \\
    =& \E\left[ \left\|\begin{bmatrix}
        \bar{\bm{\delta}}(o_k;\bar{\vtheta}_k) \\
        \bm{0}_{Nq}
    \end{bmatrix} -\begin{bmatrix}
        \bar{\mA}\bar{\vtheta}_k+\bar{\vb}\\
        \bm{0}_{Nq}
    \end{bmatrix}\right\|_2^2 \middle|\gF_{k-1}\right] \nonumber \\
    =& \E\left[\left\|\bar{\bm{\delta}}(o_k;\bar{\vtheta}_k)\right\|_2^2\middle | \gF_{k-1}\right]-2\E\left[\begin{bmatrix}
        \bar{\bm{\delta}}(o_k;\bar{\vtheta}_k)\\
        \bm{0}_{Nq}
    \end{bmatrix}^{\top} \middle|\gF_{k-1} \right]\begin{bmatrix}
        \bar{\mA}\bar{\vtheta}_k+\bar{\vb}\\
        \bm{0}_{Nq}
    \end{bmatrix}+\E\left[\left\|\begin{bmatrix}
        \bar{\mA}\bar{\vtheta}_k+\bar{\vb}\\
        \bm{0}_{Nq}
    \end{bmatrix}\right\|_2^2\right] \nonumber \\
    =&\E\left[\left\|\bar{\bm{\delta}}(o_k;\bar{\vtheta}_k)\right\|_2^2\middle | \gF_{k-1}\right] -\E\left[\left\|\begin{bmatrix}
        \bar{\mA}\bar{\vtheta}_k+\bar{\vb}\\
        \bm{0}_{Nq}
    \end{bmatrix}\right\|_2^2\right] \nonumber\\
    \leq & \E\left[\left\|\bar{\bm{\delta}}(o_k;\bar{\vtheta}_k)\right\|_2^2\middle | \gF_{k-1}\right].\nonumber
\end{align}

Taking total expectation, we get

\begin{align*}
  \E\left[\left\|\bm{\bar{\eps}}(o_k;\bar{\vtheta}_k)\right\|_2^2\right] \leq &  \E\left[\left\|\bar{\bm{\delta}}(o_k;\bar{\vtheta}_k)\right\|^2_2\right]\\ = & \E\left[\sum_{i=1}^N \left\|\delta(o_k;\vtheta^i_k)\bm{\phi}(s_k)\right\|_2^2\right]\\
    = & \E\left[\sum_{i=1}^N \left\| (r^i_k + \gamma \bm{\phi}^{\top}(s_k^{\prime})\vtheta^i_k - \bm{\phi}^{\top}(s_k) \vtheta_k^i) \bm{\phi}(s_k) \right\|_2^2 \right]\\
    \leq & \E\left[ \sum_{i=1}^N\left( 2\left\|r^i_k\bm{\phi}(s_k)\right\|_2^2+2\left\|\gamma\bm{\phi}(s_k) \bm{\phi}^{\top}(s_k)-\bm{\phi}(s_k) \bm{\phi}(s_k)^{\top} \right\|_2^2\left\|\vtheta^i_k\right\|_2^2 \right)\right]\\
    \leq & \E\left[ 2\sum^N_{i=1}\left(R_{\max}^2+4 \left\|\vtheta_k^i\right\|^2_2  \right)\right]\\
    =& 2NR_{\max}^2+ 8 \left\|\bar{\vtheta}_k\right\|^2_2.
\end{align*}

The second last inequality follows from the fact that \(\left\|\va+\vb\right\|^2_2\leq 2\left\|\va\right\|^2_2+2\left\|\vb\right\|^2_2 \) for \(\va,\vb\in\R^{Nq}\).  The last inequality follows from the assumption that \(\left\|\bm{\phi}(s)\right\|_2 \leq 1\) for \(s \in \gS\) in Assumption~\ref{assmp:feature}. Using triangle inequality, we get

\begin{align*}
     \E\left[\left\|\bm{\bar{\eps}}(o_k;\bar{\vtheta}_k)\right\|_2^2\right] \leq & 2NR^2_{\max} + 8 \left\|\bar{\vtheta}_k-\bm{1}_N\otimes \vtheta_c+ \bm{1}_N\otimes \vtheta_c \right\|^2_2\\
     \leq & 2NR^2_{\max} + 16 \left\|\bm{1}_N\otimes \vtheta_c\right\|^2_2
     + 16\left\|\tilde{\vtheta}_k\right\|^2\\
     \leq &  2NR^2_{\max} + 16N \left\|\vtheta_c \right\|^2_2
     + 16\left\|\tilde{\vtheta}_k\right\|^2\\
     \leq &  2NR^2_{\max} + 16N \left(\frac{R_{\max}}{w(1-\gamma)}\right)^2
     + 16\left\|\tilde{\vtheta}_k\right\|^2\\
     =& \frac{32NR_{\max}^2}{w^2(1-\gamma)^2}+ 16\left\|\tilde{\vtheta}_k\right\|^2. 
\end{align*}

The second inequality follows from the fact that \(\left\|\va+\vb\right\|^2_2\leq 2\left\|\va\right\|^2_2+2\left\|\vb\right\|^2_2 \) for \(\va,\vb\in\R^{Nq}\).  The last inequality follows from Lemma~\ref{theta_c_bound}.

\end{proof}

%% file: app/ode.tex
\subsection{Proof of Lemma~\ref{lem:ode_lyapunov}}\label{app:proof:lem:ode_lyapunov}

We will consider the following positive definite matrix: 
\begin{align}
    \mS = \begin{bmatrix}
        \beta \mI_{n} & \mM\\
        \mM &  \beta\mI_n
    \end{bmatrix} \in \R^{2n\times 2n}, \label{eq:S} 
\end{align}
where the choice of positive constant \(\beta\in\R\) in the statement of Lemma~\ref{lem:ode_lyapunov} will be deferred. Using the Schur complement in Lemma~\ref{lem:schur_complement} in the Appendix Section~\ref{subsec:technical}, we can see that if \(\beta>2\lambda_{\max}(\mM)\), the following holds:

\begin{align*}
    \begin{bmatrix}
        \frac{\beta}{2}\mI_n & \bm{0}_{n\times n}\\
       \bm{0}_{n\times n}& \frac{\beta}{2}\mI_n
    \end{bmatrix}
    \prec \mS
    \prec
    \begin{bmatrix}
        2\beta \mI_n &\bm{0}_{n\times n}\\
        \bm{0}_{n\times n}& 2\beta \mI_n
    \end{bmatrix}.
\end{align*}

Now, we have the following relation:
\begin{align*}
   &2\begin{bmatrix}
        \vtheta\\
        \mM\mM^{\dagger}\vw
    \end{bmatrix}^{\top}\mS\begin{bmatrix}
        -\mU & -\mM\\
        \mM & \bm{0}_{n\times n}
    \end{bmatrix}\begin{bmatrix}
        \vtheta\\
        \mM\mM^{\dagger}\vw
    \end{bmatrix} \\
    =& \begin{bmatrix}
        \vtheta\\
        \mM\mM^{\dagger}\vw
    \end{bmatrix}^{\top}\begin{bmatrix}
        \beta\mI_n & \mM\\
        \mM & \beta\mI_n
    \end{bmatrix}
    \begin{bmatrix}
        -\mU & -\mM\\
        \mM & \bm{0}_{n\times n}
    \end{bmatrix}\begin{bmatrix}
        \vtheta\\
        \mM\mM^{\dagger}\vw
    \end{bmatrix}+\begin{bmatrix}
        \vtheta\\
        \mM\mM^{\dagger}\vw
    \end{bmatrix}^{\top}\begin{bmatrix}
        -\mU^{\top} & \mM\\
        -\mM & \bm{0}_{n\times n}
    \end{bmatrix}
    \begin{bmatrix}
        \beta\mI_n & \mM\\
        \mM & \beta\mI_n
    \end{bmatrix}\begin{bmatrix}
        \vtheta\\
        \mM\mM^{\dagger}\vw
    \end{bmatrix} \\
 =& \begin{bmatrix}
        \vtheta\\
        \mM\mM^{\dagger}\vw
    \end{bmatrix}^{\top}\begin{bmatrix}
     -\beta\mU+\mM^2 &-\beta\mM\\
     -\mM\mU+\beta \mM & -\mM^2
 \end{bmatrix}   \begin{bmatrix}
        \vtheta\\
        \mM\mM^{\dagger}\vw
    \end{bmatrix}
 + \begin{bmatrix}
        \vtheta_t\\
        \mM\mM^{\dagger}\vw
    \end{bmatrix}^{\top}\begin{bmatrix}
     -\beta\mU^{\top}+\mM^2 & -\mU^{\top}\mM+\beta\mM\\
     -\beta\mM & - \mM^2
 \end{bmatrix}\begin{bmatrix}
        \vtheta\\
        \mM\mM^{\dagger}\vw
    \end{bmatrix}\\
 =& \begin{bmatrix}
        \vtheta\\
        \mM\mM^{\dagger}\vw
    \end{bmatrix}^{\top}\begin{bmatrix}
     -\beta(\mU+\mU^{\top}) + 2\mM^2 & -\mU^{\top}\mM\\
     -\mM\mU & -2\mM^2
 \end{bmatrix}\begin{bmatrix}
        \vtheta\\
        \mM\mM^{\dagger}\vw
    \end{bmatrix},
\end{align*}
where the first equality follows from plugging in \(\mS\) in~(\ref{eq:S}). Expanding the terms, we get

\begin{align*}
 &2 \begin{bmatrix}
        \vtheta\\
        \mM\mM^{\dagger}\vw
    \end{bmatrix}^{\top} \mS   \begin{bmatrix}
        -\mU & -\mM\\
        \mM & \bm{0}_{n\times n}
    \end{bmatrix}       \begin{bmatrix}
        \vtheta\\
        \mM\mM^{\dagger}\vw
    \end{bmatrix} \\
  = &  \begin{bmatrix}
        \vtheta\\
        \mM\mM^{\dagger}\vw
    \end{bmatrix}^{\top} \begin{bmatrix}
        -\beta (\mU+\mU^{\top})+2\mM^2 & -\mU^{\top}\mM \\
        -\mM\mU & -2 \mM^2
    \end{bmatrix}       \begin{bmatrix}
        \vtheta\\
        \mM\mM^{\dagger}\vw
    \end{bmatrix}  \\
 =& \vtheta^{\top}(-\beta(\mU+\mU^{\top})+2\mM^2) \vtheta - \vw^{\top}\mM\mU\vtheta-
 \vtheta^{\top}\mU^{\top} \mM \vw - 2 \vw^{\top}\mM^2\vw \\
 =& \begin{bmatrix}
     \vtheta\\
        \mM\vw
 \end{bmatrix}^{\top}
 \begin{bmatrix}
     -\beta(\mU+\mU^{\top})+2\mM^2 & -\mU^{\top}\\
     -\mU & -2\mI_n
 \end{bmatrix}
 \begin{bmatrix}
     \vtheta\\
        \mM\vw
 \end{bmatrix},
\end{align*}
where the second last equality follows from the axiom of Moore-Penrose pseudo inverse of symmetric matrices in Lemma~\ref{lem:moore_penrose} in the Appendix Section~\ref{subsec:technical}, i.e., \(\mM\mM^{\dagger}\mM=\mM\mM\mM^{\dagger}=\mM^{\dagger}\mM\mM=\mM\).

Now, it is enough to choose \(\beta >0\) that satisfies following relation:
\begin{align*}
 &    \begin{bmatrix}
    - \beta(\mU+\mU^{\top})+2\mM^2 & -\mU^{\top}\\
   -  \mU & -2\mI_n
 \end{bmatrix} \prec -\begin{bmatrix}
     \mI_n & \bm{0}_{n\times n} \\
    \bm{0}_{n\times n} & \mI_n
 \end{bmatrix}\\
\iff & \begin{bmatrix}
   - \beta (\mU+\mU^{\top})+2\mM^2+\mI_n & -\mU^{\top}\\
     -\mU & -\mI_n
\end{bmatrix} \prec \bm{0}_{2n\times 2n}.
\end{align*}
The above relation can be shown using Schur's complement Lemma~\ref{lem:moore_penrose} in the Appendix Section~\ref{subsec:technical},
\begin{align*}
   - \beta(\mU+\mU^{\top})+2\mM^2 +\mI_n +\mU\mU^{\top} \prec 0,
\end{align*}
which holds when \(\beta \) satisfies
\begin{align*}
     \beta \lambda_{\min}(\mU+\mU^{\top}) &> 2\lambda_{\max}(\mM)^2 +1+ \left\| \mU \right\|_2^2 \\
    \iff  \beta &> \frac{2\lambda_{\max}(\mM)^2+1 + \left\| \mU \right\|_2^2}{\lambda_{\min}(\mU+\mU^{\top})}.
\end{align*}

Therefore, we get

\begin{align*}
    \begin{bmatrix}
     \vtheta\\
        \mM\vw
 \end{bmatrix}^{\top}
 \begin{bmatrix}
     -\beta(\mU+\mU^{\top})+2\mM^2 & \mU^{\top}\\
     \mU & -2\mI
 \end{bmatrix}
 \begin{bmatrix}
     \vtheta\\
        \mM\vw
 \end{bmatrix} \leq & - \begin{bmatrix}
     \vtheta\\
        \mM\vw
 \end{bmatrix}^{\top}
 \begin{bmatrix}
     \vtheta\\
        \mM\vw
 \end{bmatrix}\\
 \leq & -\left\|\vtheta\right\|^2_2- \left\| \mM\vw\right\|^2_2\\
 \leq & - \min \{1,\lambda^+_{\min}(\mM)^2 \} \left\| \begin{bmatrix}
     \vtheta\\
     \mM\mM^{\dagger} \vw
 \end{bmatrix} \right\|_2^2,
\end{align*}
where the last inequality follows from the inequality that \(\left\|\mM\mM^{\dagger} \vw\right\|_2=  \left\| \mM^{\dagger}\mM \vw\right\|_2 \leq \left\|\mM^{\dagger} \right\|_2 \left\|\mM\vw \right\|_2 \leq \frac{1}{\lambda_{\min}^+(\mM)} \left\|\mM \vw \right\|_2 \). Hence, it is sufficient to choose \( \beta = \max\left\{ \frac{2\lambda_{\max}(\mM)^2 +2+ \left\| \mU \right\|_2^2}{\lambda_{\min}(\mU+\mU^{\top})} ,4\lambda_{\max}(\mM) \right\}\).

\subsection{Proof of Theorem~\ref{thm:ode:bound}}\label{app:thm:ode:bound}
\begin{proof}
Let us consider the quadratic Lyapunov function candidate \(V(\vtheta,\vw)=\begin{bmatrix}
    \vtheta \\
    \mM\mM^{\dagger}\vw
\end{bmatrix}^{\top}\mS\begin{bmatrix}
    \vtheta \\
    \mM\mM^{\dagger}\vw
\end{bmatrix}\) where \(\mS\in \R^{2n\times 2n}\) is symmetric positive definite matrix in Lemma~\ref{lem:ode_lyapunov}. The time derivative of \(V(\vtheta_t,\vw_t)\) along the solution of~(\ref{eq:primal_dual_ode}) becomes
\begin{align*}
    \frac{d}{dt}V(\vtheta_t,\vw_t) =&  2 \left(\frac{d}{dt}\begin{bmatrix}
    \vtheta_t\\
    \mM\mM^{\dagger}\vw_t
    \end{bmatrix}\right)^{\top} \mS \begin{bmatrix}
    \vtheta \\
    \mM\mM^{\dagger}\vw
\end{bmatrix}\\
=& 2 \begin{bmatrix}
    -\mU\vtheta_t-\mM \vw_t\\
    \mM\mM^{\dagger}\mM \vtheta_t
\end{bmatrix}^{\top}\mS
\begin{bmatrix}
    \vtheta_t \\
    \mM\mM^{\dagger}\vw_t
\end{bmatrix}\\
=& 2 \begin{bmatrix}
    -\mU\vtheta_t-\mM\mM\mM^{\dagger}\vw_t\\
    \mM \vtheta_t
\end{bmatrix}^{\top}\mS \begin{bmatrix}
    \vtheta_t \\
    \mM\mM^{\dagger}\vw_t
\end{bmatrix}\\\
=&2\begin{bmatrix}
    \vtheta_t \\
    \mM\mM^{\dagger}\vw_t
\end{bmatrix}^{\top} \begin{bmatrix}
   - \mU & -\mM\\
    \mM & \bm{0}_{n\times n}
\end{bmatrix}^{\top} \mS \begin{bmatrix}
    \vtheta_t \\
    \mM\mM^{\dagger}\vw_t
\end{bmatrix}\\
\leq & -2\min \{1,\lambda^+_{\min}(\mM)^2 \}\left\|\begin{bmatrix}
        \vtheta_t\\
        \mM\mM^{\dagger}\vw_t
    \end{bmatrix} \right\|^2_2\\
\leq & -2\min \{1,\lambda^+_{\min}(\mM)^2 \} \frac{1}{\lambda_{\max}(\mS)} V(\vtheta_t,\vw_t),
\end{align*}
where the second last inequality comes from Lemma~\ref{lem:ode_lyapunov}. The last inequality follows from the fact that \( V(\vtheta_t,\vw_t) \leq \lambda_{\max}(\mS) \left\| \begin{bmatrix}
    \vtheta_t\\
    \mM\mM^{\dagger}\vw_t
\end{bmatrix}\right\|_2^2  \).
From the Lyapunov method, this inequality results in
\begin{align*}
    V(\vtheta_t,\vw_t) \leq  \exp \left(-\frac{ \min \{1,\lambda^+_{\min}(\mM)^2\} }{\max\left\{ \frac{2\lambda_{\max}(\mM)^2 +2+ \left\| \mU \right\|_2^2}{\lambda_{\min}(\mU+\mU^{\top})} ,4\lambda_{\max}(\mM) \right\}} t\right) V(\vtheta_0,\vw_0).
\end{align*}
This completes the proof.    
\end{proof}

\section{Comparison with the result of~\citealp{ozaslan2023global,cisneros2020distributed,gokhale2023contractivity}}\label{app:sec:compairosn}

We will consider \( f(\vx) = \frac{1}{2}\left\|\vx\right\|_{\mB}^2\) where \(\vx\in\R^n \) and \(\mB\in\R^{n\times n}\) is symmetric positive definite matrix. Then, \(\nabla^2 f(\vx) = \mB \), and \(f(\vx)\) is \(\lambda_{\min}(\mB)\)-strongly convex and \(\lambda_{\max}(\mB)\)-smooth. Theorem 8 in~\citealp{gokhale2023contractivity} states exponential convergence rate of $\gO\left( \exp\left(- \min \left\{ \frac{\lambda_{\min}^+(\mM)^2}{\lambda_{\max}(\mU)},\frac{\lambda_{\min}^+(\mM)^2}{\lambda_{\max}(\mM)^2} \lambda_{\min}(\mU) \right\}  t\right) \right)$. When $\frac{\lambda_{\min}^+(\mM)^2}{\lambda_{\max}(\mM)^2} $ is the dominant term, the bound yields the convergence rate $\gO\left(  \exp \left(- \frac{\lambda_{\min}^+(\mM)^2}{\lambda_{\max}(\mM)^2} t \right)\right)$. Our bound in Theorem~\ref{thm:ode:bound} also results to the convergence rate of $\gO\left(  \exp \left(- \frac{\lambda_{\min}^+(\mM)^2}{\lambda_{\max}(\mM)^2}t  \right)\right)$ when $\lambda_{\min}^+(\mM)$ is small, which matches the convergence provided in~\citealp{gokhale2023contractivity}.

 Letting \(V(\vtheta_t,\vw_t)=\left\|\vtheta_t\right\|_2^2+ \left\| \mM\mM^{\dagger}\vw_t-\vw^* \right\|^2_2\), the result of 
Theorem 2 in~\citealp{ozaslan2023global} leads to

\begin{align*}
   & V(\vtheta_t,\vw_t) \\
   \leq & 2\exp\left(- \frac{2\lambda_{\min}(\mB) \min\{\lambda_{\min}(\mB)^2 , \lambda_{\min}^+(\mM)^2 \}}{(\lambda_{\max}(\mB)^2+\lambda_{\max}(\mM)^2+1) (1+2\lambda_{\min}(\mB)\lambda_{\max}(\mB))} t\right) \left(\left\|\nabla L(\vtheta_0,\vw_0) \right\|_2^2+V(\vtheta_0,\vw_0)\right).
\end{align*}

When \(\lambda_{\min}(\mB)\to 0\), the above convergence rate becomes  \( \gO(\exp(-\lambda_{\min}(\mB)
^3t))\). Whereas, from Theorem~\ref{thm:ode:bound}, our result states \(\gO\left(\exp(-\lambda_{\min}(\mB)t\right)\) convergence rate under the same condition, which implies  tighter convergence rate.



        ~\citealp{cisneros2020distributed} proved exponential convergence rate for \(\begin{bmatrix}
            \vtheta_t\\
            \mR \vw_t
        \end{bmatrix}\), where \( \mM := \mR \bm{\Sigma} \mR^{\top}\) is the singular value decomposition of \(\mM\). Theorem 4 in~\citealp{cisneros2020distributed} leads to the following convergence rate: 
        \begin{align*}
            \gO\left( \exp\left( - \frac{\lambda_{\min}(\mB)}{\lambda_{\max}(\mM)^2+\frac{3}{4}\lambda_{\max}(\mM)\lambda^+_{\min}(\mM)^2 +\lambda_{\max}(\mB)^2}\frac{\lambda_{\max}(\mM)\lambda_{\min}^+(\mM)^2}{\lambda_{\max}(\mM)+1} t \right)  \right).
        \end{align*}

When \(\lambda_{\max}(\mM)\approx \lambda_{\min}^+(\mM)\to 0\), the bound implies 
\begin{align*}
    \gO \left( \exp \left( -\lambda_{\max}(\mM)\lambda_{\min}^+(\mM)^2 t \right) \right),
\end{align*}
where as our bound in Theorem~\ref{thm:ode:bound} implies tighter convergence rate of
\begin{align*}
    \gO \left( \exp\left( - \lambda^+_{\min}(\mM)^2t \right)\right).
\end{align*}
The overall comparison with~\citealp{ozaslan2023global,cisneros2020distributed,gokhale2023contractivity} is summarized in the Table~\ref{tab:comparison_finite_bound}.

\begin{table}[h!]
\begin{adjustbox}{width=\columnwidth,center}
\centering
 \begin{tabular}{|c| c | c|} 
 \hline
  & Convergence rate & Condition \\ [0.5ex] 
 \hline\hline
\multicolumn{1}{|c|}{~\citealp{ozaslan2023global}}  & \multicolumn{1}{|c|}{$\gO\left(\exp\left(-\lambda_{\min}(\mU)^3t\right)\right)$ }& \multirow{2}{*}{$\lambda_{\min}(\mU)\to 0$}\\\cline{1-2}
 \multicolumn{1}{|c|}{\cellcolor{yellow}Ours} & \multicolumn{1}{|c|}{\cellcolor{yellow}$\gO\left(\exp\left(-\lambda_{\min}(\mU)t\right)\right)$ } &  \\
 \hline\hline
\multicolumn{1}{|c|}{~\citealp{cisneros2020distributed}}  & \multicolumn{1}{|c|}{$\gO\left(\exp\left(-\lambda^+_{\min}(\mM)^3t\right)\right)$ }& \multirow{2}{*}{$\lambda_{\max}(\mM)\approx \lambda_{\min}^+(\mM)\to 0$}\\\cline{1-2}
 \multicolumn{1}{|c|}{\cellcolor{yellow}Ours} & \multicolumn{1}{|c|}{\cellcolor{yellow}$\gO\left(\exp\left(-\lambda^+_{\min}(\mM)^2t\right)\right)$ } &  \\
 \hline
 \hline
\multicolumn{1}{|c|}{~\citealp{gokhale2023contractivity}}  & \multirow{2}{*}{$ \gO\left( \exp\left(- \frac{\lambda^+_{\min}(\mM)^2}{\lambda_{\max}(\mM)^2}t\right) \right)$ }& \multirow{2}{*}{$\lambda_{\min}^+(\mM)\to 0,\;\lambda_{\max}(\mM)\to\infty$}\\
 \multicolumn{1}{|c|}{\cellcolor{yellow}Ours  } & & \\
 \hline
 \end{tabular}
 \end{adjustbox}
 \caption{$t\ge 0$ stands for time.}\label{table:ode}
\end{table}

%% file: app/algo1.tex
\subsection{Proof of Lemma~\ref{lem:algo1:lyapunov_equation_for_projected_iterate}}\label{app:proof:lem:lyapunov_equation_for_projected_iterate}

We will consider the following positive definite matrix:

\begin{align}
    \mG:= \begin{bmatrix}
        \beta\mI_{Nq} & \bar{\mL}\\
        \bar{\mL} & \beta\mI_{Nq}
    \end{bmatrix} \in \R^{2Nq\times 2Nq}, \label{eq:G}
\end{align}

where the choice of positive constant \(\beta\in\R\) will be deferred. Using the Schur complement in Lemma~\ref{lem:schur_complement} in the Appendix Section~\ref{subsec:technical}, we can see that if \( \beta >2  \lambda_{\max}(\bar{\mL})\), the following holds:
\begin{align*}
    \begin{bmatrix}
        \frac{\beta}{2}\mI & \bm{0}_{Nq\times Nq}\\
        \bm{0}_{Nq\times Nq} & \frac{\beta}{2}\mI
    \end{bmatrix}
    \prec \mG
    \prec
    \begin{bmatrix}
        2\beta \mI &\bm{0}_{Nq\times Nq}\\
       \bm{0}_{Nq\times Nq} & 2\beta \mI
    \end{bmatrix}.
\end{align*}


Now, we have the following relation:
\begin{align*}
   &         2\begin{bmatrix}
            \tilde{\vtheta}\\
         \bar{\mL}\bar{\mL}^{\dagger}  \tilde{\vw}
        \end{bmatrix}^{\top}
        \mG\begin{bmatrix}
    \bar{\mA}-\eta\bar{\mL} & -\eta \bar{\mL}\\
   \eta \bar{\mL} & \bm{0}_{Nq\times Nq}
\end{bmatrix}        
\begin{bmatrix}
            \tilde{\vtheta}\\
            \bar{\mL}\bar{\mL}^{\dagger}  \tilde{\vw}
        \end{bmatrix} \\
=& \begin{bmatrix}
            \tilde{\vtheta}\\
         \bar{\mL}\bar{\mL}^{\dagger}  \tilde{\vw}
        \end{bmatrix}^{\top}
        \mG\begin{bmatrix}
    \bar{\mA}-\eta \bar{\mL} & -\eta\bar{\mL}\\
   \eta \bar{\mL} & \bm{0}_{Nq\times Nq},\\
\end{bmatrix}        
\begin{bmatrix}
            \tilde{\vtheta}\\
            \bar{\mL}\bar{\mL}^{\dagger}  \tilde{\vw}
        \end{bmatrix} +
        \begin{bmatrix}
            \tilde{\vtheta}\\
         \bar{\mL}\bar{\mL}^{\dagger}  \tilde{\vw}
        \end{bmatrix}^{\top}
        \begin{bmatrix}
    \bar{\mA}^{\top}-\eta\bar{\mL} & \eta\bar{\mL}\\
    -\eta\bar{\mL} & \bm{0}_{Nq\times Nq},\\
\end{bmatrix}\mG        
\begin{bmatrix}
            \tilde{\vtheta}\\
            \bar{\mL}\bar{\mL}^{\dagger}  \tilde{\vw}
        \end{bmatrix} \\
=&   \begin{bmatrix}
            \tilde{\vtheta}\\
         \bar{\mL}\bar{\mL}^{\dagger}  \tilde{\vw}
        \end{bmatrix}^{\top}  \begin{bmatrix}
        \beta(\bar{\mA}+\bar{\mA}^{\top}-2\eta\bar{\mL})+2\eta\bar{\mL}^2
        &(\bar{\mA}^{\top}-\eta\bar{\mL})\bar{\mL}\\
        \bar{\mL}(\bar{\mA}-\eta\bar{\mL}) &-2\eta\bar{\mL}^2
    \end{bmatrix}\begin{bmatrix}
            \tilde{\vtheta}\\
            \bar{\mL}\bar{\mL}^{\dagger}  \tilde{\vw}
        \end{bmatrix} ,
\end{align*}

where the last equality follows from plugging the choice of \(\mG\) in~(\ref{eq:G}). Expanding the terms, we get
\begin{align}
   & \begin{bmatrix}
            \tilde{\vtheta}\\
         \bar{\mL}\bar{\mL}^{\dagger}  \tilde{\vw}
        \end{bmatrix}^{\top}  \begin{bmatrix}
         \beta (\bar{\mA}+\bar{\mA}^{\top}-2\eta\bar{\mL})+2\eta\bar{\mL}^2
        &(\bar{\mA}^{\top}-\eta\bar{\mL})\bar{\mL}\\
        \bar{\mL}(\bar{\mA}-\eta \bar{\mL}) &-2\eta\bar{\mL}^2
    \end{bmatrix}\begin{bmatrix}
            \tilde{\vtheta}\\
            \bar{\mL}\bar{\mL}^{\dagger}  \tilde{\vw}
        \end{bmatrix} \nonumber \\
=& \tilde{\vtheta}^{\top}(\beta(\bar{\mA}+\bar{\mA}^{\top}-2\eta\bar{\mL})+2\eta\bar{\mL}^2)\tilde{\vtheta}
    +\tilde{\vtheta}^{\top}(\bar{\mA}^{\top}-\eta\bar{\mL})\bar{\mL}\bar{\mL}\bar{\mL}^{\dagger}\tilde{\vw} \nonumber \\
    &+ \tilde{\vw}^{\top}\bar{\mL}^{\dagger}\bar{\mL}\bar{\mL}(\bar{\mA}-\eta\bar{\mL})\bar{\vtheta}-2\eta\tilde{\vw}^{\top}\bar{\mL}^{\dagger}\bar{\mL} \bar{\mL}^2\bar{\mL}\bar{\mL}^{\dagger}\tilde{\vw} \nonumber \\
    =&\tilde{\vtheta}^{\top}(\beta(\bar{\mA}+\bar{\mA}^{\top}-2\eta\bar{\mL})+2\eta\bar{\mL}^2)\tilde{\vtheta}
    +\tilde{\vtheta}(\bar{\mA}^{\top}-\eta\bar{\mL})\bar{\mL}\tilde{\vw} \nonumber \\
    &+ \tilde{\vw}^{\top}\bar{\mL}(\bar{\mA}-\eta\bar{\mL})\bar{\vtheta} -2 \eta \left\|\bar{\mL} \tilde{\vw} \right\|^2_2 \nonumber \\
    =&     \begin{bmatrix}
        \tilde{\vtheta}\\
        \bar{\mL}\tilde{\vw}
    \end{bmatrix}^{\top}
    \begin{bmatrix}
        \beta(\bar{\mA}+\bar{\mA}^{\top}-2\eta\bar{\mL})+ 2\eta \bar{\mL}^2 & \bar{\mA}^{\top}-\eta\bar{\mL}\\
        \bar{\mA}-\eta\bar{\mL} & -2 \eta \mI
    \end{bmatrix}
        \begin{bmatrix}
        \tilde{\vtheta}\\
        \bar{\mL}\tilde{\vw}
    \end{bmatrix} \label{ineq:primal_dual_1},
\end{align}
where the second last equality follows from the axiom of Moore-Penrose axiom of symmetric matrices in Lemma~\ref{lem:moore_penrose} in the Appendix Section~\ref{subsec:technical}, i.e., \(\bar{\mL}\bar{\mL}^{\dagger}\bar{\mL}=\bar{\mL}\bar{\mL}\bar{\mL}^{\dagger}=\bar{\mL}^{\dagger}\bar{\mL}\bar{\mL}=\bar{\mL}\).

Now, it is enough to choose \(c\) that satisfies following relation:
\begin{align}   
&\begin{bmatrix}
\beta(\bar{\mA}+\bar{\mA}^{\top}-2\eta\bar{\mL})+2\eta\bar{\mL}^2 & \bar{\mA}^{\top}-\eta\bar{\mL}\\
        \bar{\mA}-\eta\bar{\mL} & -2\eta  \mI
    \end{bmatrix}  \preceq - \begin{bmatrix}
       \mI_{Nq} & \bm{0}_{Nq\times Nq}\\
        \bm{0}_{Nq\times Nq} & \eta \mI_{Nq} 
    \end{bmatrix}  \label{ineq:primal_dual_2} \\
 \iff &    \begin{bmatrix}
\beta(\bar{\mA}+\bar{\mA}^{\top}-2\eta\bar{\mL})+2\eta\bar{\mL}^2 & \bar{\mA}^{\top}-\eta\bar{\mL}\\
        \bar{\mA}-\eta\bar{\mL} & -2 \eta \mI
    \end{bmatrix}  + \begin{bmatrix}
        \mI_{Nq} & \bm{0}_{Nq\times Nq}\\
        \bm{0}_{Nq\times Nq} & \eta \mI_{Nq} 
    \end{bmatrix}  \preceq \bm{0}_{2Nq\times 2Nq} .  \nonumber
\end{align}

Using the result \(\bar{\mA}+\bar{\mA}^{\top}\preceq 2(\gamma-1)w\) from Lemma~\ref{lem:A_bound}, we have

\begin{align*}
    &    \begin{bmatrix}
\beta(\bar{\mA}+\bar{\mA}^{\top}-2\eta\bar{\mL})+2\eta\bar{\mL}^2 & \bar{\mA}^{\top}-\eta\bar{\mL}\\
        \bar{\mA}-\eta\bar{\mL} & -2\eta \mI
    \end{bmatrix}  + \begin{bmatrix}
        \mI_{Nq} & \bm{0}_{Nq\times Nq}\\
        \bm{0}_{Nq\times Nq} &\eta\mI_{Nq} 
    \end{bmatrix} \\
    \preceq & \begin{bmatrix}
        (2\beta(\gamma-1)w+1+2\eta\lambda_{\max}(\bar{\mL})^2)\mI_{Nq} & \bar{\mA}^{\top}-\eta\bar{\mL}\\
        \bar{\mA}-\eta\bar{\mL} & -\eta\mI
    \end{bmatrix}.
\end{align*}
The inequality follows from the fact that \( \bar{\mL}^2 \) is positive semi-definite matrix. To make the above matrix negative definite, according to the Schur complement argument in Lemma~\ref{lem:schur_complement}, we need
\begin{align}
    (2\beta(\gamma-1)w+1+2\eta\lambda_{\max}(\bar{\mL})^2) \mI_{Nq}+\frac{1}{\eta}(\bar{\mA}-\eta\bar{\mL})(\bar{\mA}^{\top}-\eta\bar{\mL}) \prec 0, \label{ineq:c:nd}
\end{align}
which can be satisfied if the following holds for \(c\):
\begin{align*}
    &(2\beta(\gamma-1)w+1+2\eta \lambda_{\max}(\bar{\mL})^2) + \frac{1}{\eta}\left\| \bar{\mA}- \eta \bar{\mL} \right\|_2^2 < 0\\
\iff &  \frac{\frac{1}{\eta}\left\| \bar{\mA}-\eta \bar{\mL} \right\|_2^2+1+2\eta \lambda_{\max}(\bar{\mL})^2}{2(1-\gamma)w} < \beta . 
\end{align*}

Since $\left\|\bar{\mA}\right\|^2_2 \leq 4$ from Lemma~\ref{lem:A_bound}, and $a^2+b^2\geq 2ab$ for $a,b\in \R$, it suffices to satisfy
\begin{align*}
    \beta > \frac{8+\eta+4\eta^2\lambda_{\max}(\bar{\mL})^2}{2\eta(1-\gamma)w}.    
\end{align*}

Therefore, choosing 
\begin{align*}
    \beta = \frac{8+\eta+4\eta^2 \lambda_{\max}(\bar{\mL})^2}{\eta (1-\gamma)w}
\end{align*}
suffices to satisfy~(\ref{ineq:c:nd}). Note that $\beta \geq \frac{1}{(1-\gamma)w}+\frac{8}{\eta(1-\gamma)w}+\frac{4\eta\lambda_{\max}(\bar{\mL})^2}{(1-\gamma)w} > 4 \lambda_{\max}(\bar{\mL}) \frac{1}{(1-\gamma)w}\geq 4\lambda_{\max}(\bar{\mL})$. Applying the relation~(\ref{ineq:primal_dual_2}) to~(\ref{ineq:primal_dual_1}) yields the following result:


\begin{align*}
      \begin{bmatrix}
        \tilde{\vtheta}\\
        \bar{\mL}\tilde{\vw}
    \end{bmatrix}^{\top}
    \begin{bmatrix}
    \beta(\bar{\mA}+\bar{\mA}^{\top}-2\eta\bar{\mL})+2\eta\bar{\mL}^2 & \bar{\mA}^{\top}-\eta\bar{\mL}\\
        \bar{\mA}-\eta\bar{\mL} & -2\eta \mI
    \end{bmatrix}
        \begin{bmatrix}
        \tilde{\vtheta}\\
        \bar{\mL}\tilde{\vw}
    \end{bmatrix} 
    \leq & -\begin{bmatrix}
        \tilde{\vtheta}\\
        \bar{\mL}\tilde{\vw}
    \end{bmatrix}^{\top} \begin{bmatrix}
        \mI_{Nq} & \bm{0}_{Nq\times Nq}\\
        \bm{0}_{Nq\times Nq} &\eta \mI_{Nq} 
    \end{bmatrix} \begin{bmatrix}
        \tilde{\vtheta}\\
        \bar{\mL}\tilde{\vw}
          \end{bmatrix}\\
        =&-  \left\|\tilde{\vtheta}\right\|_2^2
        - \eta \left\| \bar{\mL}\tilde{\vw} \right\|_2^2  \\
     \leq & -  \left\|\tilde{\vtheta} \right\|_2^2
         -\eta \lambda^+_{\min}(\bar{\mL})^2\left\| \bar{\mL}^{\dagger}\bar{\mL}\tilde{\vw}\right\|_2^2\\
    =& -\min\left\{1,\eta \lambda^+_{\min}(\bar{\mL})^2\right\}\left\|\begin{bmatrix}
        \tilde{\vtheta}\\
        \bar{\mL}^{\dagger}\bar{\mL} \tilde{\vw}
    \end{bmatrix}  \right\|_2^2,
\end{align*}

where the last inequality follows from the following relation:
\begin{align*}
    \left\| \bar{\mL}^{\dagger}\bar{\mL}\tilde{\vw}\right\|_2 \leq \left\| \bar{\mL}^{\dagger} \right\|_2 \left\|\bar{\mL}\tilde{\vw}\right\|_2 = \frac{1}{\lambda_{\min}^+(\bar{\mL})}\left\| \bar{\mL} \tilde{\vw}\right\|_2  .
\end{align*}

\section{Stochastic recursive update : i.i.d. observation model}\label{app:sec:iid}

In this section, we will consider the i.i.d. observation model of the sequence \(\{ o_k \}_{k\in \sN_0}\) and \( o_k \in \gS \times \gS \times \Pi_{i=1}^N I \) where $I$ is the closed interval \([-R_{\max},R_{\max}]\) in \(\R\). We consider the following general stochastic recursive update~\citep{robbins1951stochastic}, for \(k\in\sN_0\) and \(\vz_0\in\R^{2Nq}\):
\begin{align}
    \vz_{k+1} = \vz_k+\alpha_k (\mE\vz_k+\bm{\xi}(o_k;\vz_k)), \label{eq:general_sa_iid}
\end{align}
where \(\mE\in\R^{2Nq\times 2Nq}\), \(\bm{\xi}(\cdot;\vz): \gS \times \gS \times\Pi^N_{i=1}I \to \R^{2Nq}\) is a function parameterized by \(\vz\in\R^{2Nq}\), and \(\alpha_k\in(0,1)\).

\begin{assumption}\label{assmp:iid_general}
    \begin{enumerate}
        \item[1.] For \(k\in\sN_0\), \(\bm{\xi}(o_k;\vz_k)\) satisfies the following bound:
\begin{align*}
 \E\left[  \left\|\bm{\xi}(o_k;\vz_k)\right\|_2^2\right] \leq C_1 \E\left[ \left\|\vz_k\right\|_2^2\right] + C_2.
\end{align*}
\item[2.] For \(k\in\sN_0\), \(\{o_i\}_{i=1}^k\) is sampled from i.i.d. distribution, and
\begin{align*}
\E\left[\bm{\xi}(o_k;\vz_k) \middle | \gF_{k-1}\right] =0 ,
\end{align*}
where \(\gF_{k}:=\sigma(o_1,o_2,\dots,o_{k})\) for \(k\in \sN\).
\item[3.] There exists a positive symmetric definite matrix \(\mQ\in\R^{2Nq\times 2Nq}\) and positive real constant \(\kappa\) such that, for \(k\in\sN_0\),
\begin{align*}
    \vz_k^{\top}\mE \mQ \vz_k \leq -\kappa \left\|\vz_k\right\|_2.
\end{align*}
    \end{enumerate}
\end{assumption}

We will introduce one lemma:.

\begin{lemma}\label{lem:iid:vz_k+1-vz_k}
Under the Assumption~\ref{assmp:iid_general}, for \( k \in \sN_0 \), we have
    \begin{align*}
       \E\left[ (\vz_{k+1}-\vz_k)^{\top}\mQ(\vz_{k+1}-\vz_k) \right] \leq  2 \alpha_k^2\left\|\mQ\right\|_2 \left( \left(\left\|\mE\right\|^2_2+C_1\right)\E\left[ \left\|\vz_k\right\|^2_2 \right]+C_2\right).
    \end{align*}
\end{lemma}
\begin{proof}
We have
    \begin{align*}
        &\E\left[ (\vz_{k+1}-\vz_k)^{\top}\mQ(\vz_{k+1}-\vz_k) \right]\\
        \leq &\left\|\mQ\right\|_2\E\left[ \left\| \vz_{k+1}-\vz_k\right\|_2^2 \right] \\
         =& \left\|\mQ\right\|_2 \E\left[\left\| \alpha_k \mE\vz_k+\alpha_k\bm{\xi}(o_k;\vz_k) \middle |\right\|_2^2\right]\\
         \leq & 2\alpha_k^2\left\|\mQ\right\|_2 \left(\E\left[\left\|\mE\right\|_2^2\left\|\vz_k\right\|_2^2\right]+\E\left[\left\|\bm{\xi}(o_k;\vz_k)\right\|_2^2\right]\right)\\
         \leq & 2\alpha_k^2\left\|\mQ\right\|_2 \left( \left(\left\|\mE\right\|^2_2+C_1\right)\E\left[ \left\|\vz_k\right\|^2_2 \right]+C_2\right).
    \end{align*}
    The first inequality follows from positive definiteness of \(\mQ\). The first equality follows from the update in~(\ref{eq:general_sa_iid}). The second inequality follows from the relation \(\left\|\va+\vb\right\|^2_2\leq 2\left\|\va\right\|_2^2+2\left\|\vb\right\|_2^2 \) for \(\va,\vb\in\R^{2Nq}\). The last inequality follows from the first item in Assumption~\ref{assmp:iid_general}. 
\end{proof}

\begin{theorem}\label{thm:iid_general}
Suppose Assumption~\ref{assmp:iid_general} holds, and let \(V(\vz):=\vz^{\top}\mQ\vz\) for \(\vz\in\R^{2Nq} \).
    \begin{enumerate}
        \item[1.] Suppose we use constant step-size, i.e., \(\alpha_0=\alpha_1=\cdots=\alpha_k\), and \(\alpha_0 \leq \frac{\kappa\lambda_{\min}(\mQ) }{2\lambda_{\max}(\mQ)\left\|\mQ\right\|_2\left(E^2+C_1\right)}\), where $E$ is a positive constant that satisfies \(\left\|\mE \right\|_2\leq E\). For \(k\in\sN_0\), we have
            \begin{align*}
        \E\left[V(\vz_{k+1})\right] \leq \exp\left(-\frac{\kappa}{\lambda_{\max}(\mQ)} k\alpha_0\right)V(\vx_0) +2 \alpha_0 C_2 \left\|\mQ\right\|_2 \frac{\lambda_{\max}(\mQ)}{\kappa} +2 \alpha_0^2\left\|\mQ\right\|_2C_2.
    \end{align*}
    \item[2.] Suppose we have \(\alpha_t=\frac{h_1}{t+h_2}\) for \(t\in \sN_0\) and \(h_1 \geq \max\{2,\frac{2\lambda_{\max}(\mQ)}{\kappa} \}\) and \(\max\left\{2,h_1,h_1 \frac{2\lambda_{\max}(\mQ)\left\|\mQ\right\|_2\left(E^2+C_1\right)}{\kappa\lambda_{\min}(\mQ)}\right\} \leq  h_2 \). Then, we have
    \begin{align*}
     \E\left[ V(\vz_{k+1})\right] \leq &  \left(\frac{h_2}{k+h_2}\right)^{\frac{h_1\kappa}{\lambda_{\max}(\mQ)}}V(\vx_0)+\frac{2\left\|\mQ\right\|_2C_2h_1^2}{(k-1+h_2)} \frac{2^{\frac{2h_1\kappa}{\lambda_{\max}(\mQ)}}}{\frac{h_1\kappa}{\lambda_{\max}(\mQ)}-1}  +2\alpha_k^2\left\|\mQ\right\|_2C_2.
    \end{align*}
    \end{enumerate}
\end{theorem}

\begin{proof}
    From simple algebraic manipulation in~\citealp{srikant2019finite}, we have the following decomposition:

    \begin{align}
 & \E\left[V(\vz_{k+1})-V(\vz_k)\right] \nonumber   \\
=& \E\left[ (\vz_{k+1}-\vz_k)^{\top}\mQ (\vz_{k+1}-\vz_k) \right]+\E\left[ 2\vz_k^{\top}\mQ \vz_{k+1} \right] -2 \E\left[V(\vz_k)\right] \nonumber \\
=& \E\left[ (\vz_{k+1}-\vz_k)^{\top}\mQ (\vz_{k+1}-\vz_k) \right]+\E \left[ 2\vz_k^{\top}\mQ (\vz_{k+1}-\vz_k)\right] \nonumber\\
=&\underbrace{\E\left[ (\vz_{k+1}-\vz_k)^{\top}\mQ (\vz_{k+1}-\vz_k ) \right]}_{I_1}+\underbrace{\E\left[2\vz_k^{\top}\mQ(\vz_{k+1}-\vz_k-\alpha_k\mE \vz_k) \right]}_{I_2}+\underbrace{2\alpha_k \E\left[ \vz_k^{\top} \mQ \mE \vz_k \right]}_{I_3}.  \label{iid:decomposition_bound}
\end{align}

To bound \(I_1\), the result in Lemma~\ref{lem:iid:vz_k+1-vz_k} yields 

\begin{align*}
           \E\left[ (\vz_{k+1}-\vz_k)^{\top}\mQ(\vz_{k+1}-\vz_k) \right] \leq   2\alpha_k^2\left\|\mQ\right\|_2 \left( \left(\left\|\mE\right\|^2_2+C_1\right)\E\left[ \left\|\vz_k\right\|^2_2 \right]+C_2\right).
\end{align*}

The term \(I_2\) becomes zero due to the second item in Assumption~\ref{assmp:iid_general}, which leads to \(\E\left[2\vz_k^{\top}\mQ(\vz_{k+1}-\vz_k-\alpha_k\mE \vz_k) \right]= \alpha_k \E\left[2\vz_k^{\top}\mQ\E\left[\bm{\xi}(o_k;\vz_k)\middle|\gF_{k-1} \right]\right]=0\). Finally we can apply the third item in Assumption~\ref{assmp:iid_general} to bound $I_3$. Collecting the terms to bound~(\ref{iid:decomposition_bound}), we get

\begin{align}
    \E[V(\vz_{k+1})-V(\vz_k)] \leq  & 2\alpha_k^2\left\|\mQ\right\|_2 \left( \left(\left\|\mE\right\|^2_2+C_1\right)\E\left[ \left\|\vz_k\right\|^2_2 \right]+C_2\right)
    -2 \kappa \alpha_k \left\|\vz_k\right\|_2^2 \nonumber\\
    \leq & 2\alpha_k^2\left\|\mQ\right\|_2 \left(  \frac{\left\|\mE\right\|^2_2+C_1}{\lambda_{\min}(\mQ)}\E\left[V(\vz_k) \right]+C_2\right)
    -2 \frac{\kappa}{\lambda_{\max}(\mQ)} \alpha_k \E\left[V(\vz_k)\right]\nonumber\\
    =& \left(2\alpha_k^2\left\|\mQ\right\|_2 \frac{\left\|\mE\right\|^2_2+C_1}{\lambda_{\min}(\mQ)}-2 \frac{\kappa}{\lambda_{\max}(\mQ)} \alpha_k\right)\E\left[V(\vz_k)\right]+2\alpha_k^2\left\|\mQ\right\|_2 C_2 \label{ineq:iid_general_square_alpha-alpha}.
\end{align}

The second inequality follows from \( \lambda_{\min}(\mQ)\left\| \vz \right\|_2^2 \leq \left\|\vz\right\|_{\mQ}^2 \leq \lambda_{\max}(\mQ) \left\|\vz\right\|_2^2\). Moreover, the step-size conditions for both constant step-size and diminishing step-size leads to  
\begin{align*}
&2\left\|\mQ\right\|_2\frac{\left\|\mE\right\|^2_2+C_1}{\lambda_{\min}(\mQ)}\alpha_k^2-2\frac{\kappa}{\lambda_{\max}(\mQ)}\alpha_k \leq 2\left\|\mQ\right\|_2\frac{E^2+C_1}{\lambda_{\min}(\mQ)}\alpha_k^2-2\frac{\kappa}{\lambda_{\max}(\mQ)}\alpha_k \leq -\frac{\kappa}{\lambda_{\max}(\mQ)}\alpha_k.
\end{align*}

Applying the above result to~(\ref{ineq:iid_general_square_alpha-alpha}), we get

\begin{align}
    &\E[V(\vz_{k+1})] \nonumber \\
    \leq &  \left(1-\frac{\kappa}{\lambda_{\max}(\mQ)} \alpha_k\right)  \E\left[V(\vz_k)\right]
    + 2\alpha_k^2 \left\|\mQ\right\|_2C_2\nonumber\\
    \leq  & \Pi^k_{i=0}\left(1-\frac{\kappa}{\lambda_{\max}(\mQ)} \alpha_i\right)  \E\left[V(\vz_0)\right]
    +2\sum_{i=0}^{k-1} \alpha^2_i\left\|\mQ\right\|_2 C_2 \Pi_{j=i+1}^{k} \left(1-\frac{\kappa}{\lambda_{\max}(\mQ)}\alpha_j\right)
    +2\alpha_k^2\left\|\mQ\right\|_2C_2 \nonumber\\
    \leq & \exp\left(-\frac{\kappa}{\lambda_{\max}(\mQ)}\sum_{i=0}^k\alpha_i \right)\E\left[V(\vz_0)\right]
   + 2\sum_{i=0}^{k-1} \alpha^2_i\left\|\mQ\right\|_2 C_2\exp\left( -\frac{\kappa}{\lambda_{\max}(\mQ)} \sum_{j=i+1}^k \alpha_j\right)+2\alpha_k^2 \left\|\mQ \right\|_2C_2,\label{ineq:iid_general_step_size_result}
\end{align}
where the last inequality follows from the relation \(1-x\leq \exp(-x)\) for \( x\in \R\). 

\begin{enumerate}
    \item[1.] First, we will consider the case for the constant step-size. Using the fact that the step-size is constant, we can rewrite in~(\ref{ineq:iid_general_step_size_result}) into
\begin{align*}
   & \E\left[V(\vz_{k+1})\right] \\
   \leq & \exp\left(-\frac{\kappa}{\lambda_{\max}(\mQ)} k\alpha_0\right)\E\left[V(\vz_0)\right] \\
   &+ 2\sum^{k-1}_{i=0}\alpha_0^2\left\|\mQ \right\|_2 C_2 \exp\left(-\frac{\kappa}{\lambda_{\max}(\mQ)} \alpha_0(k-i)\right)+2\alpha_0^2 \left\|\mQ \right\|_2C_2\\
    \leq & \exp\left(-\frac{\kappa}{\lambda_{\max}(\mQ)} k\alpha_0\right) \E\left[V(\vz_0)\right]+ 2\alpha_0^2\left\|\mQ \right\|_2 C_2 \frac{\exp\left(-\frac{\kappa}{\lambda_{\max}(\mQ)}\alpha_0\right)}{1-\exp\left(-\frac{\kappa}{\lambda_{\max}(\mQ)}\alpha_0\right)}+2\alpha_0^2\left\|\mQ\right\|_2C_2.
\end{align*}

The second inequality follows from summation of geometric series. Since \( \exp (x) -1 \geq x\) for \(x >  0\), we have \(\frac{1}{\exp(x)-1} \leq \frac{1}{x}\), and this leads to

\begin{align*}
    \E\left[V(\vz_{k+1})\right] \leq & \exp\left(-\frac{\kappa}{\lambda_{\max}(\mQ)} k\alpha_0\right)\E\left[V(\vz_0)\right] + 2\alpha_0^2 C_2\left\|\mQ\right\|_2 \frac{1}{\frac{\kappa}{\lambda_{\max}(\mQ)} \alpha_0} + 2\alpha_0^2\left\|\mQ\right\|_2C_2\\
    =& \exp\left(-\frac{\kappa}{\lambda_{\max}(\mQ)} k\alpha_0\right)\E\left[V(\vz_0)\right] +2 \alpha_0 C_2 \left\|\mQ\right\|_2\frac{\lambda_{\max}(\mQ)}{\kappa} +2 \alpha_0^2\left\|\mQ\right\|_2C_2.
\end{align*}
\item[2.] The result for diminishing step-size becomes

\begin{align*}
    &\E\left[V(\vz_{k+1})\right] \\
    \leq & \exp\left(-\frac{\kappa}{\lambda_{\max}(\mQ)}\sum_{i=0}^k\alpha_i\right) V(\vz_0)\\
    &+ 2\sum^{k-1}_{i=0}\alpha_i^2 \left\|\mQ\right\|_2C_2 \exp\left(-\frac{\kappa}{\lambda_{\max}(\mQ)}\sum^{k-1}_{j=i+1}\alpha_j\right) +2\alpha_k^2\left\|\mQ\right\|_2C_2\\
    \leq & \exp\left(-\frac{h_1\kappa}{\lambda_{\max}(\mQ)}\log\left(\frac{k+h_2}{h_2}\right)\right) \E[V(\vz_0)]\\
    &+ 2\sum^{k-1}_{i=0} \frac{h_1^2}{(i+h_2)^2} \left\|\mQ\right\|_2C_2  \exp\left(-\frac{h_1\kappa}{\lambda_{\max}(\mQ)} \log\left(\frac{k-1+h_2}{i+1+h_2}\right) \right)
    +2\alpha_k^2\left\|\mQ\right\|_2C_2\\
    \leq & \left(\frac{h_2}{k+h_2}\right)^{\frac{h_1\kappa}{\lambda_{\max}(\mQ)}}V(\vz_0)
    +2\sum^{k-1}_{i=0} \frac{h_1^2}{(i+h_2)^2}\left\|\mQ\right\|_2C_2 \left(\frac{i+1+h_2}{k-1+h_2}\right)^{\frac{h_1\kappa}{\lambda_{\max}(\mQ)}} +2\alpha_k^2\left\|\mQ\right\|_2C_2 ,
\end{align*}

The second inequality follows from \(\int^k_{t=0} \frac{h_1}{t+h_2}dt \leq \sum^k_{i=0}\alpha_i\). From the choice of step-size, we have \(\frac{h_1\kappa}{\lambda_{\max}(\mQ)}\geq 2\), which leads to

\begin{align*}
        \E\left[V(\vz_{k+1})\right] \leq &\left(\frac{h_2}{k+h_2}\right)^{\frac{h_1\kappa}{\lambda_{\max}(\mQ)}}V(\vz_0)\\
        &+2\left\|\mQ\right\|_2C_2\sum^{k-1}_{i=0}\frac{h_1^2}{(i+h_2)^2} \left(\frac{i+1+h_2}{k-1+h_2}\right)^{\frac{h_1\kappa}{\lambda_{\max}(\mQ)}}+2\alpha_k^2\left\|\mQ\right\|_2C_2 \\
        \leq & \left(\frac{h_2}{k+h_2}\right)^{\frac{h_1\kappa}{\lambda_{\max}(\mQ)}}V(\vz_0)\\
        &+\frac{2\left\|\mQ\right\|_2C_2h_1^2}{(k-1+h_2)^{\frac{h_1\kappa}{\lambda_{\max}(\mQ)}}}2^{\frac{h_1\kappa}{\lambda_{\max}(\mQ)}}\sum^{k-1}_{i=0} (i+h_2)^{\frac{h_1\kappa}{\lambda_{\max}(\mQ)}-2}  +2\alpha_k^2\left\|\mQ\right\|_2C_2 \\
        \leq & \left(\frac{h_2}{k+h_2}\right)^{\frac{h_1\kappa}{\lambda_{\max}(\mQ)}}V(\vz_0)\\
        &+\frac{2\left\|\mQ\right\|_2C_2h_1^2}{(k-1+h_2)^{\frac{h_1\kappa}{\lambda_{\max}(\mQ)}}}2^{\frac{h_1\kappa}{\lambda_{\max}(\mQ)}} \int_0^k (s+h_1)^{\frac{h_1\kappa}{\lambda_{\max}(\mQ)}-2}ds+2\alpha_k^2\left\|\mQ\right\|_2C_2\\
        \leq & \left(\frac{h_2}{k+h_2}\right)^{\frac{h_1\kappa}{\lambda_{\max}(\mQ)}}V(\vz_0)\\
        &+\frac{2\left\|\mQ\right\|_2C_2h_1^2}{(k-1+h_2)^{\frac{h_1\kappa}{\lambda_{\max}(\mQ)}}} \frac{2^{\frac{h_1\kappa}{\lambda_{\max}(\mQ)}}}{\frac{h_1\kappa}{\lambda_{\max}(\mQ)}-1} (k+h_1)^{\frac{h_1\kappa}{\lambda_{\max}(\mQ)}-1} +2\alpha_k^2\left\|\mQ\right\|_2C_2\\
        \leq & \left(\frac{h_2}{k+h_2}\right)^{\frac{h_1\kappa}{\lambda_{\max}(\mQ)}}V(\vz_0)\\
        &+\frac{2\left\|\mQ\right\|_2C_2h_1^2}{(k-1+h_2)} \frac{2^{2\frac{h_1\kappa}{\lambda_{\max}(\mQ)}}}{\frac{h_1\kappa}{\lambda_{\max}(\mQ)}-1}  +2\alpha_k^2\left\|\mQ\right\|_2C_2.
\end{align*}

The second inequality follows from the fact that \(i+h_2+1\leq 2i+2h_2\) for \( i \in \sN\). The last inequality follows from the fact that \( k+h_1\leq 2k-2+2h_2\).

\end{enumerate}

\end{proof}

\subsection{Proof of Theorem~\ref{thm:algo1}}\label{app:sec:proof:algo1_iid}

Let us prove the first item in Theorem~\ref{thm:algo1}, which is the constant step-size case. To this end, we will apply Theorem~\ref{thm:iid_general} in the Appendix Section~\ref{app:sec:iid}, and it is enough to check the conditions in Assumption~\ref{assmp:iid_general} in the Appendix Section~\ref{app:sec:iid}. Let $\vz_k:=\begin{bmatrix}
    \tilde{\vtheta}_k\\
    \bar{\mL}\bar{\mL}^{\dagger}\tilde{\vw}_k
\end{bmatrix}$. The first item in Assumption~\ref{assmp:iid_general} follows from Lemma~\ref{lem:bar_eps_theta_bound} in the Appendix Section~\ref{app:sec:iid}. That is, the constants in the first item in Assumption~\ref{assmp:iid_general} becomes
\begin{align*}
    C_1 = 16 ,\quad C_2= \frac{32NR^2_{\max}}{w^2(1-\gamma)^2},\quad E=2+2\lambda_{\max}(\bar{\mL}).
\end{align*}

The second item in Assumption~\ref{assmp:iid_general} is straightforward from the fact that \((s_k,s_k^{\prime},r_k)\) is sampled from  i.i.d. distribution.

The third item in Assumption~\ref{assmp:iid_general} is satisfied by letting \(\kappa=\min\left\{1,\eta \lambda^+_{\min}(\bar{\mL})^2\right\}/2\), which follows from Lemma~\ref{lem:algo1:lyapunov_equation_for_projected_iterate}. Therefore, from the first item in Theorem~\ref{thm:iid_general}, letting the constant step-size to satisfy
\begin{align*}
    \alpha_0 \leq & \frac{\min\left\{1,\eta \lambda^+_{\min}(\bar{\mL})^2\right\}}{4(20+8\lambda_{\max}(\bar{\mL})+4\lambda_{\max}(\bar{\mL})^2)}\frac{\lambda_{\min}(\mG)}{\lambda_{\max}(\mG)^2} ,
\end{align*}
Hence, there exists $\bar{\alpha}$ such that 
\begin{align*}
    \bar{\alpha}= & \gO\left( \frac{\min\left\{1,\eta \lambda^+_{\min}(\bar{\mL})^2\right\}  }{\left(20+8\lambda_{\max}(\bar{\mL})+4\lambda_{\max}(\bar{\mL})^2\right)\left(\frac{8+\eta+4\eta^2 \lambda_{\max}(\bar{\mL})^2}{\eta (1-\gamma)w}\right)} \right)\\
    =& \gO\left( \frac{\min\left\{1,\eta \lambda^+_{\min}(\bar{\mL})^2\right\}}{ \lambda_{\max}(\bar{\mL})^2 \left(\frac{8}{\eta}+4\eta \lambda_{\max}(\bar{\mL})^2 \right)  } (1-\gamma) w \right).
\end{align*}
Letting \(\vx_0 := \begin{bmatrix}
    \bar{\vtheta}_0-\bm{1}_N\otimes \vtheta_c\\
    \bar{\mL}\bar{\mL}^{\dagger}\left(\bar{\vw}_0-\frac{1}{\eta}\bar{\vw}_{\infty}\right)
\end{bmatrix} \), This leads to the following result for the convergence rate:

 \begin{align*}
       & \E\left[\left\|\tilde{\vtheta}_{k+1} \right\|_2^2+\left\|\bar{\mL}\bar{\mL}^{\dagger}\tilde{\vw}_{k+1} \right\|_2^2\right] \\ \leq & \frac{\lambda_{\max}(\mG)}{\lambda_{\min}(\mG)} \exp\left(-\frac{\kappa}{\lambda_{\max}(\mG)} k\alpha_0\right) \left\|\vx_0 \right\|_2^2 \\
        &+ \frac{\left\|\mG \right\|_2}{\lambda_{\min}(\mG)}2 \alpha_0 C_2 \frac{\lambda_{\max}(\mG)}{\kappa} +\frac{1}{\lambda_{\min}(\mG)}2 \alpha_0^2\left\|\mG\right\|_2C_2\\
        \leq & 4\exp\left(-\frac{\min\left\{1,\eta \lambda^+_{\min}(\bar{\mL})^2\right\} }{2\left(\frac{8+\eta+4\eta^2 \lambda_{\max}(\bar{\mL})^2}{\eta (1-\gamma)w}\right)} k \alpha_0 \right) \left\|\vx_0 \right\|_2^2  \\
        &+ 16 \alpha_0 \left( \frac{32NR^2_{\max}}{w^2(1-\gamma)^2} \right) \frac{8+\eta+4\eta^2 \lambda_{\max}(\bar{\mL})^2}{\eta (1-\gamma)w}\frac{1}{\min\{1,\eta\lambda_{\min}(\bar{\mL})^2\}}\\
        &+8\alpha_0^2   \left( \frac{32NR^2_{\max}}{w^2(1-\gamma)^2} \right)\\
        =& \gO \left( \exp\left( -(1-\gamma)w \frac{\min\{1,\eta\lambda^+_{\min}(\bar{\mL})^2\}}{\frac{8}{\eta}+4\eta \lambda_{\max}(\bar{\mL})^2} k\alpha_0\right) \left\|\vx_0 \right\|_2^2 + \alpha_0\frac{NR^2_{\max}}{w^3(1-\gamma)^3}   \frac{2+\eta^2\lambda_{\max}(\bar{\mL})^2}{\eta \min\{1,\eta\lambda_{\min}(\bar{\mL})^2\} }\right)  ,
    \end{align*}
where the second inequality follows from Lemma~\ref{lem:algo1:lyapunov_equation_for_projected_iterate}.
Dividing by the number of agents, \(N\), leads to the desired result.

Similarly we can derive the second item in Theorem~\ref{thm:algo1}, which corresponds to the diminishing step-size case. From the second item in Theorem~\ref{thm:iid_general}, the step-size parameters have the following constraints:
\begin{align*}
    h_1  & \geq \max \left\{ \frac{2\lambda_{\max}(\mG)}{\kappa}, 2  \right\}\geq  \max\left\{\frac{8+\eta+4\eta^2 \lambda_{\max}(\bar{\mL})^2}{\eta (1-\gamma)w}\frac{2}{\min\{1,\eta\lambda^+_{\min}(\bar{\mL})^2\}},2\right\}, \\
    h_2 &\geq \max\left\{2,h_1, h_1 \frac{2 \frac{8+\eta+4\eta^2 \lambda_{\max}(\bar{\mL})^2}{\eta (1-\gamma)w}\left( (2+2\lambda_{\max}(\bar{\mL}))^2+16\right) }{\min\{1,\eta\lambda^+_{\min}(\bar{\mL})^2\}}\right\}.
\end{align*}

It suffices to choose \(h_1\) and \(h_2\) to have the following order:
\begin{align*}
    h_1 =& \Theta \left( \frac{2+\eta^2\lambda_{\max}(\bar{\mL})^2}{\eta(1-\gamma)w \min\{1,\eta\lambda^+_{\min}(\bar{\mL})^2 \}} \right),\\
    h_2 =& \Theta\left( \frac{2+\eta^2\lambda_{\max}(\bar{\mL})^2}{\eta(1-\gamma)w\min\{1,\eta\lambda^+_{\min}(\bar{\mL})^2 \}} \lambda_{\max}(\bar{\mL})^2h_1 \right)\\
    =&\Theta\left( \frac{\left(2+\eta^2\lambda_{\max}(\bar{\mL})^2\right)^2 \lambda_{\max}(\bar{\mL})^2}{\eta^2(1-\gamma)^2w^2\min\{1,\eta\lambda^+_{\min}(\bar{\mL})^2 \}^2} \right).
\end{align*}

Therefore, the convergence rate becomes
\begin{align*}
    &\E\left[\left\|\tilde{\vtheta}_{k+1} \right\|_2^2+\left\|\bar{\mL}\bar{\mL}^{\dagger}\tilde{\vw}_{k+1} \right\|_2^2\right] \\
    \leq & \frac{\lambda_{\max}(\mG)}{\lambda_{\min}(\mG)} \left(\frac{h_2}{k+h_2}\right)^{2} \left\|\vx_0\right\|_2^2
    \\
    &+\frac{8h_1^2}{k-1+h_2} \frac{32NR^2_{\max}}{w^2(1-\gamma)^2}\frac{4^{\frac{h_1\kappa}{\lambda_{\max}(\mG)}}}{\frac{h_1\kappa}{\lambda_{\max}(\mG)}-1}+16\left(\frac{h_1}{k+h_2}\right)^2\frac{32NR^2_{\max}}{w^2(1-\gamma)^2}\\
    =& \gO\left( \frac{1}{k} \frac{(1+\eta^2\lambda_{\max}(\bar{\mL})^2)^2}{\eta^2 \min\{1,\eta\lambda^+_{\min}(\bar{\mL})^2 \}^2} \frac{NR^2_{\max}}{w^4(1-\gamma)^4} \right).
\end{align*}
Dividing by the number of agents, $N$, completes the proof.

%% file: app/markov_proof.tex
We will consider a general stochastic recursive model with Markovian observation samples, for $k\in\sN_0$:
\begin{align}
    \vz_{k+1} = \vz_k+\alpha_k (\mE\vz_k+\bm{\xi}(o_k;\vz_k)), \label{eq:sa_markov}
\end{align}
where \(\mE\in \R^{2Nq\times 2Nq}\), \(\vz_k\in\R^{2Nq}\) and \(\bm{\xi}(o_k;\vz_k):=\mW(o_k)\vz_k + \vw(o_k)\) for \(\mW: \gS \times   \gS\times \Pi^N_{i=1}I\to  \R^{2Nq\times 2Nq}\), where \(I\) is closed interval \([-R_{\max},R_{\max}]\) in \(\R\), and \(\vw:\gS \times   \gS\times \Pi^N_{i=1}I \to \R^{2Nq}\). We assume that the  the sequence \( \{ o_k \in \gS\times \gS \times \Pi_{i=1}^N I  \}_{k\in \sN} \) is generated by an ergodic Markov chain. The proof follows the spirit of~\citealp{srikant2019finite}. We will denote \(T \in \sN \) as the total number of iterations and the mixing time \(\tau:=\tau(\alpha_T)\) will be defined as in~(\ref{eq:mixing_time}). We first introduce a set of assumptions:

\begin{assumption}\label{assmp:markov_general_sa}
    \begin{enumerate}
        \item[1.] For any \(o\in \gS \times   \gS\times \Pi^N_{i=1}I\), we have
        \begin{align*}
            \left\|\mW(o)\right\|_2 \leq C_1,\quad \left\|\vw(o)\right\|_2\leq C_2.
        \end{align*}
\item[2.] For \(k\geq \tau\), there exists a positive constant \(\Xi\) such that  
\begin{align*}
    \left\|\E\left[\bm{\xi}(o_k;\vz_{k-\tau})\middle|\gF_{k-\tau}\right]\right\|_2 \leq \Xi \alpha_T (\left\|\vz_{k-\tau}\right\|_2+1),
\end{align*}
where \(\gF_{k-\tau}=\sigma(o_1,o_2,\dots,o_{k-\tau})\).
\item[3.] For \(k\in \sN_0\), there exists a positive definite matrix \(\mQ\in \R^{2Nq\times 2Nq}\) and a positive constant \(\kappa\) such that
\begin{align*}
   2 \vz_k^{\top}\mE \mQ \vz_k \leq -\kappa \left\|\vz_k\right\|_2.
\end{align*}
    \end{enumerate}
\end{assumption}

For simplicity of the proof, we will denote \(E_1:= C_1+E\) where $E$ is a positive constant such that $\left\|\mE\right\|_2 \leq E$. We first present several useful lemmas.

\begin{lemma}\label{lem:z_s-z_k-tau}
    \begin{enumerate}
        \item[1.] For \(k\geq \tau \) and \(k-\tau+1 \leq s \leq k -1\), using constant step-size, i.e., \(\alpha_0=\alpha_1=\cdots=\alpha_T\) such that \(\tau\alpha_0E_1 \leq \ln 2\), we have
        \begin{align*}
            \left\|\vz_{s+1}\right\|_2 \leq 2\left\|\vz_{k-\tau}\right\|_2
              + \frac{4C_2}{E_1}.
        \end{align*}
        \item[2.] For \(k\geq \tau \) and \(k-\tau+1 \leq s \leq k -1 \), using diminishing step-size, i.e., \(\alpha_t=\frac{h_1}{t+h_2}\) for \(t\in\sN_0\) such that \(\frac{\tau-1+2^{1/E_1h_1}}{2^{1/E_1h_1}-1} \leq  h_2\), we have
    \begin{align*}
        \left\| \vz_{s+1}\right\|_2 \leq  2 \left\| \vz_{k-\tau} \right\|_2
        +4C_2\tau\alpha_{k-\tau} .
    \end{align*}
    \end{enumerate}
\end{lemma}
\begin{proof}
    Applying triangle inequality to the recursion in~(\ref{eq:sa_markov}), we get
    \begin{align*}
        \left\|\vz_{s+1}\right\|_2 \leq (1+\alpha_s E_1)\left\| \vz_s \right\|_2+\alpha_s C_2.
    \end{align*}
Recursive formula leads to
    \begin{align}
        \left\|\vz_{s+1}\right\|_2 &\leq \Pi_{j=k-\tau}^s (1+\alpha_jE_1)\left\| \vz_{k-\tau} \right\|_2
        + \sum_{i=k-\tau}^{s-1}  C_2\alpha_i \Pi_{j=i+1}^s (1+\alpha_j E_1)
        +\alpha_sC_2\nonumber\\
        & \leq \exp \left(\sum^s_{i=k-\tau} \alpha_i E_1 \right)\left\|\vz_{k-\tau}\right\|_2+\sum^{s-1}_{i=k-\tau} C_2 \alpha_i \exp\left(\sum_{j=i+1}^s \alpha_j E_1\right)+\alpha_sC_2 \label{ineq:lem:z_s-z_k-tau:1},
    \end{align}
    where the last inequality follows from the relation \(1+x\leq \exp(x)\) for \(x\in \R\).

    \begin{enumerate}
        \item[1.] We will first prove the case when the step-size is constant. Using the fact that \(\alpha_0=\alpha_1=\cdots=\alpha_s\), we can rewrite~(\ref{ineq:lem:z_s-z_k-tau:1}) as follows:
        \begin{align*}
              \left\|\vz_{s+1}\right\|_2 \leq & \exp\left( \tau \alpha_0 E_1\right)\left\| \vz_{k-\tau} \right\|_2+\sum^{s-1}_{i=k-\tau} C_2\alpha_0\exp(\alpha_0 E_1 (s-i)) + \alpha_0 C_2\\
              \leq & \exp\left( \tau \alpha_0 E_1\right)\left\| \vz_{k-\tau} \right\|_2+C_2\alpha_0 \frac{\exp((\tau-1)\alpha_0E_1)}{1-\exp(-\alpha_0E_1)}+\alpha_s C_2\\
              =&  \exp\left( \tau \alpha_0 E_1\right)\left\| \vz_{k-\tau} \right\|_2+C_2\alpha_0 \frac{\exp(\tau\alpha_0E_1)}{\exp(\alpha_0E_1)-1}+\alpha_s C_2 \\
              \leq & 2\left\|\vz_{k-\tau}\right\|_2
              +C_2 \frac{2}{E_1}+\alpha_{0}C_2\\
              \leq & 2\left\|\vz_{k-\tau}\right\|_2+\frac{4C_2}{E_1}.
        \end{align*}

        The second last inequality follows from the condition on the step-size, \( \tau\alpha_0E_1 \leq \ln 2\), and the fact that \(\exp(x)  \geq x+1 \) for \(x\in\R\).

        \item[2.] We will prove the case for diminishing step-size. Plugging in \(\alpha_t=\frac{h_1}{t+h_2}\) for \(t\in \sN\) to~(\ref{ineq:lem:z_s-z_k-tau:1}), we have
            \begin{align}
                 \left\|\vz_{s+1}\right\|_2 
                   &\leq \exp\left(E_1 h_1\int^s_{k-\tau-1} \frac{1}{t+h_2} dt \right) \left\|\vz_{k-\tau}\right\|_2 \nonumber\\
                   &+ C_2 \sum^{s-1}_{i=k-\tau}\alpha_i\exp\left( E_1h_1 \int^s_{i} \frac{1}{t+h_2}dt \right) +\alpha_sC_2 \nonumber\\
                 &\leq \left( \frac{s+h_2}{k-\tau-1+h_2} \right)^{E_1h_1}\left\| \vz_{k-\tau} \right\|_2
        +C_2 \sum^{s-1}_{i=k-\tau}\alpha_i \left(\frac{s+h_2}{i+h_2} \right)^{E_1h_1}+\alpha_sC_2 \nonumber\\
        &\leq 2\left\| \vz_{k-\tau} \right\|_2
        +C_2\sum^{k-1}_{i=k-\tau} 2\alpha_i +\alpha_sC_2 \label{ineq:_<2}\\
        &\leq 2 \left\| \vz_{k-\tau} \right\|_2
        +2C_2\tau\alpha_{k-\tau} +\alpha_sC_2 \nonumber\\
        &\leq 2 \left\| \vz_{k-\tau} \right\|_2
        +4C_2\tau\alpha_{k-\tau} . \nonumber
            \end{align}
        
            The first inequality follows from the fact that \(\sum_{i=a}^b \frac{1}{t+h_2}\leq \int^b_{a-1} \frac{1}{t+h_2}dt\) for \(a,b\in\sN_0\). The inequality in~(\ref{ineq:_<2}) follows from the following relation that for \(k\geq \tau\), \(k-\tau+1 \leq s \leq k-1\) and \(k-\tau\leq i \leq s-1\), the condition \(\frac{\tau-1+2^{1/E_1h_1}}{2^{1/E_1h_1}-1} \leq  h_2\) leads to
    \begin{align*}
        &\left( \frac{s+h_2}{i+h_2} \right)^{E_1h_1} \leq  \left( \frac{s+h_2}{k-\tau-1+h_2} \right)^{E_1h_1}\leq \left( \frac{k-1+h_2}{k-\tau-1+h_2} \right)^{E_1h_1}\leq 2.
        \end{align*}
            The last inequality follows since \(\frac{k-1+h_2}{k-\tau-1+h_2} \) is decreasing function in $k$ and it suffices to satisfy the inequality when $k=\tau$. This completes the proof.
    \end{enumerate}
\end{proof}





The following lemma shows that the difference between \(\vz_k\) and \(\vz_{k-\tau}\) for \(k\geq \tau\) will not be large:

\begin{lemma}\label{lem:vz_k-vz_k_tau:diff}
\begin{enumerate}
    \item[1.] Considering constant step-size, i.e., \(\alpha_0=\alpha_1=\cdots=\alpha_T\), with \(\alpha_0\leq \frac{1}{100\tau \max\{E_1,C_2\}}\), for \(k\geq \tau\), we have
    \begin{align*}
        \left\|\vz_k-\vz_{k-\tau}\right\|_2        \leq & 4E_1\alpha_0\tau \left\|\vz_k\right\|_2+10C_2\alpha_0\tau , \\
       \left\|\vz_k-\vz_{k-\tau}\right\|_2^2 \leq &  E_1\alpha_0\tau \left\|\vz_{k}\right\|^2_2+ C_2\alpha_0\tau.
    \end{align*}
    \item[2.] Considering diminishing step-size, i.e., \(\alpha_t=\frac{h_1}{t+h_2}\) for \( t\in \sN \) such that \(\max\left\{\frac{\tau-1+2^{1/E_1h_1}}{2^{1/E_1h_1}-1}, 32\tau E_1 h_1 ,32\tau C_2 h_1 \right\}\leq  h_2\), for \(k\geq \tau\), we have
\begin{align}
    \left\| \vz_k-\vz_{k-\tau} \right\|_2 &\leq   4E_1\alpha_{k-\tau} \tau \left\| \vz_k \right\|_2+4C_2\alpha_{k-\tau}\tau , \label{lem:vz_k-vz_k_tau:diff_1}\\
     \left\| \vz_k-\vz_{k-\tau} \right\|_2^2 &\leq E_1\alpha_{k-\tau}\tau\left\|\vz_k\right\|_2^2+C_2\alpha_{k-\tau}\tau. \label{lem:vz_k-vz_k_tau:diff_2}
\end{align}
\end{enumerate}
\end{lemma}
\begin{proof}
We have the following relation:
    \begin{align}
        \left\| \vz_k-\vz_{k-\tau} \right\|_2 &\leq \sum_{i=0}^{\tau-1}\left\| \vz_{i+1+k-\tau}-\vz_{i+k-\tau} \right\| \nonumber\\
        &= \sum^{\tau-1}_{i=0} \alpha_{i+k-\tau}\left\|\mE\vz_{i+k-\tau}+\bm{\xi}(o_{i+k-\tau};\vz_{i+k-\tau})\right\|_2\nonumber\\
        &\leq \sum^{\tau-1}_{i=0}\alpha_{i+k-\tau} (E_1\left\|\vz_{i+k-\tau} \right\|_2+C_2) . \label{ineq:1:lem:vz_k-vz_k_tau:diff}
    \end{align}
    The first inequality follows from triangle inequality. The first equality follows from the update in~(\ref{eq:sa_markov}). The last inequality follows from the first item in Assumption~\ref{assmp:markov_general_sa}.
    \begin{enumerate}
        \item[1.] Considering the constant step-size, we have
        \begin{align*}
            \left\|\vz_k-\vz_{k-\tau} \right\|_2 \leq & \alpha_0\sum^{\tau-1}_{i=0} \left(E_1\left(2\left\|\vz_{k-\tau}\right\|_2
              + \frac{4C_2}{E_1}\right)+C_2\right)\\
              = & \alpha_0\sum^{\tau-1}_{i=0}\left(2E_1\left\|\vz_{k-\tau}\right\|_2+5C_2\right)\\
              = & 2 E_1 \alpha_0\tau\left\|\vz_{k-\tau}\right\|_2+5C_2\alpha_0\tau.
        \end{align*}
        The first inequality follows applying Lemma~\ref{lem:z_s-z_k-tau} to~(\ref{ineq:1:lem:vz_k-vz_k_tau:diff}). Since we have \(E_1\alpha_0\tau \leq \frac{1}{4}\), using triangle inequality we get
        \begin{align*}
            \left\|\vz_k-\vz_{k-\tau}\right\|_2 \leq &2E_1\alpha_0\tau \left\|\vz_k-\vz_{k-\tau}\right\|_2+2E_1\alpha_0\tau \left\|\vz_k\right\|_2+5C_2\alpha_0\tau,\\
               \left\|\vz_k-\vz_{k-\tau}\right\|_2        \leq & 4E_1\alpha_0\tau \left\|\vz_k\right\|_2+10C_2\alpha_0\tau.
        \end{align*}
        Moreover, using the relation \((a+b)^2\leq 2a^2+2b^2\) for \(a,b\in \R\), we have
        \begin{align*}
            \left\|\vz_k-\vz_{k-\tau}\right\|_2^2 \leq & 2( 4E_1\alpha_0\tau)^2\left\|\vz_{k-\tau}\right\|^2_2+(10C_2\alpha_0\tau)^2\\
            \leq & E_1\alpha_0\tau \left\|\vz_{k-\tau}\right\|^2_2+ C_2\alpha_0\tau.
        \end{align*}
        The last inequality follows from the step-size condition that \(\alpha_0\leq \frac{1}{100\tau \max\{E_1,C_2\}}\).
        \item[2.] Considering diminishing step-size, applying Lemma~\ref{lem:z_s-z_k-tau} to~(\ref{ineq:1:lem:vz_k-vz_k_tau:diff}), we get 
        \begin{align*}
            \left\|\vz_k-\vz_{k-\tau} \right\|_2 &\leq \alpha_{k-\tau} \sum^{\tau-1}_{i=0}\left(E_1\left( 2 \left\| \vz_{k-\tau} \right\|_2
        +4C_2\tau\alpha_{k-\tau}  \right)+C_2\right)\\
        &= \alpha_{k-\tau}\left( 2\tau E_1 \left\|\vz_{k-\tau}\right\|_2 +4E_1C_2\tau^2\alpha_{k-\tau}+C_2\tau \right)\\
        &\leq \alpha_{k-\tau}( 2\tau E_1 \left\|\vz_{k-\tau}\right\|_2+2C_2\tau)\\
        &= 2E_1\alpha_{k-\tau}\tau \left\|\vz_{k-\tau}\right\|_2 +2C_2 \alpha_{k-\tau}\tau.
        \end{align*}
        The first inequality follows from the second item in Lemma~\ref{lem:z_s-z_k-tau}.  The condition \(h_2\geq  32E_1\tau h_1\) leads to the last inequality. Moreover, since \( \alpha_{k-\tau}  \leq \frac{1}{4\tau E_1}\) for \( k \geq \tau\), we have:
\begin{align*}
    \left\|\vz_k-\vz_{k-\tau} \right\|_2 &\leq 2E_1\alpha_{k-\tau}\tau\left\|\vz_{k-\tau}-\vz_k \right\|_2 +2E_1\alpha_{k-\tau}\tau\left\|\vz_k \right\|_2+2\alpha_{k-\tau}C_2\tau\\
    &\leq 4E_1\alpha_{k-\tau} \tau \left\| \vz_k \right\|_2+4C_2\alpha_{k-\tau}\tau.
\end{align*}
The first inequality follows triangle inequality. Furthermore, using the relation \((a+b)^2\leq 2a^2+2b^2\) for \(a,b\in \R\), we have
\begin{align*}
    \left\|\vz_k-\vz_{k-\tau}\right\|^2_2 \leq &2(4E_1\alpha_{k-\tau}\tau)^2\left\|\vz_k\right\|^2_2 + 2(4C_2\alpha_{k-\tau}\tau)^2\\
    \leq & E_1\alpha_{k-\tau}\tau\left\|\vz_k\right\|_2^2+C_2\alpha_{k-\tau}\tau.
\end{align*}
The last inequality follows from the step-size condition \(\max\left\{32\tau E_1 h_1 ,32\tau C_2 h_1\right\}\leq  h_2\).
    \end{enumerate}
\end{proof}

\begin{lemma}\label{lem:markov:cross_term}
\begin{enumerate}
    \item[1.] Considering constant step-size, i.e., \(\alpha_0=\alpha_1=\cdots=\alpha_T\), with \(\alpha_0\leq \min \left\{ \frac{1}{100\tau \max\{E_1,C_2\}},\frac{C_1}{2\Xi} \right\}\), for \(k\geq \tau\), we have
    \begin{align*}
    &\E[\vz_k^{\top}\mQ(\bm{\xi}(o_k;\vz_k)) ]\\
    \leq & 
\left\|\mQ \right\|_2 \left((4\Xi+13C_1E_1+20C_1C_2+4E_1C_2)\alpha_0\tau \E\left[\left\|\vz_k\right\|^2_2\right]\right.\\
    &\left.+\left(25C_1C_2
+ 10C_2^2+2\Xi +4E_1C_2 \right)  \alpha_0\tau \right).
    \end{align*}
    \item[2.] Considering diminishing step-size, i.e., \(\alpha_t=\frac{h_1}{t+h_2}\) for \( t\in \sN \) such that \(\max\left\{\frac{\tau-1+2^{1/E_1h_1}}{2^{1/E_1h_1}-1}, 32\tau E_1 h_1 ,32\tau C_2 h_1,\frac{\Xi h_1}{2C_1} \right\}\leq  h_2\), for \(k\geq \tau\), we have
    \begin{align*}
&\E[\vz_k^{\top}\mQ(\bm{\xi}(o_k;\vz_k)) ]\\
\leq & 
\left\|\mQ\right\|_2 \left(  (4\Xi+ 13E_1C_1+8C_1C_2+4C_2E_1  ) \E\left[\left\|\vz_k\right\|^2_2\right]\right.\\
    & \left.+ (13C_1C_2+4C_2^2 +2\Xi+4C_2E_1 )  \right)\alpha_{k-\tau}\tau. 
\end{align*}
\end{enumerate}
\end{lemma}

\begin{proof}
    Following the spirit of~\citealp{srikant2019finite} we can decompose the cross term in to follows four terms:
    \begin{align*}
        &\E[\vz_k^{\top}\mQ (\vz_{k+1}-\vz_k-\alpha_k \mE\vz_k)]\\
       =&\alpha_k\E[\vz_k^{\top}\mQ (\vw(o_k)+\mW(o_k)\vz_k)) ] \\
       =&\alpha_k \left(\underbrace{\E[\vz_{k-\tau}^{\top}\mQ (\vw(o_k)+\mW(o_k)\vz_{k-\tau})]}_{I_1}
       + \underbrace{\E[(\vz_k-\vz_{k-\tau})^{\top}\mQ(\vw(o_k)+\mW(o_k)(\vz_k-\vz_{k-\tau}))]}_{I_2}\right.\\
       &\left.+ \underbrace{\E[(\vz_k-\vz_{k-\tau})^{\top}\mQ\mW(o_k)\vz_{k-\tau}]}_{I_3} + \underbrace{\E[\vz_{k-\tau}^{\top}\mQ\mW(o_k)(\vz_k-\vz_{k-\tau})]}_{I_4}\right).
    \end{align*}

The term $I_1$ can be bounded from the second item in Assumption~\ref{assmp:markov_general_sa}, which uses the geometric mixing property of the Markov chain.
\begin{align*}
    \E\left[\vz_{k-\tau}^{\top}\mQ (\vw(o_k)+\mW(o_k)\vz_{k-\tau}) \right]  =& \E\left[\vz_{k-\tau}^{\top}\mQ \E\left[(\mW(o_k)\vz_{k-\tau}+\vw(o_k)) \middle|\gF_{k-\tau}\right]\right]\\
    \leq & \E\left[ \left\|\vz_{k-\tau}\right\|_2\left\|\mQ\right\|_2  \left\| \E\left[(\mW(o_k)\vz_{k-\tau}+\vw(o_k)) \middle|\gF_{k-\tau}\right]\right\|_2\right] \\
    \leq & \E\left[\left\|\vz_{k-\tau}\right\|_2\left\|\mQ\right\|_2  \Xi \alpha_T (\left\|\vz_{k-\tau}\right\|_2+1)\right]\\
    \leq & \left\|\mQ\right\|_2  \Xi \alpha_T \left( 2\E\left[\left\| \vz_{k-\tau}\right\|_2^2\right]+1 \right) \\
    \leq &\left\|\mQ\right\|_2  \Xi \alpha_T \E\left[(4\left\|\vz_k-\vz_{k-\tau}\right\|_2^2+4\left\|\vz_k\right\|_2^2+2) \right].
\end{align*}
The first inequality follows from Cauchy-Schwartz inequality. The second inequality follows from the second item in Assumption~\ref{assmp:markov_general_sa}. The third inequality follows from the relation $a\leq a^2+1$ for $a\in\R$. The last inequality follows from the relation \((a+b)^2\leq 2a^2+2b^2\) for \(a,b\in \R\).


The term $I_2$ can be bounded as follows:
\begin{align*}
    &\E\left[(\vz_k-\vz_{k-\tau})^{\top}\mQ(\bm{\xi}(o_k;\vz_k-\vz_{k-\tau}))\right]\\
    \leq & \E\left[ \left\| \vz_k-\vz_{k-\tau}\right\|_2 \left\| \mQ \right\|_2 \left\| \bm{\xi}(o_k;\vz_k-\vz_{k-\tau}) \right\|_2 \right]\\
    \leq &\E\left[\left\| \vz_k-\vz_{k-\tau}\right\|_2 \left\| \mQ \right\|_2\left(C_1\left\|\vz_k-\vz_{k-\tau} \right\|_2+C_2 \right) \right]\\
    = & \E\left[ \left\|\mQ  \right\|_2C_1 \left\|\vz_k-\vz_{k-\tau} \right\|_2^2 +\left\| \mQ \right\|_2 C_2\left\| \vz_k-\vz_{k-\tau}\right\|_2  \right].
\end{align*}
The first inequality follows from Cauchy-Schwartz inequality. The second inequality follows from the first item in Assumption~\ref{assmp:markov_general_sa}.


The term $I_3$ can be bounded as follows:
\begin{align*}
      & \E\left[(\vz_k-\vz_{k-\tau})^{\top}\mQ \mW(o_k)\vz_{k-\tau}\right] \\
     \leq & \E\left[ \left\| \vz_k-\vz_{k-\tau} \right\|_2\left\|\mQ\right\|_2\left( C_1\left\|\vz_{k-\tau} \right\|_2\right)\right]\\
     \leq & \E\left[C_1\left\|\mQ \right\|_2\left\| \vz_k-\vz_{k-\tau} \right\|_2^2  \right]
     +\E\left[ C_1 \left\|\mQ \right\|_2\left\|\vz_k-\vz_{k-\tau}\right\|_2\left\| \vz_k \right\|_2 \right]. 
\end{align*} 
The first inequality follows from Cauchy-Schwartz inequality and the first item in Assumption~\ref{assmp:markov_general_sa}. The second inequality follows from triangle inequality.

The term $I_4$ can be bounded as 
\begin{align*}
    &\E\left[\vz_{k-\tau}^{\top}\mQ\mW(o_k)(\vz_k-\vz_{k-\tau}))\right] \\
    \leq & \E\left[ \left\|\vz_{k-\tau} \right\|_2 \left\| \mQ \right\|_2 (C_1\left\|\vz_k-\vz_{k-\tau}\right\|_2)\right]\\
    \leq &\E\left[C_1\left\|\mQ\right\|_2 (\left\|\vz_k-\vz_{k-\tau}\right\|_2^2+ \left\|\vz_k-\vz_{k-\tau}\right\|_2\left\|\vz_k\right\|_2) \right].
\end{align*}
The first inequality follows from Cauchy-Schwartz inequality and the first item in Assumption~\ref{assmp:markov_general_sa}.


Collecting the terms to bound $I_1,I_2,I_3$ and $I_4$, we get
\begin{align}
    &\E[\vz_k^{\top}\mQ(\bm{\xi}(o_k;\vz_k)) ] \nonumber\\
    \leq & \left\|\mQ\right\|_2\left(4\Xi\alpha_T \E\left[ \left\|\vz_k\right\|_2^2 \right] +\left(3C_1+4\Xi\alpha_T\right) \E\left[\left\|\vz_k-\vz_{k-\tau}\right\|^2_2\right]\right. \nonumber\\
    &\left.+ 2C_1 \E\left[ \left\|\vz_k-\vz_{k-\tau}\right\|_2\left\|\vz_k \right\|_2\right]+C_2\E\left[\left\|\vz_k-\vz_{k-\tau}\right\|_2\right]+2\Xi \alpha_T\right)\nonumber \\
    \leq & \left\|\mQ\right\|_2\left( 4\Xi\alpha_T\E\left[ \left\|\vz_k\right\|_2^2 \right] + 5C_1 \E\left[\left\|\vz_k-\vz_{k-\tau}\right\|^2_2\right]\right. \label{ineq:1:cross_term_bound}\\
    &\left.+ 2C_1 \E\left[ \left\|\vz_k-\vz_{k-\tau}\right\|_2\left\|\vz_k \right\|_2\right]+C_2\E\left[\left\|\vz_k-\vz_{k-\tau}\right\|_2\right]+2 \Xi \alpha_T\right) .\label{ineq:2:cross_term_bound}
\end{align}
The last inequality follows from the step-size condition that  \(2\Xi\alpha_T\leq C_1\).
\begin{enumerate}
    \item[1.]  For constant step-size case, we have
\begin{align*}
        &\E[\vz_k^{\top}\mQ(\bm{\xi}(o_k;\vz_k)) ]\\
    \leq & \left\|\mQ\right\|_2\left( 4\Xi\alpha_0\E\left[ \left\|\vz_k\right\|_2^2 \right] +5C_1 \left(\E\left[ E_1\alpha_0\tau \left\|\vz_{k}\right\|^2_2+ C_2\alpha_0\tau\right]\right) \right.\\
    &+ 2C_1\left(\E\left[
    4E_1\alpha_0\tau \left\|\vz_k\right\|_2^2+10C_2\alpha_0\tau \left\|\vz_k\right\|_2\right]\right)\\
    &+\left.C_2\left(\E\left[ 4 E_1\alpha_0\tau\left\|\vz_k\right\|_2+10C_2\alpha_0\tau )\right] \right)+2 \Xi \alpha_0\right)\\
    \leq &\left\|\mQ\right\|_2\left((4\Xi\alpha_0+13 C_1E_1\alpha_0\tau) \E\left[\left\|\vz_k\right\|^2_2\right]\right.\\
    &\left.+\left(20C_1C_2+4E_1C_2\right)\alpha_0\tau\E\left[\left\|\vz_k\right\|_2\right]+ \left(5C_1C_2
+ 10C_2^2+2\Xi \right)  \alpha_0\tau \right)\\
    \leq &\left\|\mQ \right\|_2 \left((4\Xi+13C_1E_1+20C_1C_2+4E_1C_2)\alpha_0\tau \E\left[\left\|\vz_k\right\|^2_2\right]\right.\\
    &\left.+\left(25C_1C_2
+ 10C_2^2+2\Xi +4E_1C_2 \right)  \alpha_0\tau \right).
\end{align*}
The first inequality follows from applying Lemma~\ref{lem:vz_k-vz_k_tau:diff} to~(\ref{ineq:1:cross_term_bound}) and~(\ref{ineq:2:cross_term_bound}). The last inequality follows from the relation \(a\leq a^2+1\) for \(a\in\R\).
    \item[2.] Considering diminishing step-size, we get
    \begin{align*}
        &\E[\vz_k^{\top}\mQ(\bm{\xi}(o_k;\vz_k)) ]\\
    \leq & \left\|\mQ\right\|_2\left(4\Xi \alpha_T \E\left[ \left\|\vz_k\right\|_2^2 \right]
    + 5C_1 \left(\E\left[ E_1\alpha_{k-\tau}\tau\left\|\vz_k\right\|_2^2+C_2\alpha_{k-\tau}\tau  \right]\right) \right.\\
    &+ 2C_1 \left( \E\left[ 4E_1\alpha_{k-\tau}\tau  \left\|\vz_k\right\|_2^2+4C_2\alpha_{k-\tau}\tau\left\|\vz_k\right\|_2 \right]\right)\\
    &\left.+  C_2 ( 4E_1 \alpha_{k-\tau}\tau \left\| \vz_k \right\|_2+4C_2\alpha_{k-\tau}\tau)+2\Xi\alpha_T\right)\\
    \leq & \left\|\mQ\right\|_2\left( (4\Xi\alpha_T+ 13E_1C_1\alpha_{k-\tau}\tau  )\E\left[\left\|\vz_k\right\|^2_2\right]\right.\\
    &+ (8C_1C_2 + 4C_2E_1)\alpha_{k-\tau}\tau\E\left[\left\|\vz_k\right\|_2\right]\\
    & \left.+  (5C_1C_2\alpha_{k-\tau}\tau+4C_2^2 \alpha_{k-\tau}\tau +2\Xi\alpha_T )\right)\\
    \leq &\left\|\mQ\right\|_2 \left(  (4\Xi+ 13E_1C_1+8C_1C_2+4C_2E_1  )\alpha_{k-\tau}\tau \E\left[\left\|\vz_k\right\|^2_2\right]\right.\\
    & \left.+ (13C_1C_2+4C_2^2 +2\Xi+4C_2E_1 )\alpha_{k-\tau}\tau  \right).
    \end{align*}
    The first inequality follows from applying Lemma~\ref{lem:vz_k-vz_k_tau:diff} to~(\ref{ineq:1:cross_term_bound}) and~(\ref{ineq:2:cross_term_bound}). The last inequality follows from the relation \(a\leq a^2+1\) for \(a\in\R\).
\end{enumerate}



\end{proof}

\begin{lemma}\label{lem:markov:vz_k+1-vz_k}
For \(k\in \sN_0\), we have
    \begin{align*}
        (\vz_{k+1}-\vz_k)^{\top}\mQ(\vz_{k+1}-\vz_k) \leq  2\alpha_k^2\left\|\mQ\right\|_2 (E_1^2\left\|\vz_k\right\|_2^2+C_2^2).
    \end{align*}
\end{lemma}
\begin{proof}
We have
    \begin{align*}
         (\vz_{k+1}-\vz_k)^{\top}\mQ(\vz_{k+1}-\vz_k) \leq &\left\|\mQ\right\|_2\left\| \vz_{k+1}-\vz_k\right\|_2^2\\
         =& \left\|\mQ\right\|_2\left\| \alpha_k \mE\vz_k+\alpha_k\bm{\xi}(o_k;\vz_k) \right\|_2^2\\
         \leq & \alpha_k^2\left\|\mQ\right\|_2 (E_1\left\|\vz_k\right\|_2+C_2)^2\\
         \leq & 2\alpha_k^2\left\|\mQ\right\|_2 (E_1^2\left\|\vz_k\right\|_2^2+C_2^2).
    \end{align*}
    The first inequality follows from Cauchy-Schwartz inequality, and the second inequaltiy follows from the relation \(\left\|\va+\vb\right\|^2_2\leq2\left\|\va\right\|^2_2+2\left\|\vb\right\|^2_2\) for \(\va,\vb\in\R^{2Nq}\).
\end{proof}

\begin{theorem}\label{thm:markovian_general}
\begin{enumerate}
    \item[1.] Considering constant step-size, i.e., \(\alpha_0=\alpha_1=\cdots=\alpha_T\), with \(\alpha_0\leq \min\left\{\frac{1}{100\tau\max\{ E_1,C_2\} },\frac{C_1}{2\Xi}, \frac{\kappa\lambda_{\min}(\mQ) }{(4E_1^2+4K_1\tau)\lambda_{\max}(\mQ)\left\|\mQ\right\|_2}  \right\}\), we have, for \( \tau \leq k  \), 
\begin{align*}
    \E\left[\left\|\vz_{k+1}\right\|_2^2\right] \leq & \frac{\lambda_{\max}(\mQ)}{\lambda_{\min}(\mQ)}\exp\left(-\alpha_0\frac{\kappa}{2\lambda_{\max}(\mQ)} (k-\tau+1)\right)\left( 2\left\|\vz_{0}\right\|_2^2
              + \frac{4C_2}{E_1}\right)\\
     &+\frac{2\left\|  \mQ \right\|_2(C_2^2+K_2\tau)}{\lambda_{\min}(\mQ)}\left( \alpha_0\frac{2\lambda_{\max}(\mQ)}{\kappa}+\alpha_0^2 \right),
\end{align*}
    where
\begin{align*}
    K_1:=&4\Xi+13C_1E_1+20C_1C_2+4E_1C_2,\quad K_2:= 25C_1C_2
+ 10C_2^2+2\Xi +4E_1C_2.
\end{align*}
    \item[2.] Considering diminishing step-size, i.e., \(\alpha_t=\frac{h_1}{t+h_2}\) for \( t\in \sN \) such that \(\max\left\{\frac{\tau-1+2^{1/E_1h_1}}{2^{1/E_1h_1}-1}, 32\tau E_1 h_1 ,32\tau C_2 h_1,\frac{\Xi h_1}{2C_1} ,h_1 \frac{2\left\|\mQ\right\|_2\lambda_{\max}(\mQ) (2E_1^2+2L_1\tau)}{\kappa\lambda_{\min}(\mQ)}\right\}\leq  h_2   \) and \(\max\left\{\frac{4\lambda_{\max}(\mQ)}{\kappa},\frac{2}{E_1}\right\}\leq h_1\), for \( \tau\leq k \leq T\), we have
\begin{align*}
    \E\left[\left\|\vz_{k+1}\right\|_2^2\right] 
    \leq & \frac{\lambda_{\max}(\mQ)}{\lambda_{\min}(\mQ)} \left(\frac{\tau+h_2}{k+h_2} \right)^{h_1\frac{\kappa}{2\lambda_{\max}(\mQ)}} \left(2 \left\| \vz_{0} \right\|_2
        +4C_2\tau\alpha_{0} \right)\\
&+\frac{1}{\lambda_{\min}(\mQ)}\frac{16\left\|\mQ \right\|_2(L_2\tau+C_2^2)h_1^2}{k-1+h_2} \frac{2^{h_1\frac{\kappa}{2\lambda_{\max}(\mQ)}-1}}{h_1\frac{\kappa}{2\lambda_{\max}(\mQ)}-1} \\
&+\frac{2 \left\|  \mQ \right\|_2}{\lambda_{\min}(\mQ)}\left(L_2\tau\alpha_{k-\tau} \alpha_k+ C_2^2\alpha_k^2\right) , 
\end{align*}
where
\begin{align*}
    L_1:=& 4\Xi+ 13E_1C_1+8C_1C_2+4C_2E_1,\quad  L_2:=  13C_1C_2+4C_2^2 +2\Xi+4C_2E_1.
\end{align*}
\end{enumerate}
\end{theorem}

\begin{proof}
    Let \(V(\vz)=\vz^{\top}\mQ\vz\) for \(\vz\in \R^{2Nq}\). From simple algebraic manipulation in~\cite{srikant2019finite}, we have the following decomposition:
\begin{align}
 & \E\left[V(\vz_{k+1})-V(\vz_k)\right] \nonumber   \\
=& \E\left[ (\vz_{k+1}-\vz_k)^{\top}\mQ (\vz_{k+1}-\vz_k) \right]+\E\left[ 2\vz_k^{\top}\mQ \vz_{k+1} \right] -2 \E\left[V(\vz_k)\right] \nonumber \\
=& \E\left[ (\vz_{k+1}-\vz_k)^{\top}\mQ (\vz_{k+1}-\vz_k) \right]+\E \left[ 2\vz_k^{\top}\mQ (\vz_{k+1}-\vz_k)\right] \nonumber\\
=&\underbrace{\E\left[ (\vz_{k+1}-\vz_k)^{\top}\mQ (\vz_{k+1}-\vz_k ) \right]}_{I_1}+\underbrace{\E\left[2\vz_k^{\top}\mQ(\vz_{k+1}-\vz_k-\alpha_k\mE \vz_k) \right]}_{I_2}+\underbrace{2\alpha_k \E\left[ \vz_k^{\top} \mQ \mE \vz_k \right]}_{I_3}.  \label{eq:markov_decomposition}
\end{align}

\begin{enumerate}
    \item[1.] We will first consider the case using constant step-size. The term $I_1$ can be bounded using Lemma~\ref{lem:markov:vz_k+1-vz_k}, the term $I_2$ can be bounded using the first item in Lemma~\ref{lem:markov:cross_term}, and the bound on $I_3$ follows from the third item in Assumption~\ref{assmp:markov_general_sa}, which yields
\begin{align*}
    \E[V(\vz_{k+1})-V(\vz_k)] \leq & 2\alpha_0^2\left\|\mQ\right\|_2 \left(E_1^2 \E\left[\left\|\vz_k\right\|_2^2\right]+C_2^2\right)\\
    &+ 2\alpha_0\left\| \mQ \right\|_2 K_1  \alpha_0\tau\E\left[\left\| \vz_k \right\|_2^2\right] \\
&+
2\alpha_0\left\| \mQ\right\|_2 K_2 \tau \alpha_0\\
&-\alpha_0 \kappa \E\left[\left\|\vz_k\right\|_2^2\right].
\end{align*}
Considering that \(\lambda_{\min}(\mQ)\left\|\vz_k\right\|_2^2 \leq \left\|\vz_k\right\|_{\mQ}^2 \leq\lambda_{\max}(\mQ) \left\|\vz_k\right\|_2^2\), we get
\begin{align*}
    \E\left[V(\vz_{k+1})-V(\vz_k)\right] \leq & \frac{\left\|\mQ \right\|_2}{\lambda_{\min}(\mQ)} \left( 2E_1^2\alpha_0^2+2K_1\alpha_0^2\tau\right) \E\left[V(\vz_k)\right]-\alpha_0\frac{\kappa}{\lambda_{\max}(\mQ)}\E\left[V(\vz_k)\right]\\
    & +   2\left\|  \mQ \right\|_2(C_2^2+K_2\tau)\alpha_0^2 .
\end{align*}

The condition on the step-size that
\begin{align*}
    &\frac{\left\|\mQ \right\|_2}{\lambda_{\min}(\mQ)} \left( 2E_1^2\alpha_0^2+2K_1\alpha_0^2\tau\right)-\alpha_0\frac{\kappa}{\lambda_{\max}(\mQ)} \leq -\alpha_0\frac{\kappa}{2\lambda_{\max}(\mQ)}\\
    \iff & \alpha_0 \leq \frac{\kappa\lambda_{\min}(\mQ) }{(4E_1^2+4K_1\tau)\lambda_{\max}(\mQ)\left\|\mQ\right\|_2},
\end{align*}
leads to
\begin{align*}
    \E\left[V(\vz_{k+1})\right] \leq \left(1-\alpha_0\frac{\kappa}{2\lambda_{\max}(\mQ)}\right)\E\left[V(\vz_k)\right]
    + 2\left\|  \mQ \right\|_2(C_2^2+K_2\tau)\alpha_0^2.
\end{align*}

Recursively expanding the terms, we get

\begin{align*}
    & \E\left[V(\vz_{k+1})\right] \\
    \leq & \Pi_{i=\tau}^k \left(1-\alpha_0\frac{\kappa}{2\lambda_{\max}(\mQ)}\right)\E\left[V(\vz_{\tau})\right]\\
     &+\sum^{k-1}_{i=\tau} 2\left\|  \mQ \right\|_2(C_2^2+K_2\tau)\alpha_0^2  \Pi_{j=i+1}^{k} \left(1-\alpha_0\frac{\kappa}{2\lambda_{\max}(\mQ)}\right)
     +2\left\|  \mQ \right\|_2(C_2^2+K_2\tau)\alpha_0^2 \\
     \leq & \exp\left(-\alpha_0\frac{\kappa}{2\lambda_{\max}(\mQ)} (k-\tau+1)\right)\E\left[V(\vz_{\tau})\right]\\
     &+\sum^{k-1}_{i=\tau} 2\left\|  \mQ \right\|_2(C_2^2+K_2\tau)\alpha_0^2 \exp\left(-\alpha_0\frac{\kappa}{2\lambda_{\max}(\mQ)}(k-i)\right)+2\left\|  \mQ \right\|_2(C_2^2+K_2\tau)\alpha_0^2\\
     \leq & \exp\left(-\alpha_0\frac{\kappa}{2\lambda_{\max}(\mQ)} (k-\tau+1)\right)\E\left[V(\vz_{\tau})\right]\\
     &+2\left\|  \mQ \right\|_2(C_2^2+K_2\tau)\alpha_0^2 \frac{\exp\left(-\alpha_0\frac{\kappa}{2\lambda_{\max}(\mQ)}\right)}{1-\exp\left(-\alpha_0\frac{\kappa}{2\lambda_{\max}(\mQ)}\right)}+2\left\|  \mQ \right\|_2(C_2^2+K_2\tau)\alpha_0^2\\
     =&\exp\left(-\alpha_0\frac{\kappa}{2\lambda_{\max}(\mQ)} (k-\tau+1)\right)\E\left[V(\vz_{\tau})\right]\\
     &+2\left\|  \mQ \right\|_2(C_2^2+K_2\tau)\alpha_0^2 \frac{1}{\exp\left(\alpha_0\frac{\kappa}{2\lambda_{\max}(\mQ)}\right)-1}+2\left\|  \mQ \right\|_2(C_2^2+K_2\tau)\alpha_0^2\\
     \leq &\exp\left(-\alpha_0\frac{\kappa}{2\lambda_{\max}(\mQ)} (k-\tau+1)\right)\E\left[V(\vz_{\tau})\right]\\
     &+2\left\|  \mQ \right\|_2(C_2^2+K_2\tau)\alpha_0^2 \frac{2\lambda_{\max}(\mQ)}{\alpha_0\kappa} +2\left\|  \mQ \right\|_2(C_2^2+K_2\tau)\alpha_0^2\\
     =&\exp\left(-\alpha_0\frac{\kappa}{2\lambda_{\max}(\mQ)} (k-\tau+1)\right)\E\left[V(\vz_{\tau})\right]\\
     &+ 4\left\|  \mQ \right\|_2\left(C_2^2 + K_2\tau\right) \alpha_0\frac{\lambda_{\max}(\mQ)}{\kappa}+2\left\|  \mQ \right\|_2(C_2^2+K_2\tau)\alpha_0^2.
    \end{align*}

The second inequality follows from the fact that \(1-x \leq  \exp(-x)\) for \(x\in\R\). From the first item in Lemma~\ref{lem:z_s-z_k-tau}, we can bound \(\E\left[V(\vz_{\tau})\right]\), which leads to

\begin{align*}
    \E\left[\left\|\vz_{k+1}\right\|_2^2\right] \leq & \frac{\lambda_{\max}(\mQ)}{\lambda_{\min}(\mQ)}\exp\left(-\alpha_0\frac{\kappa}{2\lambda_{\max}(\mQ)} (k-\tau+1)\right)\left( 2\left\|\vz_{0}\right\|_2^2
              + \frac{4C_2}{E_1}\right)\\
     &+\frac{2\left\|  \mQ \right\|_2(C_2^2+K_2\tau)}{\lambda_{\min}(\mQ)}\left( \alpha_0\frac{2\lambda_{\max}(\mQ)}{\kappa}+\alpha_0^2 \right).
\end{align*}

    \item[2.] We will now consider the case using diminishing step-size. In~(\ref{eq:markov_decomposition}), the term $I_1$ can be bounded using Lemma~\ref{lem:markov:vz_k+1-vz_k}, the term $I_2$ can be bounded using the second item in Lemma~\ref{lem:markov:cross_term}, and the bound on $I_3$ follows from the third item in Assumption~\ref{assmp:markov_general_sa}, which yields
\begin{align*}
    \E[V(\vz_{k+1})-V(\vz_k)] \leq & 2\left\|\mQ\right\|_2 \alpha_k^2\left(E_1^2 \E\left[\left\|\vz_k\right\|_2^2\right]+C_2^2\right)\\
    &+ 2\alpha_k\left\| \mQ \right\|_2 L_1  \alpha_{k-\tau}\tau\E\left[\left\| \vz_k \right\|_2^2\right] \\
&+
2\alpha_k\left\| \mQ\right\|_2 L_2  \alpha_{k-\tau}\tau\\
&-\alpha_k \kappa \E\left[\left\|\vz_k\right\|_2^2\right],
\end{align*}
where
\begin{align*}
    L_1:=& 4\Xi+ 13E_1C_1+8C_1C_2+4C_2E_1,\quad  L_2:=  13C_1C_2+4C_2^2 +2\Xi+4C_2E_1.
\end{align*}

Considering that \(\lambda_{\min}(\mQ)\left\|\vz_k\right\|_2^2 \leq \left\|\vz_k\right\|_{\mQ}^2 \leq\lambda_{\max}(\mQ) \left\|\vz_k\right\|_2^2\), we get
\begin{align}
    \E\left[V(\vz_{k+1})-V(\vz_k)\right] \leq & \frac{\left\|\mQ \right\|_2}{\lambda_{\min}(\mQ)} \left( 2E_1^2\alpha_k^2+2L_1\alpha_k\alpha_{k-\tau}\tau\right) \E\left[V(\vz_k)\right]-\alpha_k\frac{\kappa}{\lambda_{\max}(\mQ)}\E\left[V(\vz_k)\right] \label{ineq:marovian:nesterov_decrease}\\
    & +   2\left\|  \mQ \right\|_2C_2^2 \alpha_k^2 +2\left\|  \mQ \right\|_2 L_2\tau \alpha_k\alpha_{k-\tau}. \nonumber
\end{align}
The condition on the step-size that \(h_2 \geq h_1 \frac{2\left\|\mQ\right\|_2\lambda_{\max}(\mQ) (2E_1^2+2L_1\tau)}{\kappa\lambda_{\min}(\mQ)} \) implies 
\begin{align*}
    \frac{\left\|\mQ \right\|_2}{\lambda_{\min}(\mQ)} \left(2 E_1^2\alpha_k^2+2L_1\alpha_k\alpha_{k-\tau}\tau\right)-\alpha_k\frac{\kappa}{\lambda_{\max}(\mQ)} \leq -\alpha_k\frac{\kappa}{2\lambda_{\max}(\mQ)}.
\end{align*}

Applying the above relation to~(\ref{ineq:marovian:nesterov_decrease}) results to

\begin{align*}
    \E\left[ V(\vz_{k+1}) \right] \leq \E\left[\left(1-\frac{\kappa}{2\lambda_{\max}(\mQ)}\alpha_k \right)V(\vz_k)\right]+2\left\|  \mQ \right\|_2C_2^2 \alpha_k^2 +2 \left\|  \mQ \right\|_2 L_2\tau \alpha_k\alpha_{k-\tau}.
\end{align*}

Recursively expanding the terms, we get
    \begin{align}
        &\E\left[ V(\vz_{k+1}) \right] \nonumber\\
       \leq &\left(1-\frac{\kappa}{2\lambda_{\max}(\mQ)}\alpha_k \right)\E\left[V(\vz_k)\right]+ 2 \left\|  \mQ \right\|_2L_2\tau\alpha_{k-\tau} \alpha_k+ 2\left\|\mQ\right\|_2C_2^2\alpha_k^2 \nonumber \\
        \leq & \Pi_{i=\tau}^k \left(1-\frac{\kappa}{2\lambda_{\max}(\mQ)}\alpha_i\right) \E\left[V(\vz_{\tau})\right]+2\left\|\mQ\right\|_2\sum^{k-1}_{i=\tau} (L_2\tau\alpha_{i-\tau}\alpha_i+ \alpha_i^2 C^2_2) \Pi^{k-1}_{j=i+1}\left(1-\frac{\kappa}{2\lambda_{\max}(\mQ)}\alpha_j\right) \nonumber\\
&+2 \left\|  \mQ \right\|_2L_2\tau\alpha_{k-\tau} \alpha_k+ 2\left\|\mQ\right\|_2C_2^2\alpha_k^2 \nonumber\\
\leq & \exp\left(-\frac{\kappa}{2\lambda_{\max}(\mQ)}\sum^{k}_{i=\tau}\alpha_i\right)\E\left[V(\vz_{\tau})\right]
+ 2\left\|\mQ\right\|_2 \sum^{k-1}_{i=\tau} (L_2\tau\alpha_{i-\tau}\alpha_i+ \alpha_i^2 C^2_2) \exp\left(-\frac{\kappa}{2\lambda_{\max}(\mQ)}\sum^{k-1}_{j=i+1}\alpha_j \right) \nonumber\\
&+2 \left\|  \mQ \right\|_2L_2\tau\alpha_{k-\tau} \alpha_k+ 2\left\|\mQ\right\|_2C_2^2\alpha_k^2 \nonumber\\
\leq & \left(\frac{\tau+h_2}{k+h_2} \right)^{h_1\frac{\kappa}{2\lambda_{\max}(\mQ)}}\E\left[ V(\vz_{\tau})\right]
+2\left\|\mQ\right\|_2 \sum^{k-1}_{i=\tau}(L_2\tau\alpha_{i-\tau}\alpha_i+ \alpha_i^2 C^2_2)\left(\frac{i+1+h_2}{k-1+h_2}\right)^{h_1\frac{\kappa}{2\lambda_{\max}(\mQ)}} \nonumber\\
&+2 \left\|  \mQ \right\|_2L_2\tau\alpha_{k-\tau} \alpha_k+ 2\left\|\mQ\right\|_2C_2^2\alpha_k^2 \nonumber\\
\leq & \left(\frac{\tau+h_2}{k+h_2} \right)^{h_1\frac{\kappa}{2\lambda_{\max}(\mQ)}} \E\left[V(\vz_{\tau})\right]
+\frac{2\left\|\mQ \right\|_2(L_2\tau+C_2^2)h_1^2}{(k-1+h_2)^{h_1\frac{\kappa}{2\lambda_{\max}(\mQ)}}}\sum^{k-1}_{i=\tau} 8\left(i+1+h_2 \right)^{h_1\frac{\kappa}{2\lambda_{\max}(\mQ)}-2} \label{ineq:decreasing_1} \\
&+2 \left\|  \mQ \right\|_2L_2\tau\alpha_{k-\tau} \alpha_k+ 2\left\|\mQ\right\|_2C_2^2\alpha_k^2 \nonumber\\
\leq & \left(\frac{\tau+h_2}{k+h_2} \right)^{h_1\frac{\kappa}{2\lambda_{\max}(\mQ)}} \E\left[V(\vz_{\tau})\right]
+\frac{16\left\|\mQ \right\|_2(L_2\tau+C_2^2)h_1^2}{(k-1+h_2)^{h_1\frac{\kappa}{2\lambda_{\max}(\mQ)}}}\int^{k}_{\tau} \left(t+1+h_2 \right)^{h_1\frac{\kappa}{2\lambda_{\max}(\mQ)}-2}dt \label{ineq:decreasing_2} \\
&+2 \left\|  \mQ \right\|_2L_2\tau\alpha_{k-\tau} \alpha_k+ 2\left\|\mQ\right\|_2C_2^2\alpha_k^2 \nonumber\\
\leq & \left(\frac{\tau+h_2}{k+h_2} \right)^{h_1\frac{\kappa}{2\lambda_{\max}(\mQ)}} \E\left[V(\vx_{\tau})\right]
+\frac{16\left\|\mQ \right\|_2(L_2\tau+C_2^2)h_1^2}{(k-1+h_2)^{h_1\frac{\kappa}{2\lambda_{\max}(\mQ)}}} \frac{1}{h_1\frac{\kappa}{2\lambda_{\max}(\mQ)}-1} (k+1+h_2)^{h_1\frac{\kappa}{2\lambda_{\max}(\mQ)}-1} \nonumber  \\
&+2 \left\|  \mQ \right\|_2L_2\tau\alpha_{k-\tau} \alpha_k+ 2\left\|\mQ\right\|_2C_2^2\alpha_k^2 \nonumber\\
\leq & \left(\frac{\tau+h_2}{k+h_2} \right)^{h_1\frac{\kappa}{2\lambda_{\max}(\mQ)}} \E\left[V(\vz_{\tau})\right]
+\frac{16\left\|\mQ \right\|_2(L_2\tau+C_2^2) h_1^2}{k-1+h_2} \frac{2^{h_1\frac{\kappa}{2\lambda_{\max}(\mQ)}-1}}{h_1\frac{\kappa}{2\lambda_{\max}(\mQ)}-1} \label{ineq:decreasing_3}\\
&+2 \left\|  \mQ \right\|_2L_2\tau\alpha_{k-\tau} \alpha_k+ 2\left\|\mQ\right\|_2C_2^2\alpha_k^2. \nonumber
\end{align}

The inequality~(\ref{ineq:decreasing_1}) follows from the fact that \(\alpha_{i-\tau} \leq 2\alpha_i\) for \(\tau \leq i \), which holds since \(\frac{\tau-1+2^{1/E_1h_1}}{2^{1/E_1h_1}-1} \leq h_2 \) and \(2 \leq E_1h_1\). Moreover, \(\frac{i+1+h_2}{i+h_2}\leq 2\) for \(i\geq 0\). This follows from the condition \(h_2>2\tau\), which can be checked from \(h_2 \geq \frac{\tau-1+2^{1/E_1h_1}}{2^{1/E_1h_1}-1}\) and \(h_1E_1 \geq 2\). The inequality~(\ref{ineq:decreasing_2}) holds since \( \frac{4\lambda_{\max}(\mQ)}{\kappa} \leq h_1\). The inequality~(\ref{ineq:decreasing_3}) follows from the fact that \(k+1+h_2\leq 2k-2+2h_2\), which when holds \(h_2 \geq 3\) and it is satisfied by the inequalities \(h_2 \geq \frac{\tau-1+2^{1/E_1h_1}}{2^{1/E_1h_1}-1}\) and \(h_1E_1 \geq 2\).
We can bound \(\E\left[V(\vz_{\tau})\right]\) from Lemma~\ref{lem:z_s-z_k-tau}, which results to

\begin{align*}
    \E\left[\left\|\vz_{k+1}\right\|_2^2\right] 
    \leq & \frac{\lambda_{\max}(\mQ)}{\lambda_{\min}(\mQ)} \left(\frac{\tau+h_2}{k+h_2} \right)^{h_1\frac{\kappa}{2\lambda_{\max}(\mQ)}} \left(2 \left\| \vz_{0} \right\|_2
        +4C_2\tau\alpha_{0} \right)\\
&+\frac{1}{\lambda_{\min}(\mQ)}\frac{16\left\|\mQ \right\|_2(L_2\tau+C_2^2)h_1^2}{k-1+h_2} \frac{2^{h_1\frac{\kappa}{2\lambda_{\max}(\mQ)}-1}}{h_1\frac{\kappa}{2\lambda_{\max}(\mQ)}-1} \\
&+\frac{2 \left\|  \mQ \right\|_2}{\lambda_{\min}(\mQ)}\left(L_2\tau\alpha_{k-\tau} \alpha_k+ C_2^2\alpha_k^2\right) . 
\end{align*}
This completes the proof.
\end{enumerate}





\end{proof}

\subsection{Proof of Theorem~\ref{thm:markov_td}}\label{app:proof:thm_markov_td}
We will provide several building blocks for the main proof. First, for $o\in\gS\times\gS\times\Pi^N_{i=1}I$, let
    \begin{align*}
    \vw(o)& :=\begin{bmatrix}
         r^1(s,\va,s^{\prime})\bm{\phi}(s)-\vb^1 \\
        r^2(s,\va,s^{\prime})\bm{\phi}(s)-\vb^2\\
        \vdots\\
      r^N(s,\va,s^{\prime})\bm{\phi}(s)-\vb^N 
    \end{bmatrix}+ \left(\mI_q \otimes 
    (\gamma \bm{\phi}(s) \bm{\phi}^{\top}(s^{\prime})-\bm{\phi}(s)\bm{\phi}(s)^{\top})-\bar{\mA}\right) \bm{1}_N\otimes \vtheta_c, \\
     \mW(o) &:= \mI_q \otimes 
    (\gamma \bm{\phi}(s)\bm{\phi}^{\top}(s^{\prime})-\bm{\phi}(s)\bm{\phi}(s)^{\top})-\bar{\mA}.
    \end{align*}
Note that  \(   \bm{\bar{\eps}}(o_k;\bar{\vtheta}_k)\) defined in~(\ref{eq:bar_eps}) can be expressed as
\begin{align*}
    \bm{\bar{\eps}}(o_k;\bar{\vtheta}_k) =\begin{bmatrix}
         \mW(o_k)\tilde{\vtheta}_k  + \vw(o_k) \\
         \bm{0}_{Nq}
    \end{bmatrix}.
\end{align*}

\begin{lemma}\label{lem:Ww_bound}
    For $o\in\gS\times\gS\times\Pi^N_{i=1}I$, we have
    \begin{align*}
        \left\|\mW(o) \right\|_2 \leq 6,\quad \; \left\| \vw(o) \right\|_2 \leq  \frac{9\sqrt{N}R_{\max}}{(1-\gamma)w}.
    \end{align*}
\end{lemma}
\begin{proof}
First, we have
    \begin{align*}
        \left\| \mW(o) \right\| =& \left\| \mI_q \otimes 
    (\gamma \bm{\phi}(s)\bm{\phi}^{\top}(s^{\prime})-\bm{\phi}(s)\bm{\phi}(s)^{\top})-\bar{\mA}\right\|_2  \\
    \leq &  \left\|\gamma \bm{\phi}(s)\bm{\phi}^{\top}(s^{\prime})-\bm{\phi}(s)\bm{\phi}(s)^{\top}-\mA \right\|_2\\
    \leq & 6.
    \end{align*}
    The last inequality follows from Lemma~\ref{lem:A_bound} and the assumption that $\left\|\bm{\phi}(s) \right\|_2 \leq 1$ for all $s\in\gS$.

    Moreover, we have
    \begin{align*}
        \left\| \vw(o) \right\|_2 =& \left\| \bm{\bar{\eps}}(o;\bm{1}_N\otimes \vtheta_c )\right\|_2\\
        \leq &  6 \left\| \bm{1}_N\otimes \vtheta_c\right\|_2+ 3\sqrt{N} R_{\max} \\
        \leq &6\sqrt{N}\frac{R_{\max}}{(1-\gamma)w}+ 3\sqrt{N} R_{\max}\\
        \leq & \frac{9\sqrt{N}R_{\max}}{(1-\gamma)w},
    \end{align*}
    where the first equality follows from the definition of \(\bm{\bar{\eps}}\) in~(\ref{eq:bar_eps}). The first inequality follows from~(\ref{ienq:eps_pre}). The second inequality follows from Lemma~\ref{theta_c_bound}.
\end{proof}

\begin{lemma}\label{lem:algo1:o_k-vx_k-tau}
    For \( k \geq \tau\), we have
 \begin{align*}
      \left\|\E\left[ \bar{\bm{\eps}}(o_k;\bar{\vtheta}_{k-\tau})\middle| \gF_{k-\tau} \right]\right\|_2 \leq & \max\left\{ \frac{4R_{\max}\sqrt{Nq}}{w(1-\gamma)} ,2q \right\} \alpha_T (\left\| \bar{\vtheta}_{k-\tau}-\bm{1}_{N} \otimes \vtheta_c \right\|_2+1).
 \end{align*}
\end{lemma}
\begin{proof}
Applying triangle inequality to~(\ref{eq:bar_eps}), we get
    \begin{align*}
         \left\|\E\left[ \bar{\bm{\eps}}(o_k;\bar{\vtheta}_{k-\tau})\middle| \gF_{k-\tau} \right]\right\|_2 \leq  \underbrace{\left\| \E\left[ \vw(o_k)\mid \gF_{k-\tau}\right] \right\|_2}_{I_1}
         +\underbrace{ \left\| \E\left[ \mW(o_k)(\bar{\vtheta}_{k-\tau}-\bm{1}_N\otimes \vtheta_c) \middle| \gF_{k-\tau}  \right] \right\|_2 }_{I_2}.
    \end{align*}

    We will check the conditions to apply Lemma~\ref{lem:geometric_mixing} to bound $I_1$ and $I_2$ separately. Considering $I_1$, note that we have
    \begin{align*}
    &\left\|
        \begin{bmatrix}
         r^1_k\bm{\phi}(s_k) \\
        r^2_k\bm{\phi}(s_k)\\
        \vdots\\
      r^N_k\bm{\phi}(s_k) 
    \end{bmatrix}
    + \left(\mI_q \otimes 
    (\gamma \bm{\phi}(s_k)\bm{\phi}^{\top}(s_k^{\prime})-\bm{\phi}(s_k)\bm{\phi}(s_k)^{\top})\right) \bm{1}_N\otimes \vtheta_c \right\|_{\infty} \\
    \leq & \max_{1\leq i \leq N} \left\| r^i_k \bm{\phi}(s_k) \right\|_{\infty}
    + \left\|\left(\mI_q \otimes 
    (\gamma \bm{\phi}(s_k)\bm{\phi}^{\top}(s_k^{\prime})-\bm{\phi}(s_k)\bm{\phi}(s_k)^{\top})\right) \bm{1}_N\otimes \vtheta_c \right\|_{\infty}  \\ 
    \leq & R_{\max} + \left\| \mI_q \otimes 
    (\gamma \bm{\phi}(s_k)\bm{\phi}^{\top}(s_k^{\prime}) -\bm{\phi}(s_k)\bm{\phi}(s_k)^{\top}) \right\|_2\left\|\bm{1}_N \otimes \vtheta_c \right\|_{\infty}\\
    \leq & R_{\max}+ \frac{2R_{\max}}{(1-\gamma)w}  \\
    \leq & \frac{4R_{\max}}{w(1-\gamma)}.
    \end{align*}

    The second inequality follows from the assumption that \(|r^i_k| \leq R_{\max}\) for \(1\leq i\leq N, k\in \sN_0\), and \(\left\|\bm{\phi}(s)\right\|_2 \leq 1 \) for \(s \in \gS\). The third inequality follows from Lemma~\ref{theta_c_bound}.

    Furthermore, we have, for \(1 \leq i \leq N\),
    \begin{align*}
        \E\left[r^i_k\bm{\phi}(s_k)\right]=&\sum_{s\in \gS} d(s) \bm{\phi}(s) \sum_{s^{\prime}\in \gS}\sum_{\va \in \Pi^N_{i=1}\gA^i}\pi(\va|s) \gP(s,\va,s^{\prime})r^i(s,\va,s^{\prime}) \\
        &= \sum_{s\in \gS} d(s)\bm{\phi}(s) [\mR_i^{\pi}]_s\\
        &= \mPhi^{\top}\mD^{\pi}\mR_i^{\pi},
    \end{align*}

and it is straightforward to check that \( \E\left[\bm{\phi}(s_k)\bm{\phi}^{\top}(s_k^{\prime})-\bm{\phi}(s_k)\bm{\phi}^{\top}(s_k))\right]=\mA \).
    Therefore, from Lemma~\ref{lem:geometric_mixing}, we get
 \begin{align*}
   \left\|    \E\left[ \vw(o_k) \middle | \gF_{k-\tau}\right] \right\|_2 \leq \frac{4R_{\max}\sqrt{Nq}}{w(1-\gamma)}\alpha_T.
 \end{align*}

Now, we will bound $I_2$. Consider the following relations:
 \begin{align*}
      \left\|\E\left[\mW(o_k) \middle | \gF_{k-\tau}\right] \right\|_2 =& \left\|\E\left[ \gamma \bm{\phi}(s_k)\bm{\phi}^{\top}(s_k^{\prime})-\bm{\phi}(s_k)\bm{\phi}^{\top}(s_k)-\mA \middle | \gF_{k-\tau}\right] \right\|_2,
 \end{align*}
 and
  \begin{align*}
   \max_{1\leq i,j \leq q} \left| \left[
    (\gamma \bm{\phi}(s_k)\bm{\phi}^{\top}(s_k^{\prime})-\bm{\phi}(s_k)\bm{\phi}(s_k)^{\top})  \right]_{ij}\right|\leq & \left\|\gamma  \bm{\phi}(s_k)\bm{\phi}^{\top}(s_k^{\prime})-\bm{\phi}(s_k)\bm{\phi}^{\top}(s_k)) \right\|_2
    \leq  2,
 \end{align*}
 
 where the second inequality follows from the assumption that \(\left\|\bm{\phi}(s)\right\|_2 \leq 1\) for \(s \in \gS\). 

 
 From the third item in Lemma~\ref{lem:geometric_mixing}, we have
 \begin{align*}
     \left\| \E\left[\mW(o_k)\middle| \gF_{k-\tau} \right] \right\|_2 \leq 2 q \alpha_T .
 \end{align*}
 Hence, we have
 \begin{align*}
     \left\| \E\left[ \mW(o_k)(\bar{\vtheta}_{k-\tau}-\bm{1}_N\otimes \vtheta_c) \middle| \gF_{k-\tau}  \right] \right\|_2 =& \left\|  \E\left[ \mW(o_k) \middle|\gF_{k-\tau} \right] (\bar{\vtheta}_{k-\tau}-\bm{1}_N\otimes \vtheta_c) \right\|_2\\
     \leq &  \left\|  \E\left[ \mW(o_k) \middle|\gF_{k-\tau} \right]\right\|_2 \left\|\bar{\vtheta}_{k-\tau}-\bm{1}_N\otimes \vtheta_c \right\|_2\\
     \leq & 2 q \alpha_T \left\| \bar{\vtheta}_{k-\tau}-\bm{1}_{N} \otimes \vtheta_c \right\|_2.
 \end{align*}

 Collecting the bounds on $I_1$ and $I_2$, we get
 \begin{align*}
      \left\|\E\left[ \bar{\bm{\eps}}(o_k;\bar{\vtheta}_{k-\tau})\middle| \gF_{k-\tau} \right]\right\|_2 \leq & \frac{4R_{\max}\sqrt{Nq}}{w(1-\gamma)}\alpha_T+ 2 q \alpha_T \left\| \bar{\vtheta}_{k-\tau}-\bm{1}_{N} \otimes \vtheta_c \right\|_2 \\
      \leq & \max\left\{ \frac{4R_{\max}\sqrt{Nq}}{w(1-\gamma)} ,2q \right\} \alpha_T (\left\| \bar{\vtheta}_{k-\tau}-\bm{1}_{N} \otimes \vtheta_c \right\|_2+1).
 \end{align*}
    This completes the proof.
\end{proof}

Now, we are ready to prove Theorem~\ref{thm:markov_td}.
\begin{proof}[Proof of Theorem~\ref{thm:markov_td}.]
To this end, we will apply Theorem~\ref{thm:markovian_general} in the Appendix Section~\ref{app:subsec:markovian}. Let $\vz_k:=\begin{bmatrix}
    \tilde{\vtheta}_k\\
    \bar{\mL}\bar{\mL}^{\dagger}\tilde{\vw}_k
\end{bmatrix}$. Hence, it is enough to check the conditions in Assumption~\ref{assmp:markov_general_sa} in the Appendix Section~\ref{app:subsec:markovian}. The first item in Assumption~\ref{assmp:markov_general_sa} can be checked from Lemma~\ref{lem:Ww_bound}, we have
    \begin{align*}
        C_1:=6 ,\quad C_2:=\frac{9 \sqrt{N} R_{\max}}{w(1-\gamma)}  ,\quad E_1:=8+2\lambda_{\max}(\bar{\mL}).
    \end{align*}

    From Lemma~\ref{lem:algo1:o_k-vx_k-tau}, we have
    \begin{align*}
     \Xi:= \max\left\{ \frac{4R_{\max}\sqrt{Nq}}{w(1-\gamma)} ,2q \right\}  ,
    \end{align*}
    which satisfies the second assumption in Assumption~\ref{assmp:markov_general_sa}. The third item in Assumption~\ref{assmp:markov_general_sa} follows from Lemma~\ref{lem:algo1:lyapunov_equation_for_projected_iterate}. 
\begin{enumerate}
    \item[1.] For constant step-size,  \(K_1\) and \(K_2\) in Theorem~\ref{thm:markovian_general} becomes
    \begin{align*}
        K_1 =&4\max\left\{ \frac{4R_{\max}\sqrt{Nq}}{w(1-\gamma)} ,2q \right\} +624 +152\frac{9 \sqrt{N} R_{\max}}{w(1-\gamma)}+13\cdot 12\lambda_{\max}(\bar{\mL})+\frac{72\sqrt{N}R_{\max}}{w(1-\gamma)}\lambda_{\max}(\bar{\mL}) \\
        =& \gO\left( \max\left\{ \frac{\sqrt{Nq}R_{\max}}{w(1-\gamma)} \lambda_{\max}(\bar{\mL}) ,q\right\}\right), \\
        K_2 =&  2\max\left\{ \frac{4R_{\max}\sqrt{Nq}}{w(1-\gamma)} ,2q \right\}
+  810\frac{NR_{\max}^2}{w^2(1-\gamma)^2}+182\frac{9 \sqrt{N} R_{\max}}{w(1-\gamma)} +\frac{72\sqrt{N}R_{\max}}{w(1-\gamma)}\lambda_{\max}(\bar{\mL}) ,  
    \end{align*}
    which leads to
    \begin{align*}
     \Omega\left( \frac{NR^2_{\max}}{w^2(1-\gamma)^2} \right) \leq K_2 \leq \gO\left( \max\left\{ \frac{N\sqrt{q}R^2_{\max}}{w^2(1-\gamma)^2}\lambda_{\max}(\bar{\mL}) , 2q \right\}\right).
    \end{align*}
    Note that from Lemma~\ref{lem:algo1:lyapunov_equation_for_projected_iterate}, we have \(\left\| \mG  \right\|_2 = \Theta \left( \frac{\lambda_{\max}(\bar{\mL})^2}{(1-\gamma)w} \right)\).
    Therefore, from the step-size condition in the first item in Theorem~\ref{thm:markovian_general}, we need
    \begin{align*}
         &\alpha_0 \\
        \leq & \min \left\{ \frac{1}{900\tau\max\left\{\frac{\sqrt{N}R_{\max}}{w(1-\gamma)},10\lambda_{\max}(\bar{\mL}) \right\}}, \frac{6}{2 \max\left\{ \frac{4R_{\max}\sqrt{Nq}}{w(1-\gamma)} ,2q \right\} } ,\frac{\min\left\{1, \lambda^+_{\min}(\bar{\mL})^2\right\}\lambda_{\min}(\mG)}{(400\lambda_{\max}(\bar{\mL})^2+4K_1\tau )\lambda_{\max}(\mG)\left\|\mG \right\|_2}\right\}.
    \end{align*}
    Hence, there exists \(\bar{\alpha}\) such that \begin{align*}
        \bar{\alpha} = \gO \left(\frac{\min\left\{1, \lambda^+_{\min}(\bar{\mL})^2\right\}(1-\gamma)w}{\tau  \max\left\{ \frac{\sqrt{Nq}R_{\max}}{w(1-\gamma)} ,q\right\} \lambda_{\max}(\bar{\mL})^4}   \right).
    \end{align*}
Therefore, the first item in Theorem~\ref{thm:markovian_general} leads to
    \begin{align*}
    & \frac{1}{N}\left(\E\left[\left\|\tilde{\vtheta}_{k+1}\right\|^2_2\right]+\left\|\bar{\mL}\bar{\mL}^{\dagger}\tilde{\vw}_{k+1}\right\|_2^2 \right)\\
    =& \gO\left(  \exp\left( -\frac{(1-\gamma)w\min\{1,\lambda_{\min}^+(\mL)^2 \}}{\lambda_{\max}(\mL)^2}\alpha_0( k-\tau-1)  \right) \right.\\
    &\left.+  \alpha_0 \tau \max\left\{ \frac{\sqrt{q}R^2_{\max}\lambda_{\max}(\bar{\mL})}{w^3(1-\gamma)^3} , \frac{2q}{N(1-\gamma)w} \right\}    \frac{\lambda_{\max}(\mL)^2 }{\min\{1,\lambda_{\min}^+(\mL)^2 \}}
    \right) .
    \end{align*}
    

\item[2.] For diminishing step-size, we get
\begin{align*}
    L_1=& 4\max\left\{ \frac{4R_{\max}\sqrt{Nq}}{w(1-\gamma)} ,2q \right\} +624 +80\frac{9 \sqrt{N} R_{\max}}{w(1-\gamma)}+13\cdot12\lambda_{\max}(\bar{\mL})+\frac{72\sqrt{N}R_{\max}}{w(1-\gamma)}\lambda_{\max}(\bar{\mL}) \\
        =& \gO\left(\max\left\{ \frac{R_{\max}\sqrt{Nq}}{w(1-\gamma)} \lambda_{\max}(\bar{\mL}) ,2q \right\}\right), \\
    L_2 =& 2\max\left\{ \frac{4R_{\max}\sqrt{Nq}}{w(1-\gamma)} ,2q \right\}+110\frac{9 \sqrt{N} R_{\max}}{w(1-\gamma)}+4 \frac{81 N R^2_{\max}}{w^2(1-\gamma)^2}+\frac{72\sqrt{N}R_{\max}}{w(1-\gamma)}\lambda_{\max}(\bar{\mL}),
\end{align*}
which leads to 
\begin{align*}
    \Omega\left( \frac{NR^2_{\max}}{w^2(1-\gamma)^2}  \right) \leq L_2 \leq  \gO\left(  \max\left\{\frac{N\sqrt{q}R^2_{\max}}{w^2(1-\gamma)^2}\lambda_{\max}(\bar{\mL}),q\right\} \right).
\end{align*}
Following the second item in Theorem~\ref{thm:markovian_general}, the choice of step-size satisfying
\begin{align*}
    h_1 =& \Theta\left(  \frac{\lambda_{\max}(\mL)^2}{(1-\gamma)w\min\{1,\lambda_{\min}^+(\bar{\mL})^2 \}}\right),\\
    h_2 =& \Theta\left(\max\left\{1+\frac{\tau}{2^{1/E_1h_1}-1}, h_1\tau\frac{\sqrt{N}R_{\max}}{(1-\gamma)w},h_1 \frac{\lambda_{\max}(\bar{\mL})^4 \tau}{\min\{1,\lambda_{\min}^+(\bar{\mL})^2 \}}\max\left\{ \frac{R_{\max}\sqrt{Nq}}{w^2(1-\gamma)^2} ,\frac{2q}{w(1-\gamma)}\right\} \right\}  \right),
\end{align*}
yields
\begin{align*}
   &\frac{1}{N}\left(\E\left[\left\|\tilde{\vtheta}_{k+1} \right\|^2_2\right]+\left\|\bar{\mL}\bar{\mL}^{\dagger}\tilde{\vw}_{k+1}\right\|_2^2 \right) =\gO\left(  \frac{\tau}{k} \frac{qR_{\max}^2}{w^4(1-\gamma)^4} \frac{\lambda_{\max}(\mL)^5}{\min\{1,\lambda_{\min}^+(\mL)^2 \}^2}  \right).
\end{align*}
This completes the proof.
\end{enumerate}

\end{proof}

%% file: app/exp.tex
\section{Comparison with other algorithms}\label{app:exp:comparison}

\begin{figure}[!h]
     \centering
     \begin{subfigure}[!ht]{0.49\textwidth}
         \centering
         \includegraphics[width=\textwidth]{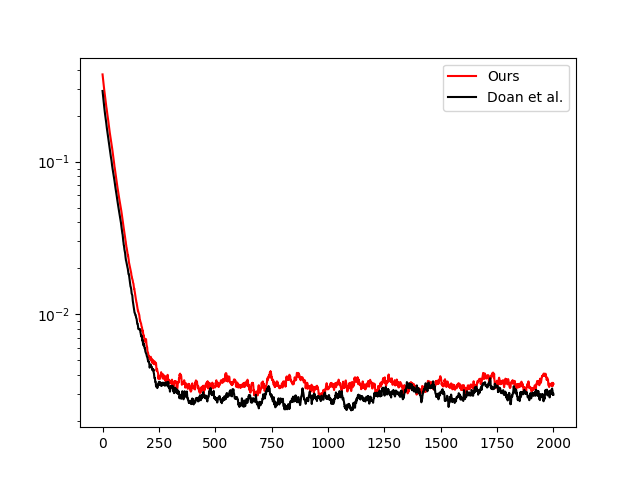}
         \caption{Number of agents 8}
         \label{fig:comparionson-least-squares-N-8}
     \end{subfigure}
          \hfill
          \begin{subfigure}[!ht]{0.49\textwidth}
         \centering
         \includegraphics[width=\textwidth]{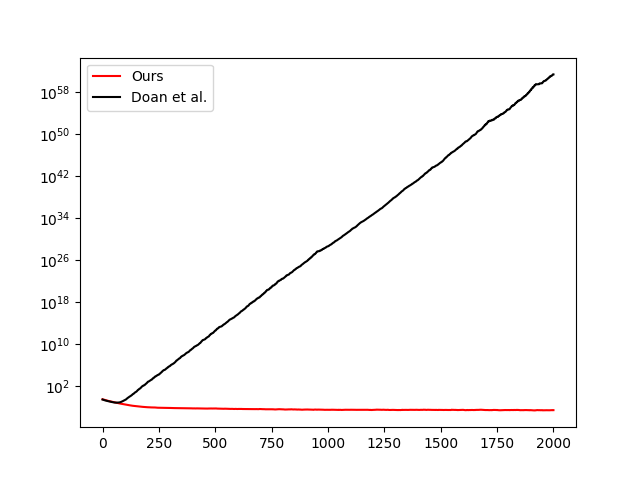}
         \caption{Number of agents 32}
         \label{fig:comparionson-least-squares-N-32}
     \end{subfigure}
     \caption{The doubly stochastic matrix was constructed by solving a least squares problem~\citep{bai2007computing}. We did not plot the result of~\citealp{wang2020decentralized}, since it diverges. The step-size was chosen as $1/2^3$.}\label{fig:least_squares_doubly_stochastic}
\end{figure}

\begin{figure}[!h]
     \centering
     \begin{subfigure}[!ht]{0.49\textwidth}
         \centering
         \includegraphics[width=\textwidth]{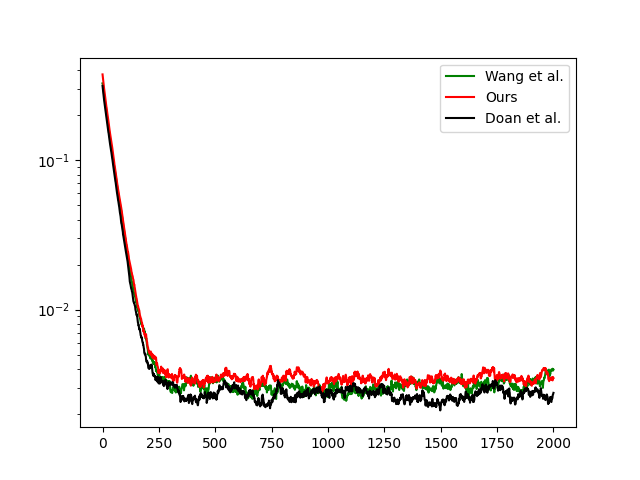}
         \caption{Number of agents 8}
         \label{fig:comparionson-sinkhorn-N-8}
     \end{subfigure}
          \hfill
          \begin{subfigure}[!ht]{0.49\textwidth}
         \centering
         \includegraphics[width=\textwidth]{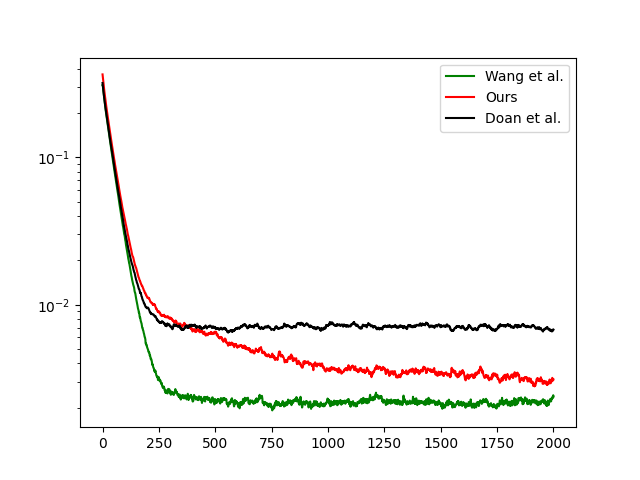}
         \caption{Number of agents 32}
         \label{fig:comparionson-sinkhorn-N-32}
     \end{subfigure}
     \caption{The doubly stochastic matrix was constructed by Sinkhorn-Knobb algorithm~\citep{knight2008sinkhorn}. The step-size was chosen as $1/2^3$.}\label{fig:sinkhorn}
\end{figure}

To compare the performance with other algorithms, we have experimented under the setting in Section~\ref{sec:exp} on cycle graph, and the rewards are generated uniformly random between $(0,1)$. The results are given in Figure~(\ref{fig:least_squares_doubly_stochastic}) and Figure~(\ref{fig:sinkhorn}). Note that the performance of distributed TD algorithms in~\citealp{wang2020decentralized} and~\citealp{doan2019finite} depend on the choice of doubly stochastic matrix. For example, when the doubly stochastic matrix was constructed by least squares method~\citep{bai2007computing}, there are divergent cases as can be seen from Figure~(\ref{fig:least_squares_doubly_stochastic}). 


\section{Additional experimental results}

\begin{figure}[!ht]
     \centering
     \begin{subfigure}[!ht]{0.49\textwidth}
         \centering
         \includegraphics[width=\textwidth]{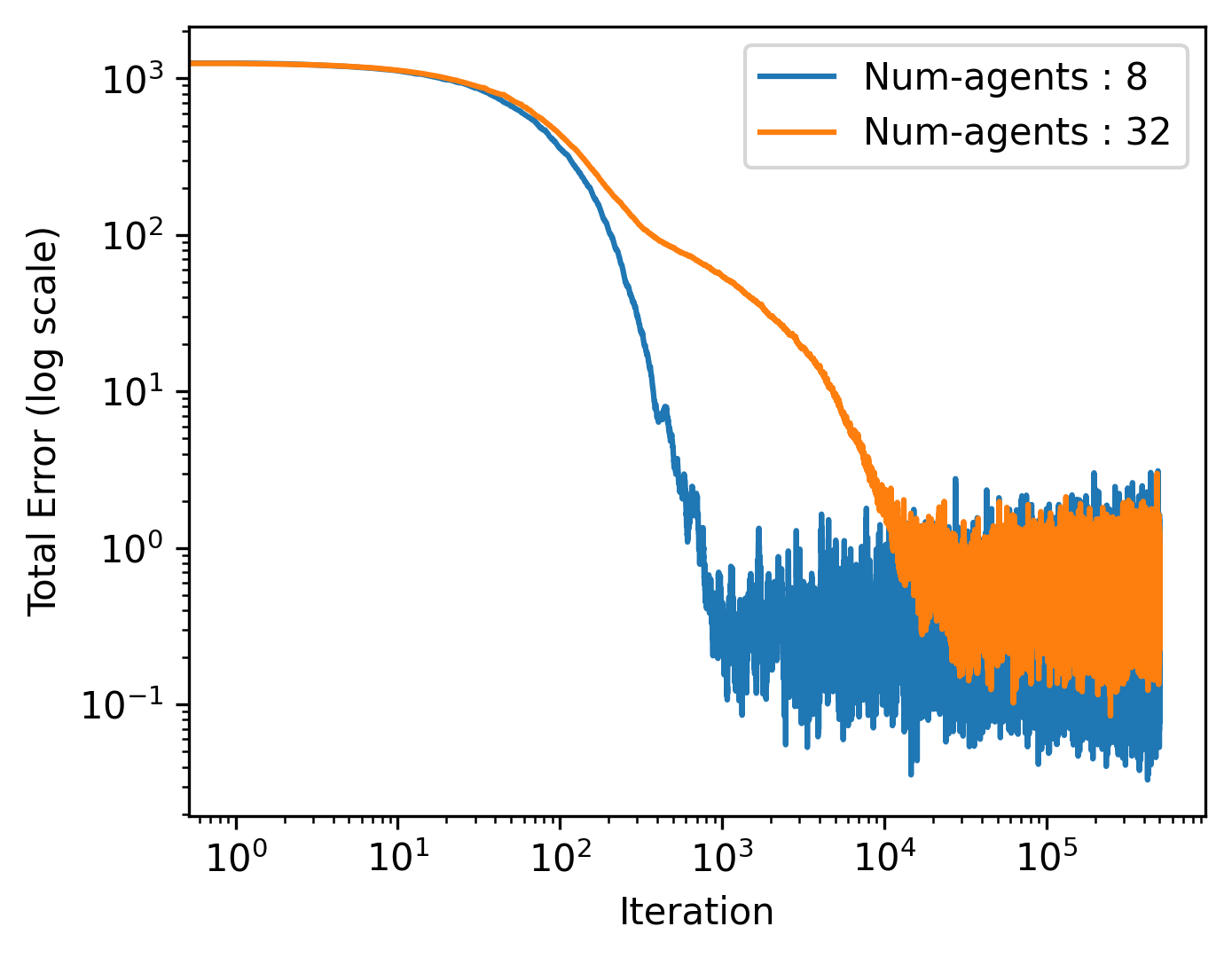}
         \caption{Full plot for Figure~(\ref{fig:step-size}) with step-size $1/2^4$.}
         \label{fig:step-size-2}
     \end{subfigure}
     \hfill
          \begin{subfigure}[!ht]{0.49\textwidth}
         \centering
         \includegraphics[width=\textwidth]{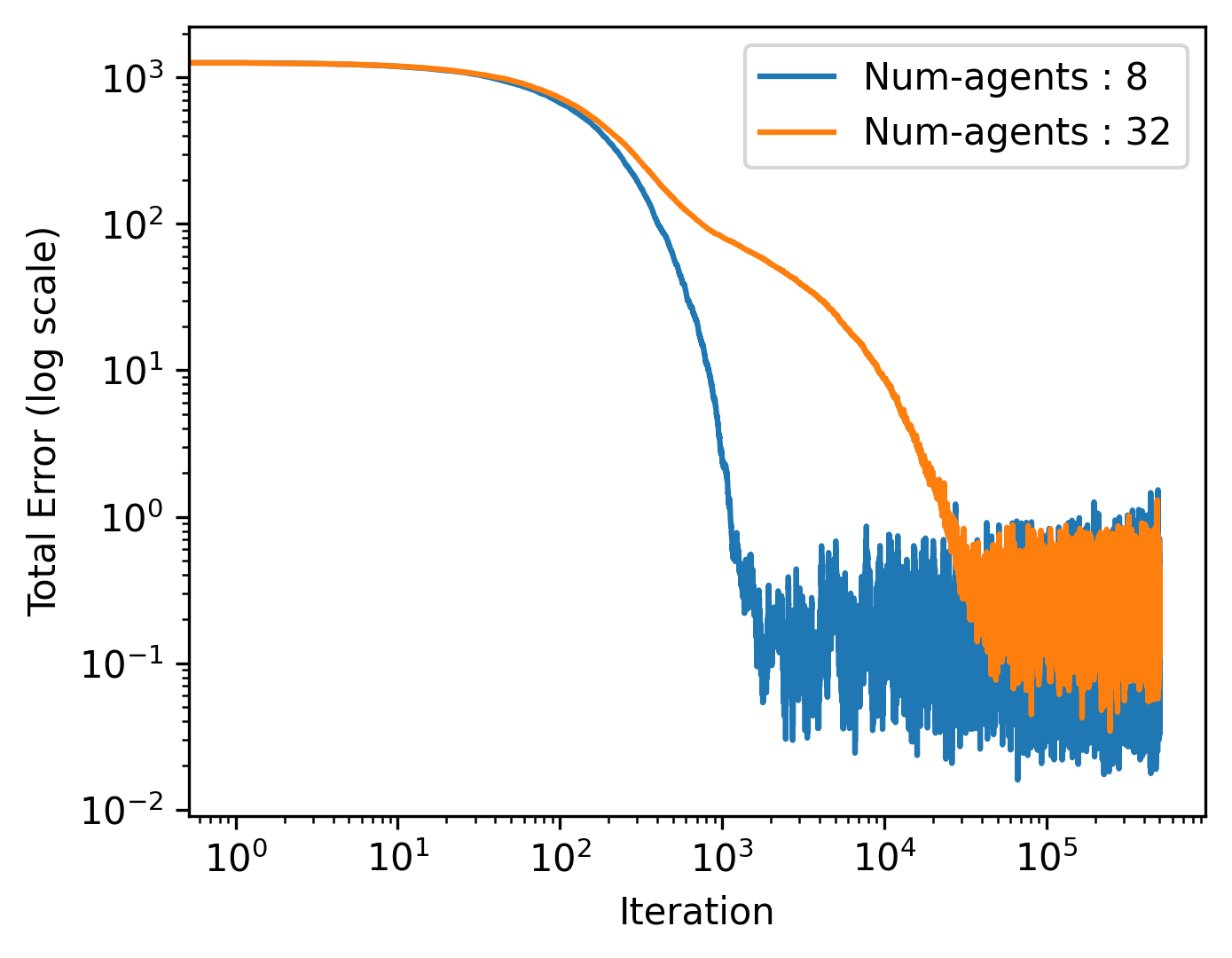}
         \caption{Full plot for Figure~(\ref{fig:step-size}) with step-size $1/2^5$.}
         \label{fig:step-size-3}
     \end{subfigure}
          \hfill
          \begin{subfigure}[!ht]{0.49\textwidth}
         \centering
         \includegraphics[width=\textwidth]{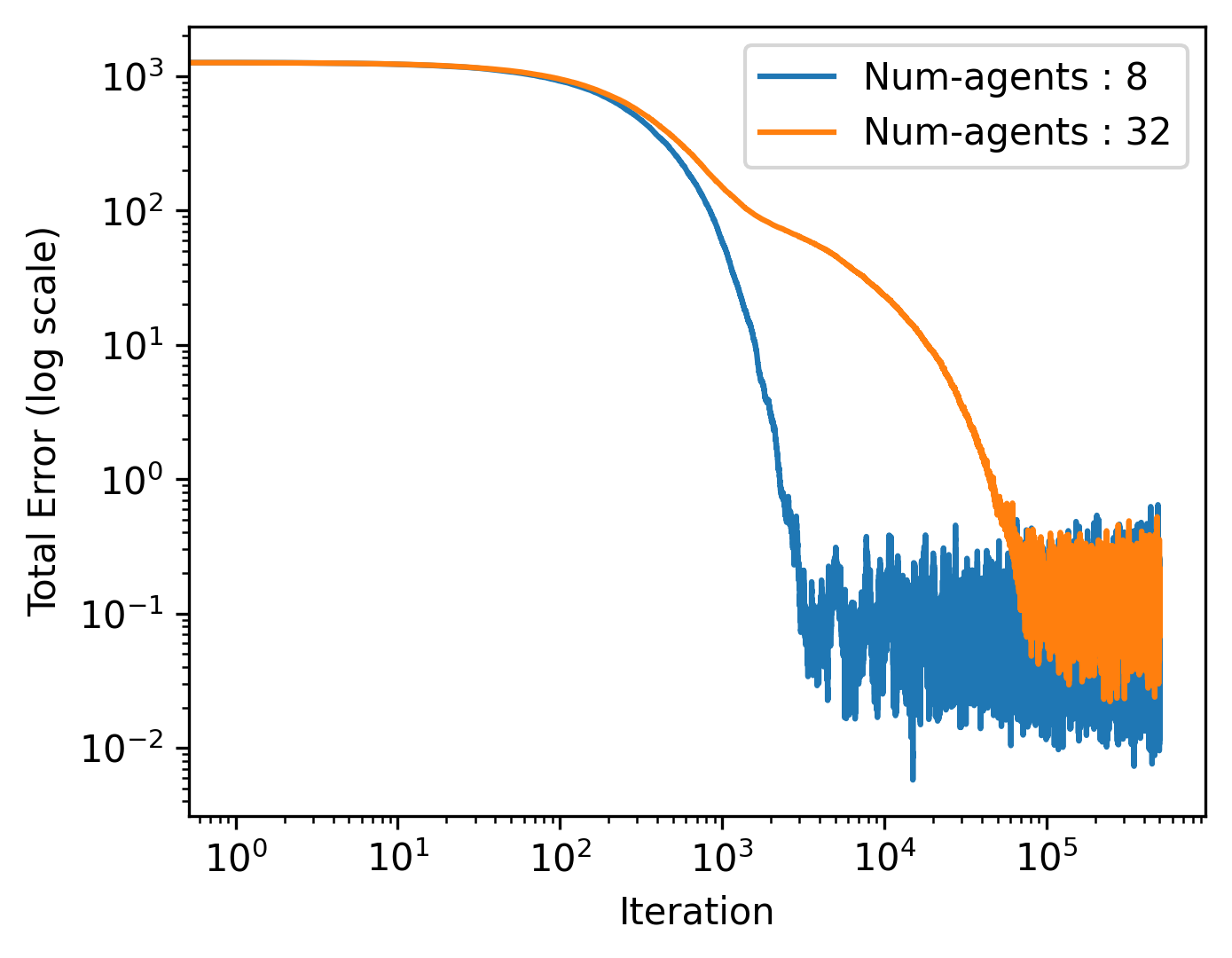}
         \caption{Full plot for Figure~(\ref{fig:step-size}) with step-size $1/2^6$.}
         \label{fig:step-size-4}
     \end{subfigure}
     \caption{Full plots for the result in Figure~(\ref{fig:step-size}). }\label{fig:constant-step-size-ring}

\end{figure}